\documentclass{article}
\usepackage{jfrExamplee}
\usepackage[utf8]{inputenc}
\usepackage{graphicx,apalike,setspace}
\usepackage{amsmath,xcolor,subcaption,hyperref}
\usepackage{algorithm}
\usepackage{algorithmic,latexsym}
\usepackage{placeins,enumitem,epstopdf}
\graphicspath{{./figures/}}

\title{Autonomous Cooperative Multi-Vehicle System for Interception of Aerial and Stationary Targets in Unknown Environments}

\author{Lima Agnel Tony\thanks{Guidance, Control, and Decision Systems Laboratory (GCDSL), Department of Aerospace Engineering,
Indian Institute of Science,
Bangalore-560012, India.}\\
 Doctoral Research Scholar\\
\texttt{limatony@iisc.ac.in} 
\And
Shuvrangshu Jana$^\star$\\
Post-Doctoral Fellow\\
\texttt{shuvra.ce@gmail.com}
\And
Varun V. P.\thanks{Robert Bosch Center for Cyber-Physical Systems, Indian Institute of Science,
Bangalore-560012, India.} \\
Technical Associate\\
\texttt{varunvp@iisc.ac.in} \\
\And
Aashay Anil Bhise$^\star$\\
Project Associate \\
\texttt{meetaashay3@gmail.com} 
\And
Aruul Mozhi Varman S.$^\dagger$ \\
Project Assistant \\
\texttt{aruuls@iisc.ac.in} 
\And
Vidyadhara B. V.$^\star$\\
Project Assistant \\
\texttt{vidyadhara.vk@gmail.com} \\
\And
Mohitvishnu S. Gadde$^\star$ \\
Project Assistant \\
\texttt{mohitvishnug@iisc.ac.in} 
\And
Raghu Krishnapuram$^\dagger$\\
Distinguished Member of Technical Staff\\
\texttt{kraghu@iisc.ac.in} 
\And
Debasish Ghose$^\star$ \\
Professor \\
\texttt{dghose@iisc.ac.in} \\
}
\begin{document}

\maketitle

\begin{abstract}
This paper presents the design, development, and testing of hardware-software systems by the IISc-TCS team for Challenge 1 of the Mohammed Bin Zayed International Robotics Challenge 2020. The goal of Challenge 1 was to grab a ball suspended from a moving and maneuvering UAV and pop balloons anchored to the ground, using suitable manipulators. The important tasks carried out to address this challenge include the design and development of a hardware system with efficient grabbing and popping mechanisms, considering the restrictions in volume and payload, design of accurate target interception algorithms using visual information suitable for outdoor environments, and development of a software architecture for dynamic multi-agent aerial systems performing complex dynamic missions. In this paper, a single degree of freedom manipulator attached with an end-effector is designed for grabbing and popping, and robust algorithms are developed for the interception of targets in an uncertain environment. Vision-based guidance and tracking laws are proposed based on the concept of pursuit engagement and artificial potential function. The software architecture presented in this work proposes an Operation Management System (OMS) architecture that allocates static and dynamic tasks collaboratively among multiple UAVs to perform any given mission. An important aspect of this work is that all the systems developed were designed to operate in completely autonomous mode. A detailed description of the architecture along with simulations of complete challenge in the Gazebo environment and field experiment results are also included in this work. The proposed hardware-software system is particularly useful for counter-UAV systems and can also be modified in order to cater to several other applications.
\end{abstract}

\section{Introduction}\label{s1}
Autonomous systems are becoming an integral part of human society due to rapid advancements in hardware, software, and algorithms; and their apparent utility in many important applications. As a result, autonomous and semi-autonomous robots have been quietly replacing humans in many daily tasks, which were repetitive in nature. Aerial robotics is one such evolving field where many complex applications are being addressed by researchers worldwide. Mohammed Bin Zayed International Robotics Competition \cite{MBZ20} is a platform that provides exciting problems to roboticists. The challenges involved in this competition provide solutions to many practical applications involving aerial and ground robots. MBZIRC 2020 has three challenges, addressing aerial and ground target interception, wall construction, and indoor and outdoor fire fighting.  
This paper presents a complete solution to the problems in Challenge 1, which involves autonomously searching, detecting, and grabbing a maneuvering target as the primary task. The secondary task is to search, detect, and intercept stationary targets randomly located above the ground in the mission environment. The primary goal of Challenge 1 is to grab a softball hanging from a moving target drone. The target moves continuously in a figure-of-eight path. The figure-of-eight can be in any orientation in space. The ball is yellow and 10 cm in diameter. The drone carrying the ball travels at a maximum speed of 10 m/s. Three UAVs which satisfy the volume constraint of 1.2 m $\times$ 1.2 m $\times$ 0.5 m can be used for the mission. Before take-off, the UAV should satisfy the volume constraint but can extend appendages in the air. For safety reasons, the cap on the UAV speed is 30 km/h. The UAVs are assigned to take-off from a marked 10 m $\times$ 10 m square in the middle of the arena. The UAV should autonomously detect the ball, approach, grab the ball hanging from the target drone and land back in the 10 m $\times$ 10 m area. The grabbing force required is less than 4 N.

The secondary target in this challenge is the popping of balloons attached to the top of 1.5 m high poles. The balloons are 65 cm in diameter and green in color. A maximum of five balloons would be placed at various random locations within the arena. The challenge is to be completed within 15 minutes. The summary of Challenge 1 specifications are given in Table \ref{t1}.
 
\begin{table}[h]
	\centering
	\caption{MBZIRC 2020 Challenge 1 specifications}
	\begin{tabular}{|l|c|l|}
        \hline
		Parameter &Value&Remarks  \\
		\hline\hline
		Arena & 100 m $\times$ 40 m $\times$ 20 m & Covered with net\\\hline
		Target UAV trajectory & Repetitive figure-of-eight & Altitude: 5 m - 15 m\\\hline
		Figure-of-eight spec.& Any orientation& Center fixed\\\hline
		Target UAV speed & 3 m/s or 7 m/s & First 7 min/rest 8 min\\\hline
		Ball color & Yellow & Softball \\\hline
		Ball specifications & 10 cm dia., 60 g & -\\\hline
		String for ball&1.45 m& Steel wire\\\hline
		Balloon specifications & 60 cm dia., Green color & Altitude: 5 m\\\hline
		Balloon rod & 1.5 m from ground & 5 nos., randomly placed\\\hline
		Time & 15 min & 5 min to set up\\
		\hline
	\end{tabular}
	\label{t1}
\end{table}

For the successful execution of Challenge 1, the primary building blocks are aerial manipulator hardware design, target detection, tracking, aerial interception, and coordination among the agents through mission planner. An aerial manipulator's design for grasping a dynamic aerial target is challenging due to various issues such as stability, control, and accuracy requirements. Moreover, aerial manipulation with UAVs has varied challenges like instability due to the forces and moments experienced during the manipulation, being an under-actuated system with nonlinear coupled dynamics, and payload constraints.

In this paper, the above-mentioned challenges are addressed by developing an efficient and stable aerial manipulation mechanism, robust and computationally efficient tracking and grabbing algorithms using visual information, and a multi-agent software architecture, called Operation Management System (OMS), for allocating tasks to UAVs ensuring zero conflicts and maximum safety. An indigenously designed and manufactured passive gripper with linear actuation is attached to a hexacopter for grabbing and interception. A 3D guidance law based on the idea of pursuit guidance is developed for the interception by using visual information from a monocular camera. An adaptive controller is used to handle the uncertainties in system dynamics during the grabbing phase and wind disturbance. A computationally efficient and robust vision algorithm is also developed, which provides precise detection of the target at varied lighting conditions, improving the algorithms' performance. The OMS is developed using a multi-master network in such a way that many critical tasks like collision avoidance are addressed in high priority without affecting the primary mission, such as grabbing. Exploring, tracking, and grabbing tasks are designed for multi-UAV collaboration without compromising the overall system's efficiency and redundancy. 
 All these algorithms are tested on Gazebo, a rigid body physics simulator, taking care to simulate the physical and sensor parameters like a manipulator attachment, the target ball hanging from the target UAV, balloons, and camera vision parameters. The complete Challenge 1 mission is simulated in the Gazebo environment using the software architecture developed on ROS. The functionality of individual algorithms is first checked through simulations, by incorporating models of sensors and manipulation mechanism in the virtual environment. The proposed hardware and software solution is tested outdoors after successful Software-In-The-Loop simulations (SITL) and Hardware-In-The-Loop simulations (HITL). The paper reports the flight test results highlighting the major aspects of the solution.

The remainder of this report is organized as follows. Section \ref{s2} gives a detailed literature survey. Section \ref{s3} presents the major challenges involved, and Section \ref{s4} presents the algorithms developed to address these challenges. Section \ref{s5} describes the mission modes associated with the software architecture, while Section \ref{s6} introduces the OMS and describes it in detail. Section \ref{s7} presents the  ROS simulations. Section \ref{s8} details the hardware design, and Section \ref{s9} brings out the experimental results. A brief discussion about this challenge and developed framework is given in Section \ref{s10} and Section \ref{s11} concludes the paper.

\section {Relevant Literature}\label{s2}
In this section, we briefly survey the literature that addresses the problems related to some components of Challenge 1.  A comprehensive survey of aerial manipulation mechanisms and UAV platforms, along with design methods, are given in \cite{khamseh2018aerial}. UAV manipulators for specific purposes like object extraction, inspection and transportation are described in \cite{suarez2016lightweight,ramon2016extracting,suarez2019compliant,suarez2020benchmarks}. Design of  aerial grasping system is reported in \cite{papachristos2014efficient,ramon2020grasp}; but only for stationary objects. A multi-link manipulator will have a high variation of center of gravity and moment of inertia during the manipulation mechanism movement than a body-fixed manipulator and can have stability issues during grabbing from a maneuvering target. The manipulator's design as a part of the UAV frame is reported in several works \cite{tsukagoshi2015aerial,alexis2016aerial}. However, these are not feasible solutions for Challenge 1 as the vehicle might crash due to the steel wire getting entangled in the propellers while attempting to grab the target. A manipulator as a rigid tool attached to the UAV is reported in 
\cite{nguyen2015mechanics,gioioso2014turning,fumagalli2014developing}, which are designed to interact with the physical environment.  In \cite{gioioso2014turning},  a quadrotor system is designed to exert force on its surrounding environment, at near hovering conditions.  In \cite{papachristos2014efficient}, a similar application is presented, where the thrust vectoring capability of a Tri-Tiltrotor UAV is utilized to achieve manipulation. There was no restriction on the maximum size in all these cases, and the relative motion between the manipulator's end effector and the target object is not aggressive.

Visual object tracking is crucial in semantic video interpretation, and several different approaches have been developed in recent years. Correlation filters were used to model the visual appearance of an arbitrary target, and correlation operation was used as an effective approach to achieving tracking-by-detection \cite{5539960}. Numerous techniques have been developed to improve the computational efficiency and tracking capabilities of a correlation filter-based tracker by employing multi-channel versions \cite{6751493}, spatial boundaries \cite{7410847}, deep features \cite{8100216}, and particle filter based methods \cite{8099995}. The correlation filter based approaches are simple to implement and computationally lighter. Though these approaches can run in real time on most modern hardware, they are limited in handling occlusions, scale variations, and variations in appearance. Considering their limitations in object detection, these approaches are not adequately equipped to track high speed objects. Hence, a two-step approach to tracking high speed targets with deep neural networks to perform object detection is employed to overcome the above-mentioned challenges \cite{7533003,8296962}. The first step involves detecting and localizing all the objects of interest in the image using object detectors such as TinyYOLOv3 \cite{redmon2018yolov3}. The second step is to uniquely identify the actual objects corresponding to the detections and associate their newer detections with their trackers over time. A Kalman filter is used to maintain the track, and the Hungarian algorithm is used to accomplish the association task. Object re-identification features (Re-ID features) \cite{10.1007/978-3-642-33863-2_39} and Intersection over Unions (IoU) or Mahalanobis distance among the detections and trackers' predictions are used to compute the cost matrix to associate detections to object trackers across multiple frames.

Traditionally, handcrafted features like color histogram \cite{10.1007/978-3-319-10584-0_1}, HOG \cite{7569092}, and dense SIFT \cite{6751425} were used as Re-ID features. With advancements in deep learning, the regions corresponding to the bounding boxes are fed through an identity embedding deep neural network \cite{6909421} to learn the Re-ID features. These deep learning based object tracking techniques typically require two deep models addressing object detection and Re-ID feature computation. Considering the fact that all the UAV detections are used in the initial phases to search for the attached target ball and their information is not critical once the target ball is located, the computational complexity involved in using Re-ID features in the association metric out-weighs its benefits in our case. Also, the Re-ID features are not helpful in balloon detection and tracking as the balloons are similar in size and appearance.

In general, vision based interception or grabbing is achieved by generating control/guidance law based on the desired position of the target in the image plane or using a planning strategy after estimating the target future pose. Target interception using visual information is reported in the literature with techniques such as the Image-Based Visual Servo (IBVS) technique using the target pixel feedback from an image plane for \cite{kim2016vision,mebarki2014image}, guidance law design based on the differential error between the desired and actual target pixel co-ordinates \cite{mehta2015vision,stepanyan2007adaptive}, and developing path-planning strategies after the prediction of target future position and velocities  \cite{cheung2015visual,triharminto2013adaptive}. Visual guidance using the concept of Image-Based Visual Servoing (IBVS) for aerial manipulation using UAVs is reported in \cite{kim2016vision,mebarki2014image,lee2012adaptive,thomas2014toward}, where the control law is developed based on the feedback directly from the image co-ordinates. The position-Based Visual Servoing (PBVS) method is used in another approach, which requires the reconstruction of the target pose with reference to interceptor UAV\cite{chaumette2006visual}. Vision based terminal guidance system using a monocular camera is developed using the error between desired and actual pixel co-ordinates \cite{mehta2015vision}, where target acceleration is considered as a time-varying disturbance. 
In \cite{stepanyan2007adaptive}, visual tracking of maneuvering target is proposed using an adaptive disturbance rejection controller where target velocity is considered as a time-varying disturbance.  An adaptive guidance law is proposed to intercept a maneuvering target in 3D 
space using LOS angular rate and target image measurements from seeker information \cite{tian2011vision}. 
Generally, it isn't easy to attach a seeker to a multi-rotor UAV system.  However,  the proposed algorithms in \cite{mehta2015vision,stepanyan2007adaptive,tian2011vision} are not tested on real images for a real interception scenario.

The interception of a moving target after estimation of the target future position and velocities is reported in \cite{kumar2017anticipated,khan2018multi}. Visual servoing based control strategy is used to capture the target after estimating the target's real time pose using the adaptive extended Kalman filter in \cite{dong2016autonomous}.  In \cite{strydom2015uav},  Kalman filter-based estimation using stereo vision is employed to intercept moving targets. In  \cite{cheung2015visual}, target position and velocities are estimated using optical flow, and then an adaptive controller is used to grab the target using the robotic manipulator. Interception of moving target using visual information is proposed in \cite{zhang2016vision}, where an online path planning strategy is used to intercept the predicted pose of target considering the target motion model as time series polynomials. In \cite{jin2018motion}, motion planning method is used to capture a tumbling satellite using visual information. In \cite{triharminto2013adaptive},  target interception in 3D is developed using dynamic path planning based on the Dubins algorithm.  Trajectory optimization technique   is  used for target interception in \cite {kritsky2019model,garcia2019minimal}. Learning-Based Model Predictive Control (LBMPC) is used to catch a ball with quadrotor in an indoor environment \cite{bouffard2012learning}. In the above cases, the algorithm is designed based on the pixel error between the desired position and the actual projection of the target on the image plane or involves the target's depth information. Any interception algorithm based on the pixel error/ target depth will be significantly affected by the pixel noise, and it could fail to achieve the terminal accuracy required for grabbing in the outdoor environment.  Grabbing an object from a maneuvering target in an uncertain outdoor environment with space, size, and payload restrictions is a far more challenging problem to solve and have not been addressed in the above papers.

Software architecture for mobile multi-robots operation are proposed in several works in the literature such as ALLIANCE-ROS  \cite{Li2016ALLIANCEROSAS}, MAFOSS \cite{Jones2019MAFOSSMF}, AMEB \cite{Gil2019ACA}, Robosmith \cite{Floroian2010RoboSmithWN}, 
\cite{Efremov2020ArchitectureOS}. Aerial robots pose further  challenges due to restrictions in communication bandwidth to satisfy the payload constraints. 
Various software architectures reported in the literature for multi-agent operation \cite{dwiyasa2020heterogeneous,guerreiro2019mission,SnchezLpez2016AEROSTACKAA,Antonelli2014ControlSA} discuss the development of a mission planner for single and multiple robots. Many of these address the aspects of multi-robot handling, but neither contributes an end-to-end solution. Such a solution would comprise algorithms and logic necessary for conflict-free task allocation among multiple agents, with the added features of redundancy and safety. Software frameworks developed for multi UAS operation are reported in the literature, such as AEROSTACK \cite{SnchezLpez2016AEROSTACKAA}, CAVIS \cite{Antonelli2014ControlSA}; however, a complex mission like Challenge 1 is not demonstrated.  In Challenge 1, the mission phases require the allocation of tasks among the agents in a dynamic environment that needs to be carried out over communication with limited bandwidth.

\section{Main Tasks and Major Challenges}\label{s3}
The main tasks for the successful execution of Challenge 1, adhering to the prescribed constraints, are listed below.
\begin{enumerate}
    \item Design of an efficient grabbing mechanism while satisfying the volume constraints.
    \item Design of a guidance law for approaching and grabbing the object from a maneuvering target using visual information.
    \item Design of a system architecture for the operation of multiple UAVs. 
\end{enumerate}
Since the ball is suspended below the target UAV, close proximity operations are required, which increases the complexity of the mission. Apart from the above requirements, we consider the following aspects for designing an overall system.
\begin{enumerate}
\item Algorithm requirements
  \begin{enumerate}
      \item  Algorithms should have real time performance.
      \item  For a reliable mission, the designed algorithms should be robust enough to handle the uncertainty in visual information. 
       \item The uniformly colored target ball lacks rich visual cues and could be indistinguishable from background objects of similar color. The vision algorithm ought to use additional cues like the location of the target UAV and information from multiple frames to get a reliable location of the target ball. In this case, the algorithm should be resilient to variations in shape, size, and appearance of the UAV. Also, as the UAVs are maneuvering at high speeds, the vision algorithm should be capable of detecting the far-off targets rapidly and reliably to allow sufficient time to generate an appropriate response.
       \item  Glare due to lens flare in the outdoor environment can have an impact on color segmentation. Color being one of the few visual features of the target ball, the vision algorithm should do its best to account for visual artifacts in the segmentation process.
  \end{enumerate}
\item  Hardware considerations
    \begin{enumerate}
        \item The manipulator configured at one side of the UAV could avoid a head-on collision with the incoming target, in case of a failed grabbing attempt. An added benefit of the sideways manipulator configuration, which we propose, is the improved redundancy of the system, by deploying the same UAV for both grabbing and balloon popping in case of any failures. Compared to a manipulator located in the front of the UAV, the sideways design maintains the pitch stability; however, it can reduce the lateral stability. The stability issues that may arise during the mission can be handled by using an adaptive controller. 
        \item Monocular camera, instead of a stereo-camera, could be used to reduce the computational load as well as delay involved in the vision module output. The vision algorithms should be quick and light, which plays a critical role in the interception of maneuvering targets.
        \item Single degree of freedom (DoF) manipulator is preferable over multi-link manipulators, as in the latter, the computational load for command generation is high and the response is comparatively sluggish for capturing maneuvering targets.
    \end{enumerate}

\item Software architecture  
    \begin{enumerate}
         \item The architecture should be capable of scheduling low-level tasks like inter-agent collision avoidance, obstacle avoidance, etc., as high priority while performing high-level mission tasks like exploration, tracking, grabbing, etc., for safe multi-UAV missions. 
         \item  The task allocation framework should be able to schedule the static and dynamic tasks among the agents dynamically without conflicts.
         \item The architecture should include basic fail-safe strategies to handle unprecedented situations.
         \item They should be real time with good computational efficiency.
    \end{enumerate}
 \end{enumerate}

\section{Algorithm Design}\label{s4}
\begin{figure}[htb!]
\centering
\begin{subfigure}{0.45\columnwidth}
    \includegraphics[scale=0.25]{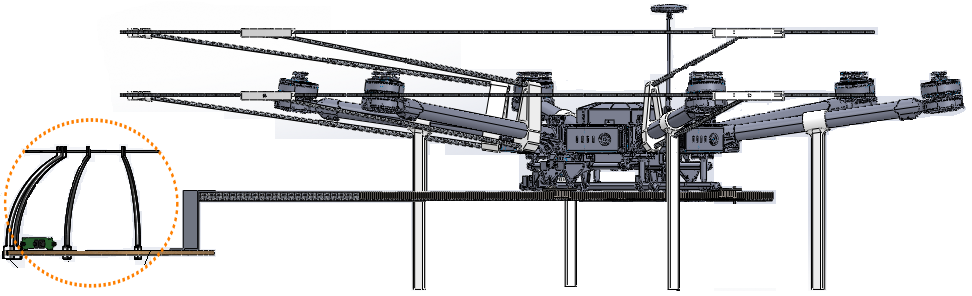}
    \caption{}
    \label{fig:SWG}  
\end{subfigure}
\begin{subfigure}{0.45\columnwidth}
    \includegraphics[scale=0.25]{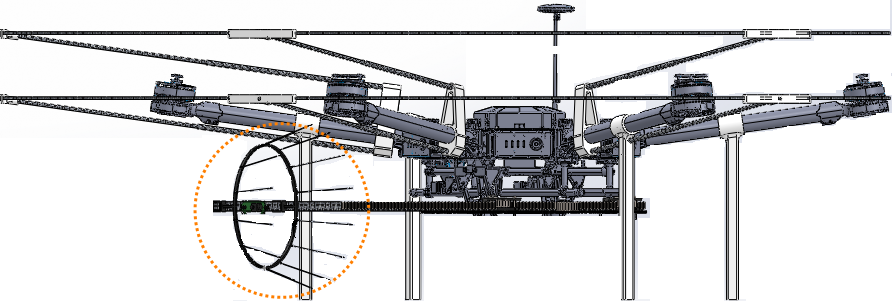}
    \caption{}
    \label{fig:SWP}  
\end{subfigure}
\caption{Conceptual designs of (a) Ball grabbing (b) Balloon popping manipulators. The proposed sideways mechanisms with their respective end effector can be seen.}
\label{fig:HD_concept}
\end{figure}

To capture an object from a high-speed maneuvering target, algorithms are developed for grabbing, target tracking, estimation, and prediction of the target location. The popping of a balloon involves its detection and an algorithm for stationary target interception using visual information. Accurate path following, inter-agent collision avoidance, and geo-fencing are important features to be included in system design to ensure the safety of multi-UAV operation in close proximity. The controller block should handle the variation in system dynamics over the mission due to wind or change in system parameters. Based on hardware considerations, the conceptual design of the UAV's manipulator system for grabbing and popping is developed as given in Fig. \ref{fig:HD_concept}. The detailed design is described in Section \ref{s8}. Algorithms are initially developed based on these conceptual designs and later modified based on flight test observations.

\subsection{Object detection}
This section presents the algorithms for detecting UAV, ball, and balloon. Monocular vision based method is developed to achieve the detection with good accuracy.
\subsubsection{UAV detection}
\begin{figure}[t]
     \centering 
     \includegraphics[scale=0.5]{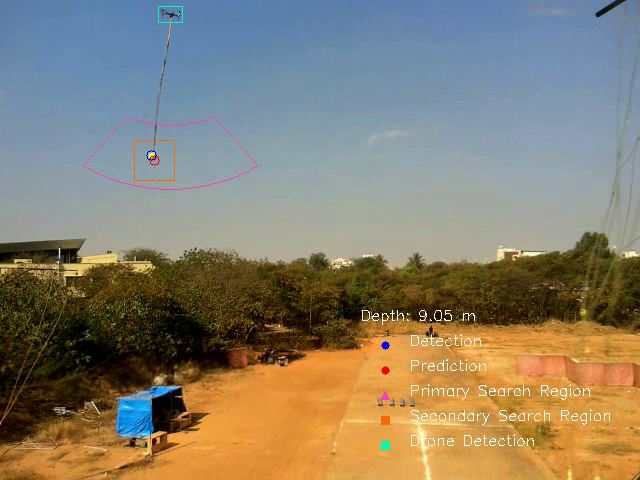}
     \caption{A snapshot showing various regions of interest for the vision module. The color of the ball is yellow.}
     \label{fig:ball_explained}
 \end{figure}

Detecting targets like UAVs flying outdoors can be difficult using traditional techniques due to its variations in appearance caused by sunlight, high velocity of the UAVs, and orientation of the camera. The proposed two-step tracking approach uses TinyYOLOv3 \cite{redmon2018yolov3}, an object detection neural network to detect the UAVs and the Kalman filter to track the targets. In the cases of multiple observations, Euclidean distance between the predictions and observations is used by the Hungarian algorithm to associate the observations with the trackers. The tracker uses linear velocity models to approximate the inter-frame displacement of the targets, and the state of the targets are modeled as:
\begin{equation*}
    [Cx, Cy, w, h]^T 
\end{equation*}
where, \textit{Cx} and \textit{Cy} are $x$ coordinate and $y$ coordinate of the bounding box centre and \textit{w} and \textit{h} are width and height of the bounding box, respectively. The algorithm detects UAV as shown in Fig. \ref{fig:ball_explained}.

\subsubsection{Ball detection} 
The vision module for ball detection and tracking involves three parts - Search, Tracking and Grab. 
\begin{figure}[htb!]
     \centering
     \includegraphics[scale=0.65]{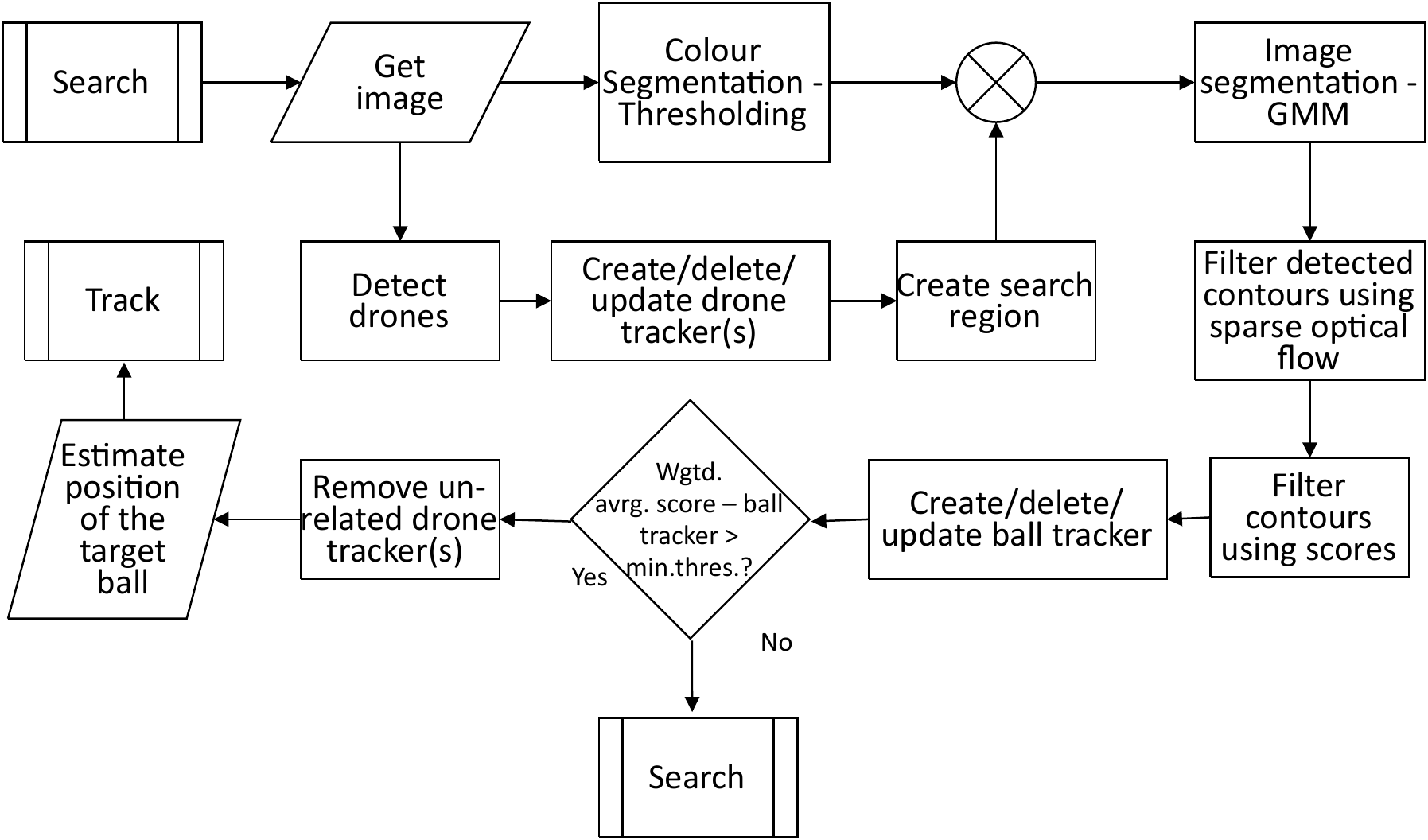}
     \caption{Schema of search phase in ball detection and tracking.}
     \label{fig:search}
 \end{figure}
\begin{enumerate}[label=(\alph*)]
    \item \textbf{Search Phase:}
    The main objective of the search phase is to locate the target ball suspended from a high-speed UAV. The first step is to identify the target UAV, followed by the target ball. Once the UAVs are detected and tracked, their position information is used to search for the ball attached to the target UAV.
    
    The image is segmented based on the color of the ball, and a vertical region below the bounding boxes of the UAVs in the segmented image is used to search for the target ball attached to the target UAV. A multivariate Gaussian Mixture is used to model the density of the target ball in the images to perform a likelihood-based segmentation \cite{doi:10.1002/rob.21816}, and the Expectation Maximization (EM) algorithm is used to estimate the parameters of the multi-modal distribution from the pixel values of the target ball from training data. In outdoor environments, varying lighting conditions can greatly affect the appearance of the objects and color segmentation in an RGB image. Whereas, HSV color space is robust to lighting conditions as it separates the color information and image intensity information. A reliable segmentation is obtained by using the likelihood of the pixel values from the HSV image. 
     Additionally, sparse optical flow calculated for multiple points inside the segmented contours over successive frames by using the Lucas-Kanade method is used to filter out other distracter contours in the search region. Assuming the target UAV is maneuvering at a higher speed, the time of the flow can be used to eliminate some of the false positives. The objects corresponding to the selected contours in the search region are assigned scores based on their geometry and location relative to the corresponding UAV’s bounding boxes in the following manner:
\begin{equation}\label{eq:circularity}
   \text{Circularity} = \dfrac{4\pi \times \text{Area}}{\text{Perimeter}^2} \\
\end{equation}
\begin{equation}\label{eq:score}
    \text{Score} = 
    \begin{cases}
      \text{Circularity}_{\text{contour}}, & \lvert l_r - \dfrac{L_r \times r_b}{R_b} \rvert \leq \epsilon\\
      -1 & \text{otherwise}
    \end{cases}       
\end{equation}
where, $l_r$ is the distance from the bottom of the UAV to the center of the selected contour and $r_b$ is the radius of the smallest circle enclosing the selected contour; $R_b$ is the actual radius of the target ball, and $L_r$ is the length of the rod used to suspend the ball.

Circularity metric given by (\ref{eq:circularity}) is used in the evaluation of the score (\ref{eq:score}). Ideally, the circularity can range between 0 to 1, with 1 for a perfectly circular contour. However, circularity might be greater than one based on the approximations used in measuring the area and perimeter of the contour in digital images. The score is essentially the circularity of the contour, as long as the observed distance of the ball from the UAV in the image is close to the distance predicted based on the size of the contour. To evaluate the suitability of the contour in consideration for the target ball, the distance of the contour from the UAV in the image $l_r$ is compared with the expected length of the rod in the image for the size of the contour in the image $r_b$ calculated using $L_r$ and $R_b$; $\epsilon$ is the tolerance on the difference between $l_r$ and the expected length of the rod for the images. It is manually tuned by experimentation as the shape of the contours can vary, and the swing of the ball can depend on several factors like the joint connecting UAV and rod, wind conditions, and the motion pattern of the UAV carrying the ball. The scores of the contours corresponding to the detected colored objects are used as a measure of the likelihood of the colored object being the target ball. Once the target ball is identified, perspective projection is employed to compute the position of the target ball using the camera intrinsics and the dimensions of the ball. If a suitable candidate could not be identified, the search for the target ball continues. The block schematic of the search phase is given in Fig. \ref{fig:search}. The algorithm detects the ball as shown in Fig. \ref{fig:ball_explained}.

\item \textbf{Tracking Phase:}
The Kalman filter trackers for the target ball and UAV in the image plane are updated throughout this stage, and the trackers for the other UAVs are removed. The target UAV and ball trackers are used to reduce the region of interest to locate the target ball in the subsequent frames. A wedge-shaped region below the UAV is used for this purpose, the location of the wedge from the UAV, and the size of the wedge is determined using the depth information of the target ball from the previous frame. As the dimensions of the ball and the rod used to suspend it are known, the depth information of the target ball from the previous frame is used to roughly estimate the length of the rod in the image plane for the current frame, and this information is used to position the wedge shaped search region accordingly. Similarly, the radius of the target ball from the previous frame is used to determine the size of the wedge-shaped search region.

\item \textbf{Grab Phase:}
In the terminal phase, the ball must be tracked precisely without any misses to attempt to grab the ball. As the interceptor UAV approaches the target ball, the target UAV moves out of the field of view of the camera, and the square-shaped search region around the ball’s predicted position is used to track the ball in this phase.
\end{enumerate}
\begin{figure}[t]
     \centering 
     \includegraphics[scale=0.5]{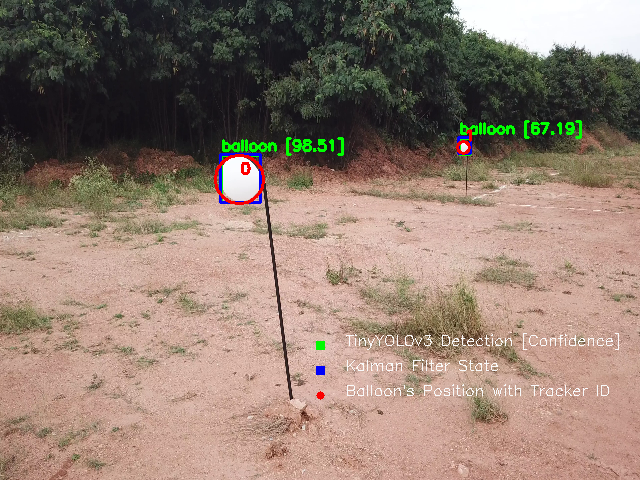}
     \caption{Balloon detection and tracking. The confidence level of each detection is shown along with the bounding box.}
     \label{fig:balloon_explained}
 \end{figure}
 
\subsubsection{Balloon detection}
\begin{figure}
     \centering
     \includegraphics[scale=0.65]{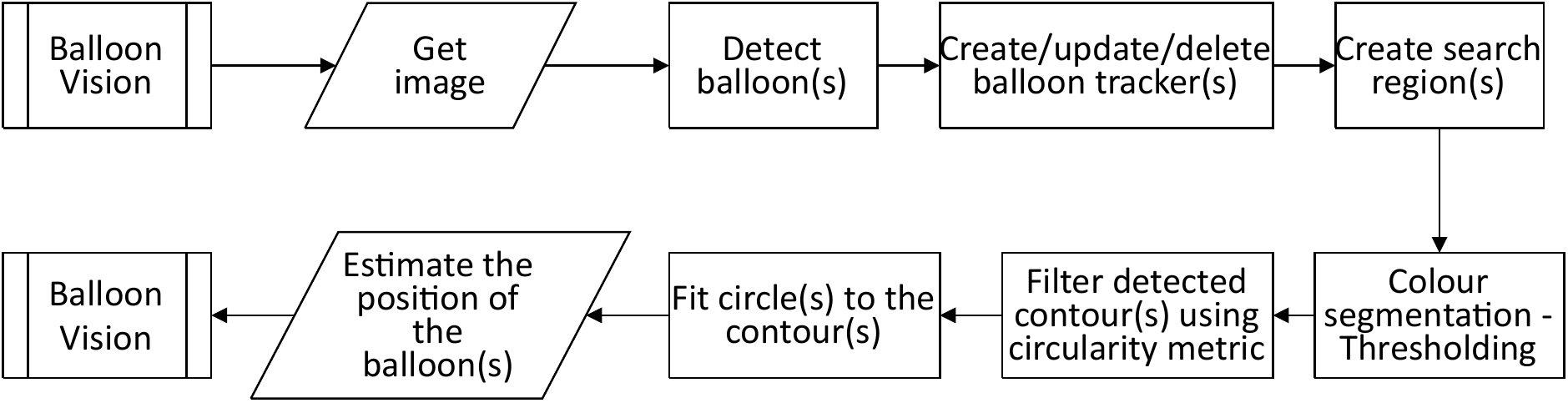}
     \caption{Schematic of balloon detection and tracking algorithm.}
     \label{fig:balloonvision}
 \end{figure}

The balloon popping UAV detects and tracks all the balloons in the image, as illustrated in Fig. \ref{fig:balloon_explained}. 
Once a balloon is detected, a new instance of the tracker with a unique ID is initialised to maintain the position of the balloon. 
Then, the regions of the bounding boxes corresponding to the tracked balloons are segmented based the color of the balloon, and the resulting blobs (that potentially represent the balloons) are filtered based on circularity. The center and major axis of each blob are computed, and the circle that passes through the intersection points of the major axis and the contour of the blob is used to estimate the positions of the balloons. This computation assumes that the balloons are spherical in shape, uniform in size, and the diameter is known. The block schematic of balloon detection and tracking is shown in Fig. \ref{fig:balloonvision}.

\subsubsection{Use of data sets}
TinyYOLOv3 trained on PASCAL VOC 2012 data set is re-purposed as a UAV/balloon detector by fine-tuning the last four convolutional layers of the network. A combination of manually annotated data and a synthetically generated dataset is used to re-train the network. A large portion of the data set is synthetically generated by blending the target images (balloon and UAV images) collected at different lighting conditions and orientations, with a diverse variety of backgrounds at different sizes and orientations, as described in \cite{Dwibedi_2017_ICCV}. To extract target images for synthetic data generation, Pixel Objectness \cite{pixelobjectness}, a deep neural network for semantic segmentation is used to segment the targets from the background. Also, many negative samples without any annotations are included in the dataset, as this improves the mean average precision for 0.5 IoU threshold (mAP@0.5) by 20\% . The inclusion of synthetically generated images in the data set used for training the network leads to a 20\% improvement in mAP@0.9. From the results, it can be observed that the inclusion of many uniformly generated annotations of the synthetic data set improves the confidence score and IoU of the detections. 

\subsection{Target tracking }
\begin{figure}[htb!]
    \centering
    \includegraphics[scale=0.75]{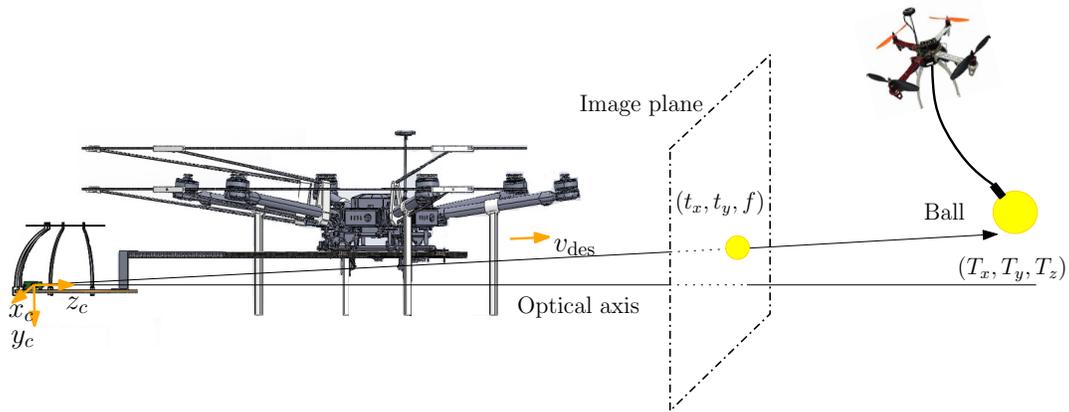}  
    \caption{Engagement geometry: The location of target, image plane and manipulator.}
    \label{fig:ball_cam}
\end{figure}

A target tracking algorithm is developed using the information from the monocular camera by combining a vision based guidance law with a distance-based attractive-repulsive potential function. The guidance law keeps the target in the field of view, whereas the attractive-repulsive potential function helps in maintaining the distance from the target. A typical engagement geometry between target and tracker/grabber UAV is shown in Fig. \ref{fig:ball_cam}. In the camera frame, the projected coordinates of the ball in image plane is ($t_{x}$, $t_{y}$, $f$).
\begin{figure}[htb!]
\begin{subfigure}{.33\textwidth}
  \centering
  \includegraphics[scale=0.305]{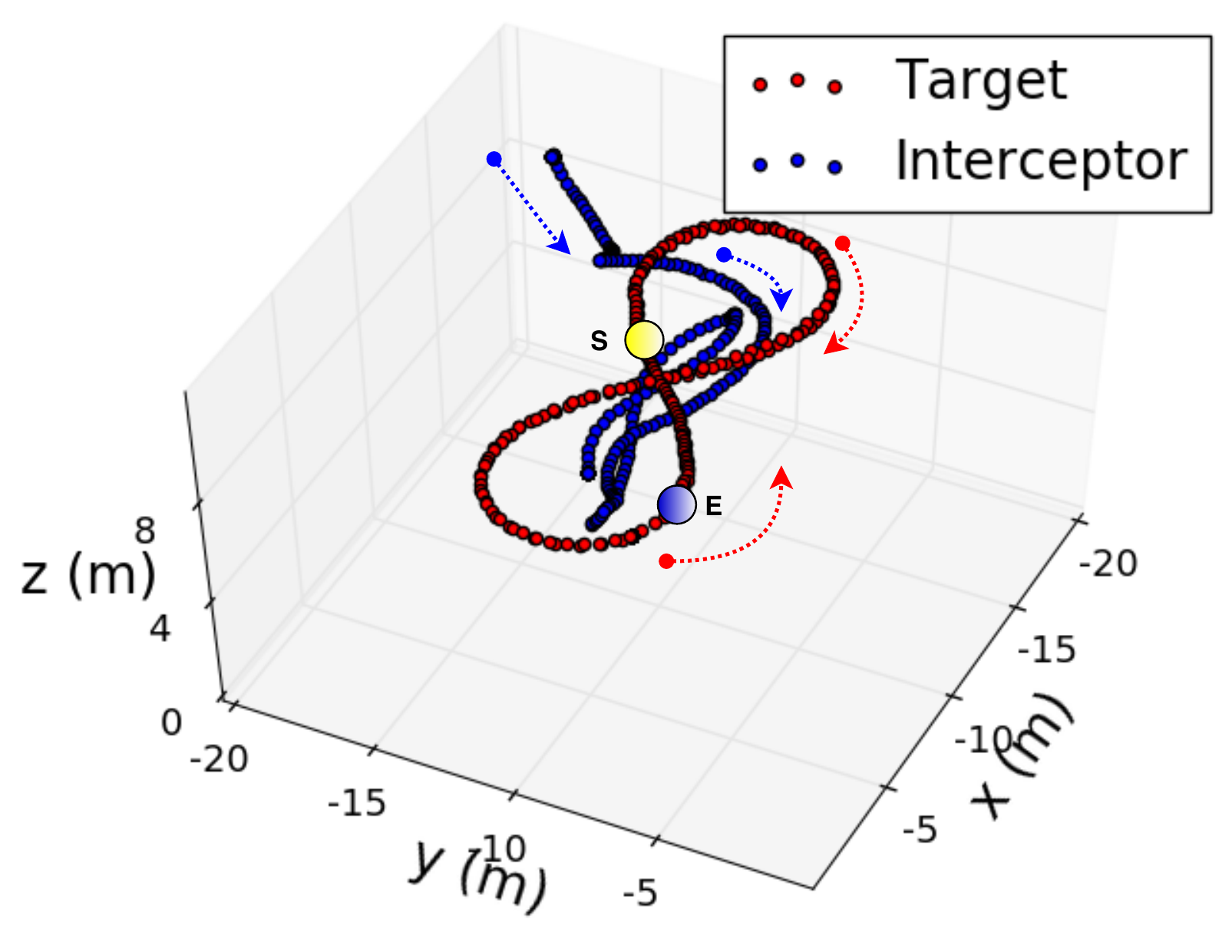}  
  \caption{}
  \label{fig:tracking_full_trace}
\end{subfigure}
\begin{subfigure}{.33\textwidth}
  \centering
  \includegraphics[scale=0.275]{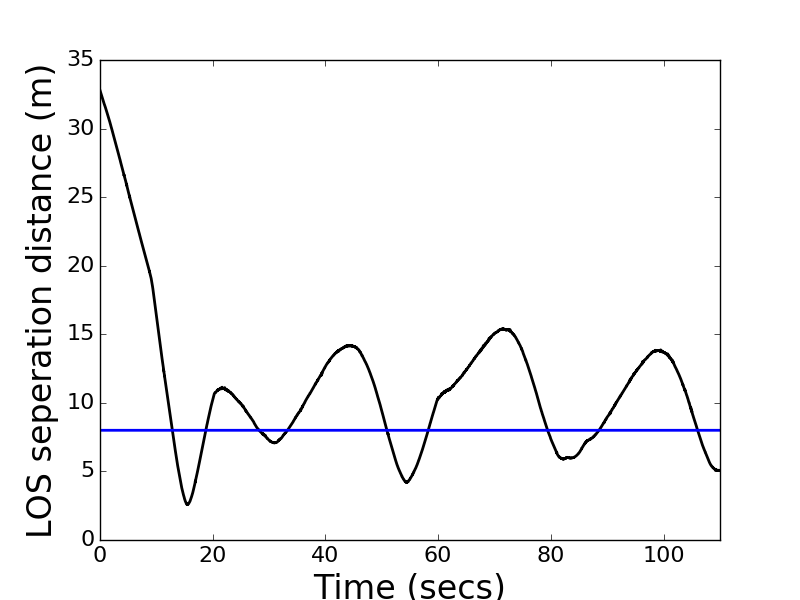}
  \caption{}
  \label{fig:tracking_distance_plot}
\end{subfigure}
\begin{subfigure}{.33\textwidth}
  \centering
  \includegraphics[scale=0.275]{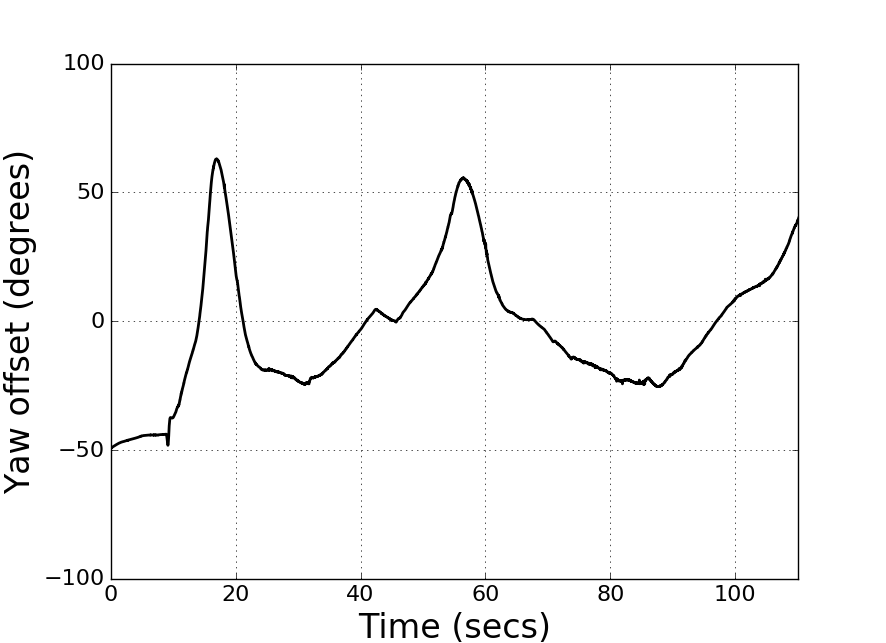}  
  \caption{}
  \label{fig:tracking_yaw_error}
\end{subfigure}
\caption{Target tracking (a) Trajectory (b) Distance between the target and interceptor (c) Yaw offset during tracking.}
\label{fig:plots_tracking_solo}
\end{figure}   
\begin{figure}[htb!]
\begin{subfigure}{.49\textwidth}
  \centering
  \includegraphics[scale=0.32]{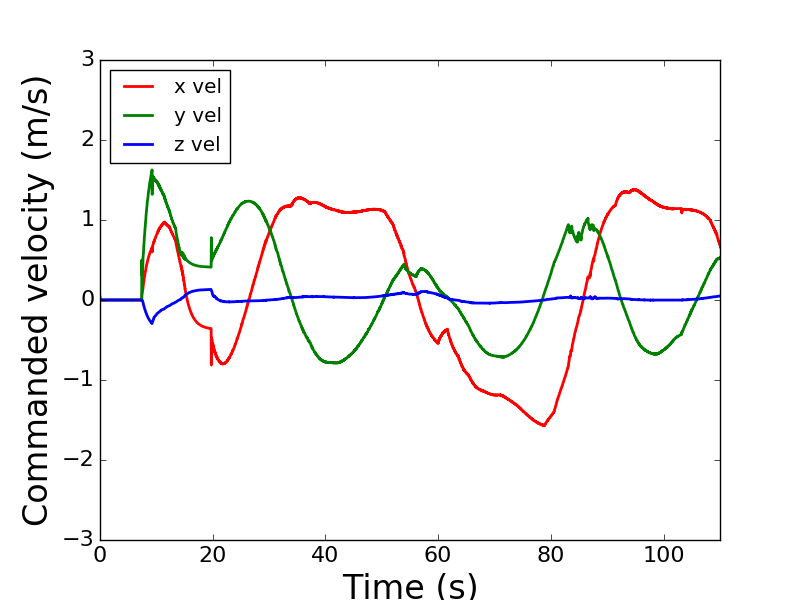}  
  \caption{}
  \label{fig:tracking_solo_cmd_vel_plot}
\end{subfigure}
\begin{subfigure}{.49\textwidth}
  \centering
  \includegraphics[scale=0.32]{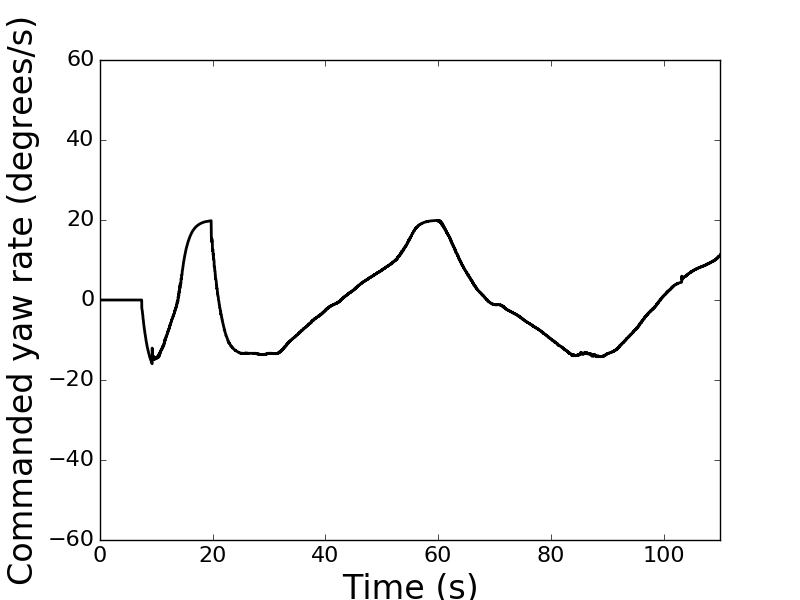}
  \caption{}
  \label{fig:tracking_solo_cmd_yaw_plot}
\end{subfigure}
\caption{(a) Commanded velocity in $x$, $y$ and $z$ direction (b) Commanded yaw rate.}
\label{fig:plots_tracking_solo_cmd}
\end{figure}
The tracker UAV is made to move parallel to the line joining the camera focal point and the pixel position of the target's center in the image plane so that it can follow the target point. If the UAV is inside the tracking radius of the target, then the UAV is subjected to velocity away from the target and \textit{vice-versa}.  Apart from the commanded velocities, the UAV is also subjected to the yaw rate command so that the target is always maintained within the field of view of the camera attached to the manipulator. Algorithm \ref{alg:tracking} shows the details of the tracking process.  A tracking scenario is simulated in the Gazebo to track a target moving in a figure-of-eight loop, where the tracking radius is 8 m. Figure \ref{fig:tracking_full_trace} shows the trajectories of the target and the tracking UAV. The separation between the target and tracking UAV is plotted in Fig. \ref{fig:tracking_distance_plot} which shows that the tracking UAV remains 8 m away from the target UAV. The oscillatory response is due to the tracker UAV being pushed away by the potential filed force when closing in to the target. Figure \ref{fig:tracking_yaw_error} shows the error between desired yaw and actual yaw to keep the target in the field of view of the camera attached to the interceptor.  The commanded values of position and heading are shown in Fig. \ref{fig:plots_tracking_solo_cmd}.

\begin{algorithm}[htb!]\label{}
\textbf{Input}: Tracking distance $(D_{\text{track}})$,  
target pixel coordinates ($t_{x}$, $t_{y}$), camera focal length ($f$),  image  width ($w$), image height ($h$), yaw attitude ($\psi$), camera frame to body frame transformation matrix ($R_{c2b}$), body frame to vehicle frame transformation matrix ($R_{b2i}$). 

\begin{enumerate} 
     \item Calculate the target pixel-coordinates in the camera frame \\
          $p_{x}\gets t_{x}-w/2$,~~~
          $p_{y}\gets t_{y}-h/2$
     \item Calculate  the distance of ball $(D_{\text{target}})$ from the camera center using the prior information of the size of the ball
     
     \item Formulate attractive-repulsive potential function considering $(D_{\text{track}})$ as the equilibrium point 
     
     \item  Calculate  the magnitude of velocity $(V_{\text{track}})$ required from the potential function  using the distance from the target
     
     \item  Calculate the unit vector along the line joining the camera center and center of the ball in the image plane, $O^{c} = (l_{x}, l_{y}, l_{z}$)'
      
        $ l_{x}\gets   \frac{p_{x}}{\sqrt{p_{x}^{2}+p_{y}^2+f^{2}}}$,~~~
        $ l_{y} \gets  \frac{p_{y}}{\sqrt{p_{x}^{2}+p_{y}^2+f^{2}}}$,~~~
        $ l_{z} \gets  \frac{f}{\sqrt{p_{x}^{2}+p_{y}^2+f^{2}}}$ 
     \item   Calculate the desired velocity in the camera frame
    
     \begin{algorithmic}
      \IF{$D_{\text{target}}$ 	$\geq$  $D_{\text{track}}$}
      \STATE $V_{c} \gets V_{\text{track}} O^{c} $
     \ELSE 
     \STATE $V_{c} \gets -V_{\text{track}} O^{c} $
     \ENDIF
    \end{algorithmic}
  
     \item Calculate the transformation from the camera frame to the vehicle frame: $ R_{c2i} \gets  R_{b2i} R_{c2b}$
     
     \item Desired velocity in the vehicle frame:
         $V_{\text{des}} \gets  R_{c2i} V_{c} $
 
      \item Calculate the $O^{c}$ in the vehicle frame: $O^{v} \gets  R_{c2i}$  $O^{c}$
 
     \item Calculate desired yaw: $\psi_{\text{des}} \gets \arctan(\frac{O^{v}(2)}{O^{v}(1)}$)
    \item Calculate yaw offset: $e_{\psi} \gets \psi_{\text{des}}-\psi$.
    \item Calculate desired yaw rate: $r_{\text{des}} \gets kp_{\psi} e_{\psi} +  kd_{\psi} \frac{\text{d}e_{\psi}}{\text{dt}}$
\end{enumerate}
\textbf{Output}: Desired  velocity ($V_{\text{des}}$), desired yaw rate ($r_{\text{des}}$).
 \caption{Ball tracking}
 \label{alg:tracking}
\end{algorithm}

\subsubsection{Computational complexity}
Consider that $i$th step in the algorithm takes $T_i$ units of time to execute. There are 12 steps in this algorithm. The total time involved is the summation of individual steps involved.
\begin{equation}
    T_{\text{Total}}=T_1+T_2+T_3+\cdots+T_{12}
\end{equation}
where, $T_{\text{Total}}$ is the total execution time of the algorithm. Thus, for a given camera and its  parameters, object detection algorithm, and vehicle state parameters, the run-time computational complexity is a constant. At every time step, a constant run time is required for tracking the ball, which makes it feasible to implement in real time.

\subsection{Guidance and interception/ball grabbing}
In this section, a guidance law is designed for ball grabbing and balloon popping. The method is inspired by the missile guidance literature, where target interception is the focus. The design of the gripper of the manipulator is also arrived at based on the proposed guidance law.

\subsubsection{Guidance for maneuvering ball grabbing}
Various guidance laws such as different forms of Proportional Navigation (PN) are reported in the literature for interception of a target using radar/seeker information \cite{zarchan2012tactical}, \cite{siouris2004missile}. However, it is difficult to extend the same algorithms using visual sensors because of the delay and uncertainty involved in visual feedback. In missile interception problems, the expected miss distance in the terminal phase is accommodated by considering the suitable lethal radius of the attached warhead. In our case, the allowable miss distance for balloon interception or ball grabbing is very low, considering the size of the manipulator to satisfy the overall size constraints. Any guidance laws involving the direct or indirect target depth information will have a large miss distance because of the huge uncertainty involved in the estimation of depth. The miss distance will further increase if the depth information is obtained from a monocular camera.  In the case of the interception of a maneuvering target, predictive guidance laws will also have large miss distance due to uncertainty involved in predicted target position from the visual information. 
   
Considering the above facts, a visual guidance law is proposed based on the idea of the classical pure pursuit guidance, where the interceptor moves towards the target along the line of sight between the camera center and ball center without including the target depth information. As shown in Fig. \ref{fig:ball_cam},  the grabber UAV is subjected to the velocity parallel to the line joining the camera centre and the projection of centre of ball ($T_{x}$, $T_{y}$, $T_{z}$) on the image plane ($t_{x}$, $t_{y}$, $f$).  The grabber UAV is also subjected to a commanded yaw rate to keep the ball in the center of the field of view of the camera.  The details of the guidance algorithm are presented in Algorithm \ref{alg:guidance}. Since the target is moving at low speed, the turn radius required at the last phase of interception is within the allowable limit of the interceptor UAV. 

\begin{algorithm} [t]
\textbf{Input}: Target pixel coordinates ($t_{x}$, $t_{y}$), camera focal length ($f$), magnitude of target velocity ($V_{\text{t}}$), excess velocity ($V_{\text{excess}}$), image  width ($w$), image height ($h$), yaw attitude ($\psi$), camera frame to body frame transformation matrix ($R_{c2b}$), body frame to vehicle frame transformation matrix ($R_{b2i}$)). 
\begin{enumerate} 
    \item Calculate the target pixel-coordinates in the camera frame \\
    $p_{x}\gets t_{x}-w/2$,~~~
    $p_{y}\gets t_{y}-h/2$
    \item  Calculate the unit vector along the line joining the camera center and center of ball in image plane\\
    $O^{c}\gets \frac{1}{\sqrt{p_{x}^{2}+p_{y}^2+f^{2}}} (p_{x}, p_{y}, f)$
    \item  Desired velocity magnitude for interception in camera frame: $ V_{\text{mag}}\gets V_{t} +V_{\text{excess}}$
    \item Desired velocity for interception in camera frame: $ V_{c}\gets V_{\text{mag}} O^{c} $ 
    \item Calculate the desired velocity components in camera frame: $V_{c}=(V_{cx}, V_{cy}, V_{cz})' $ \\
    $V_{cx}\gets V \frac{p_{x}}{\sqrt{p_{x}^{2}+p_{y}^2+f^{2}}}$,~~
    $ V_{cy} \gets V \frac{p_{y}}{\sqrt{p_{x}^{2}+p_{y}^2+f^{2}}}$,~~
    $ V_{cz} \gets V \frac{f}{\sqrt{p_{x}^{2}+p_{y}^2+f^{2}}}$ 
    \item Calculate the transformation from camera frame to vehicle frame: $ R_{c2i} \gets  R_{b2i} R_{c2b}$
    \item Desired velocity in vehicle frame:
    $ V_{\text{des}} \gets  R_{c2i} V_{c}  $
     
     \item  Calculate the desired yaw rate $r_{\text{des}}$ as described in Algorithm \ref{alg:tracking} 
\end{enumerate}
\textbf{Output}: Desired  velocity ($V_{\text{des}}$), desired yaw rate ($r_{\text{des}}$).
\caption{Ball grabbing}
\label{alg:guidance}
\end{algorithm}

\begin{figure}[t]
\begin{subfigure}{.33\textwidth}
    \centering
    \includegraphics[scale=0.31]{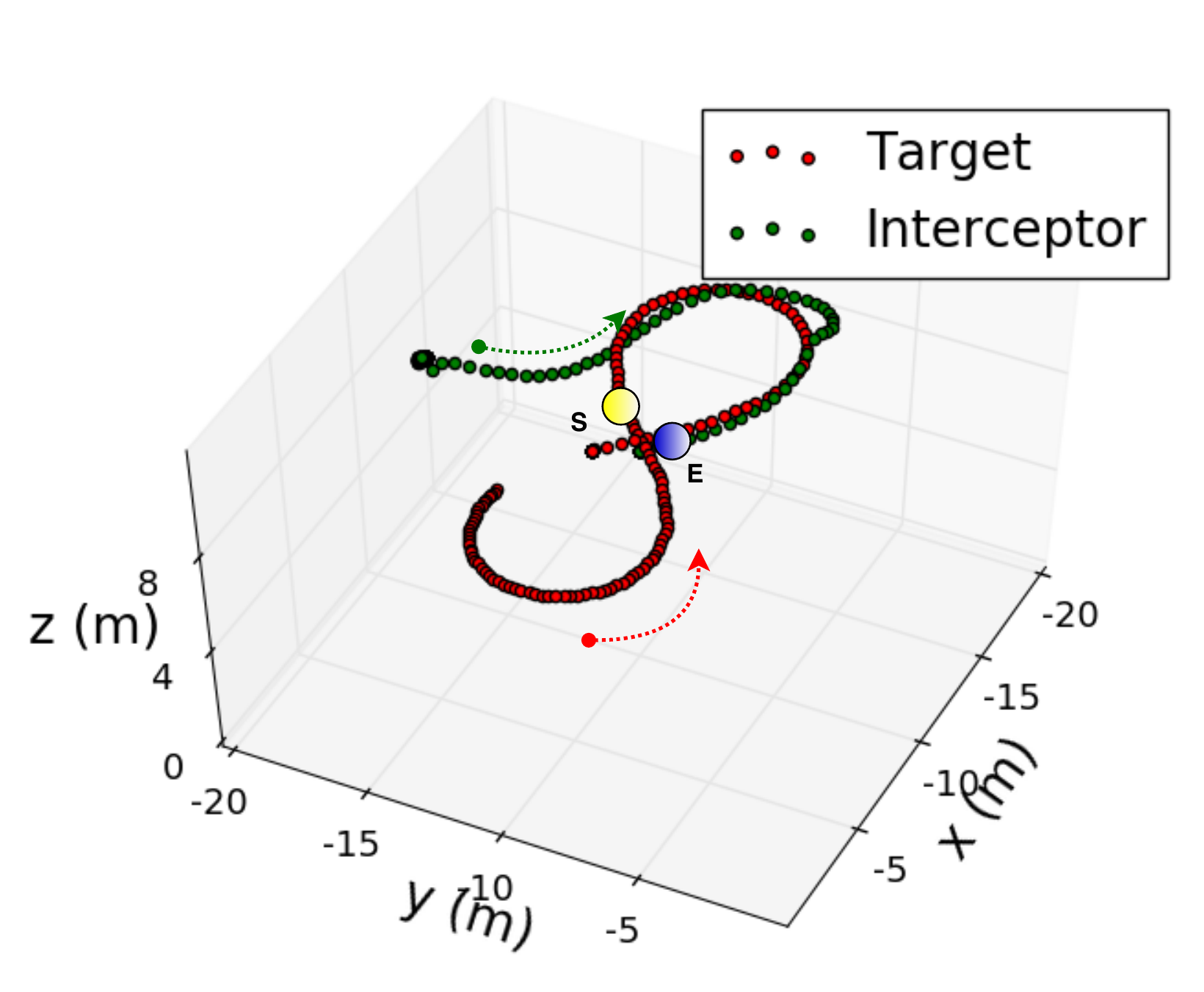}  
    \caption{}
    \label{fig:grabbing_trace_full}
\end{subfigure}
\begin{subfigure}{.33\textwidth}
    \centering
    \includegraphics[scale=0.27]{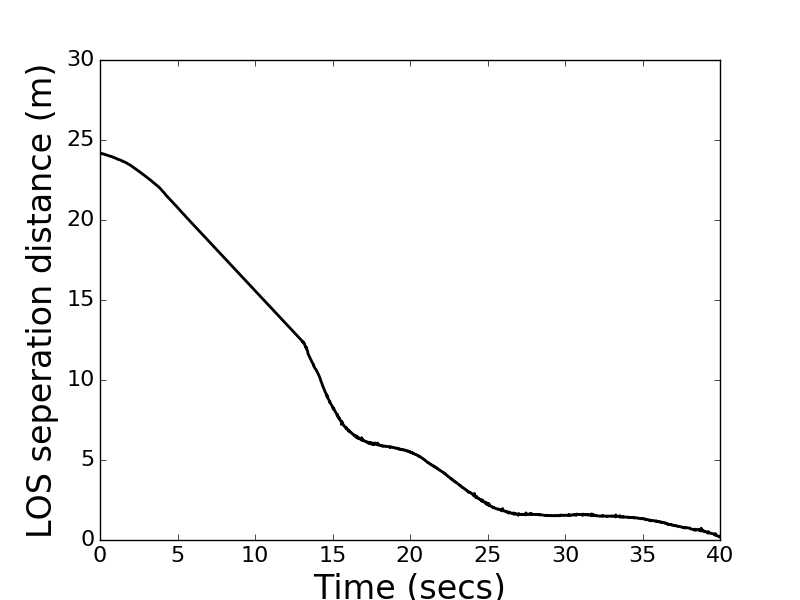}  
    \caption{}
    \label{fig:grabbing_distance}
\end{subfigure}
\begin{subfigure}{.33\textwidth}
    \centering
    \includegraphics[scale=0.27]{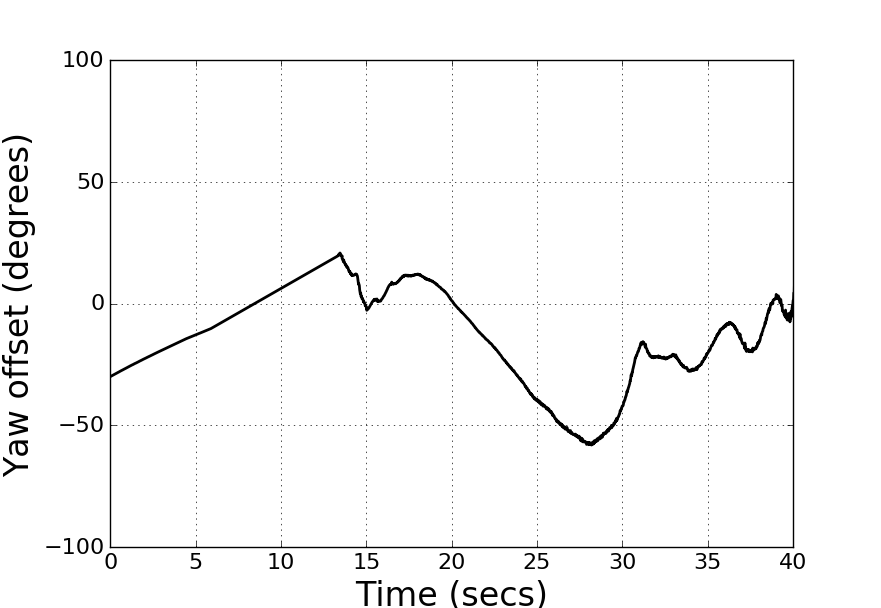}
    \caption{}
    \label{fig:grabbing_yaw_error}
\end{subfigure}
\caption{(a) UAV trajectories (b) Variation of LOS separation (c) Yaw offset during grabbing. The LOS distance converges to zero on grabbing.}
\label{fig:plots_grabbing_solo}
\end{figure}

\begin{figure}[t]
\begin{subfigure}{.49\textwidth}
  \centering
  \includegraphics[scale=0.32]{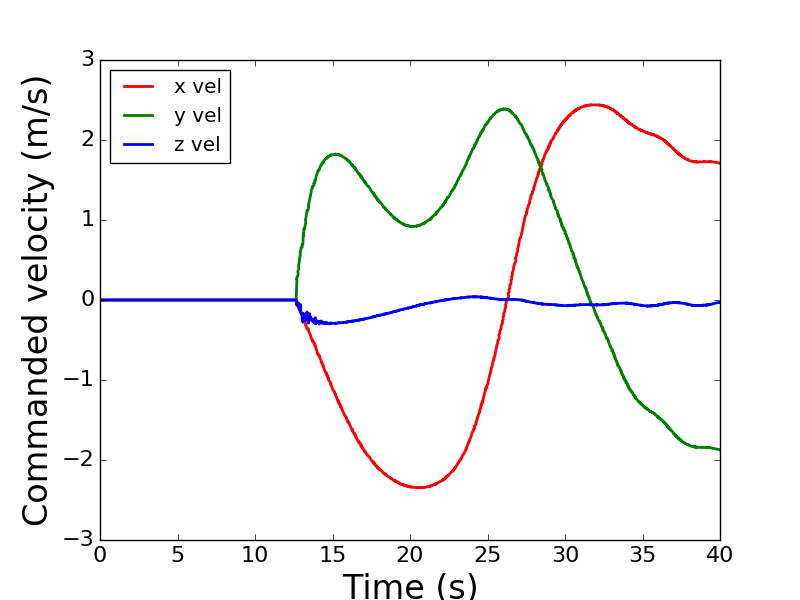} 
  \caption{}
  \label{fig:grabbing_solo_cmd_vel_plot}
\end{subfigure}
\begin{subfigure}{.49\textwidth}
  \centering
  \includegraphics[scale=0.32]{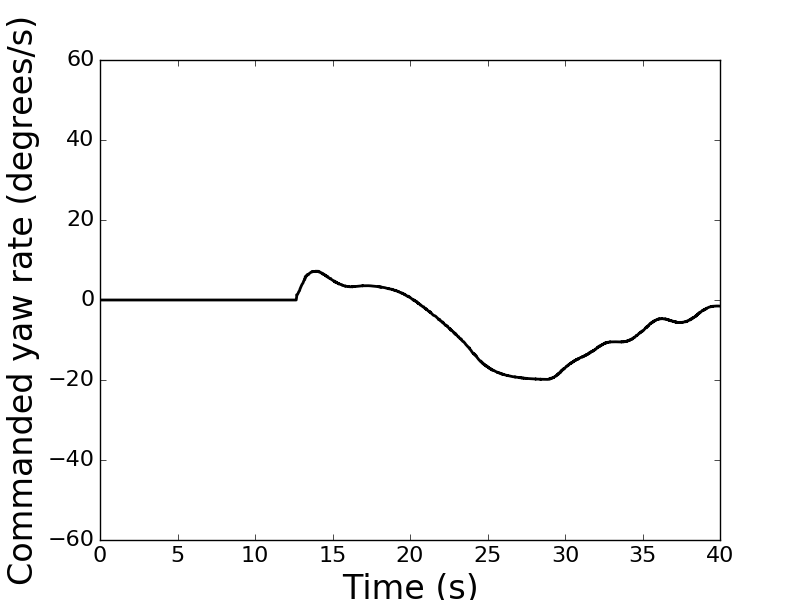}
  \caption{}
  \label{fig:grabbing_solo_cmd_yaw_plot}
\end{subfigure}
\caption{(a) Commanded velocity in $x$, $y$ and $z$ direction (b) Commanded yaw rate.}
\label{fig:plots_grabbing_solo_cmd}
\end{figure}

A typical engagement scenario is generated to validate the grabbing algorithm in Gazebo. Figure \ref{fig:grabbing_trace_full} shows the target and interceptor path for an engagement scenario where, the target is detected for the first time at point S, and the engagement ends at point E. Target and interceptor speed is considered as 2.0  m/s and 2.5 m/s, respectively. The variation of Line of Sight (LOS) separation distance between the camera center and the target during the grabbing process is plotted in Fig. \ref{fig:grabbing_distance}. Clearly, the distance goes to zero during grabbing. The error in the desired yaw to keep the target in the center of the field of view and the actual yaw is shown in  Fig. \ref{fig:grabbing_yaw_error}.  The commanded values of position and heading are shown in Fig. \ref{fig:plots_grabbing_solo_cmd}.

The effective radius of the gripper of the grabbing drone's manipulator is designed based on the form of the guidance law. It was observed during the initial flight tests that the interceptor UAV was unable to perform quick maneuvers when the ball depth was less than 0.5 m. This was attributed to the computational delay associated with the vision module. To resolve this, the algorithm for ball grabbing was designed so that the end effector of the interceptor UAV reached feasible relative positions for successful capture before it reached 0.5 m to the ball.  An analysis of relative velocity space between the target and the UAV was performed considering the target velocity constant at the terminal phase. 

\begin{figure}[t]
    \centering
    \includegraphics[scale=0.65]{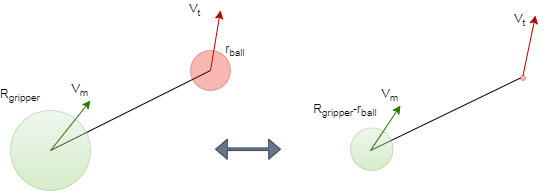}
    \caption{Ball grabber gripper design: The virtual sphere is constructed to compute the grabber size.}
    \label{fig:gripper1}
\end{figure}

Consider an engagement scenario, as shown in Fig. \ref{fig:gripper1}, where the ball with radius $r$ is moving with constant velocity  $V_t$  and a gripper having effective grabbing radius of $R_{\text{gripper}}$, moves at a speed of $V_{m}$, pursuing the ball. The grabbing can be expressed as an equivalent scenario of interception of a virtual sphere with radius $R_{eq}$, where the target is assumed to be a point.  For successful grabbing of the ball, the equivalent grabbing radius is the difference between the gripper's radius and the radius of the ball. 
\begin{equation} \label{eqrad}
    R_{eq}= R_{\text{gripper}} - R_{\text{ball}}
\end{equation}
Interception is possible if the point target lies in the collision cone of the  virtual sphere \cite{chakravarthy2012generalization}. Therefore, the radius of virtual sphere should satisfy the following condition.
\begin{equation} 
    R_{eq}^{2} \geq \frac{r_{0}^{2}(V_{\theta 0}^{2}+ V_{\phi 0}^{2})}{(V_{\theta 0}^{2}+ V_{\phi 0}^{2}+V_{r0}^{2})}  \label{virtualrad}
\end{equation}
where, $r_0$ is the initial  distance, $V_{r0}$ is the component of the relative velocity along the  line of sight, $V_{\theta 0}$, $V_{\phi 0}$ are the relative velocity components normal to the line of sight. When the ball is 0.5 m away from the manipulator, it is desired that the guidance law can position the interceptor UAV such that the error in alignment of target velocity and interceptor velocity is within 5$^\text{o}$. Considering an approximate pure pursuit engagement in the 2 D plane, a head-on scenario is considered with a target speed of 8 m/s and an interceptor speed of 2 m/s. The  value of $R_{eq}$ should be greater than 34.9 mm, as obtained from the \eqref{virtualrad}, where, the values of  $r_{0}$, $ V_{r0}$, $V_{\theta 0}$, $V_{\phi 0}$ are 0.5 m,  9.96 m/s, 0.69 m/s and 0 m/s, respectively.  A similar analysis for a tail-chase condition with a target speed of 4 m/s and interceptor speed of 6 m/s gives the minimum value of $R_{eq}$ to be 85 mm. The size of the target ball is 100 mm.  As per \eqref{eqrad},  the effective radius of the gripper should be greater than 185 mm. Thus, the design value of the effective radius of the gripper is chosen as 200 mm.       

\subsubsection{Computational complexity}
We follow the same procedure as in the previous section to evaluate the run time complexity of the ball grabbing algorithm. The total run time of the algorithm is computed as below. 
\begin{equation}
    T_{\text{Total}}=T_{1}+T_{2}+\cdots+T_{7}+T_{8}
\end{equation}
Step 8 constitutes three computations. Thus, the run-time complexity of ball grabbing algorithm is also a constant, making it a fit candidate for real-time implementation.

\subsubsection{Interception of high speed target}
The interception of a high-speed target requires the prediction of the target points with reasonable accuracy.  In our case, it is known a priori that the target will be maneuvering through a repetitive trajectory and a ball with known color will be attached to the target at a certain distance below the target.  Using these information of the type of target path and the color of the object, an estimation framework using EKF is proposed to estimate the position of the target. The information of the repetitive trajectory of the target is considered in the motion model, and the measurement model is developed using the visual observations of the target points. 
 \begin{figure}
     \centering
     \includegraphics[scale=0.5]{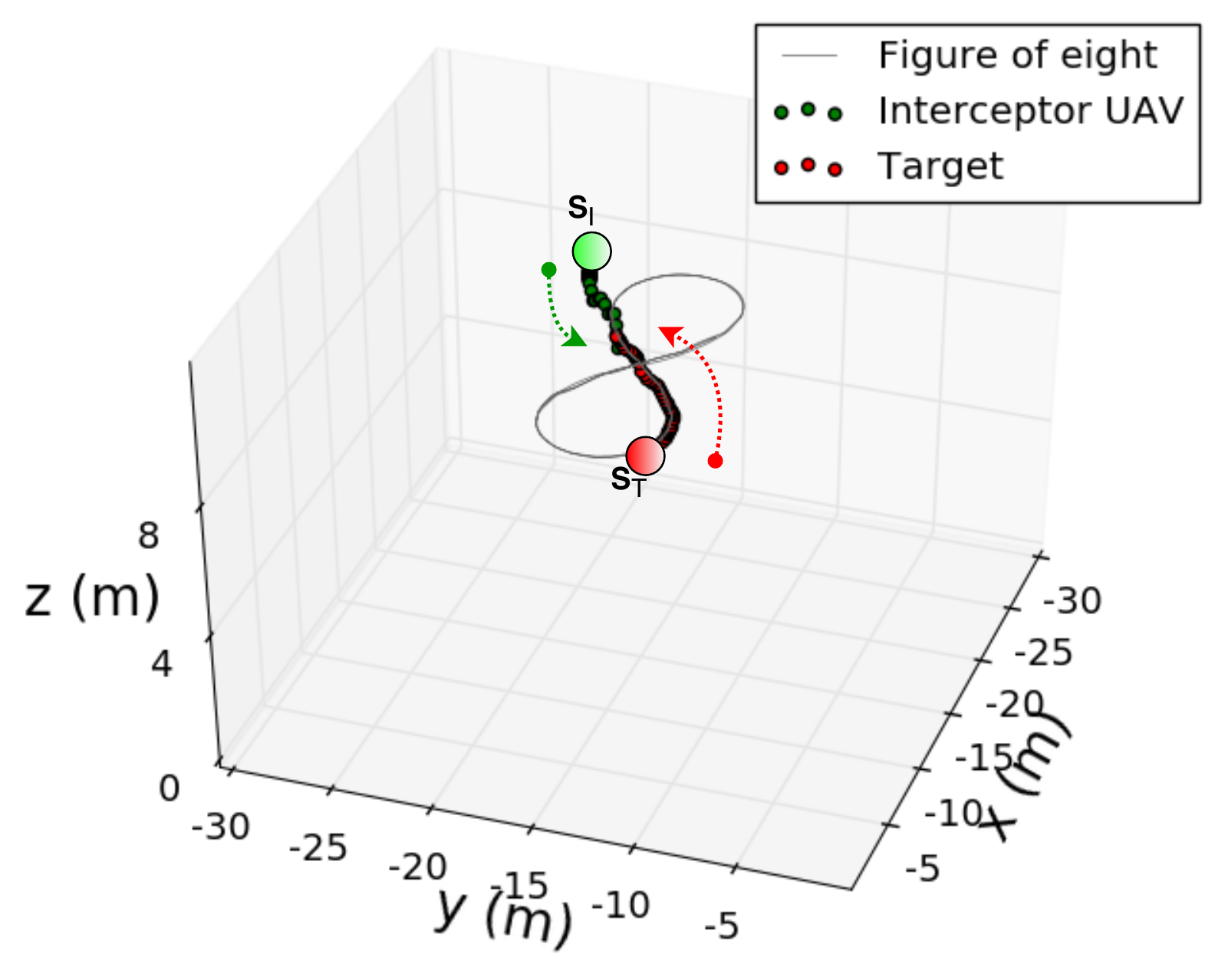}
     \caption{Engagement with high speed target.}
     \label{fig:high_speed}
 \end{figure}
\begin{algorithm}[htb!] \label{}
\textbf{Input}: Target pixel coordinates in camera frame $(p_{x}, p_{y})$,  focal length ($f$), target depth ($Z$), magnitude of target velocity $(V_{T})$.
\begin{enumerate} 
   \item  Develop a EKF using the target position as state vectors
          \begin{enumerate}
          \item  Calculate the approximate centre of curvature $(T_{x0}, T_{y0})$ of path of target
         \item Consider the  state dynamics:  \\ 
          $\dot{T_{x}}\gets -V_{T} \frac{T_{x}-T_{x0}}{\sqrt{(T_{x}-T_{x0})^{2}+(T_{y}-T_{y0})^{2} }} $,~~~
          $\dot{T_{y}}\gets -V_{T} \frac{T_{y}-T_{y0}}{\sqrt{(T_{x}-T_{x0})^{2}+(T_{y}-T_{y0})^{2} }} $ 
          \item Measure the   states vectors using the target centre pixel co-ordinates and target depth ($Z$):  \\
         $ T_{x} \gets  \frac{p_{x} Z}{f} $,~~~
         $ T_{y} \gets  \frac{p_{y} Z}{f} $ 
         \item  Estimate the   current position of  target $(\hat{T}_{x}, \hat{T}_{y})$ with the EKF formulation 
      \end{enumerate}
      \item Predict the incremental movement  of target $(\delta _{T_{x}},\delta _{T_{y}})$ at future  $m$ steps from the previous $n$  sequence of observations
     \item  Predict the target location $(\hat{T}_{mx}, \hat{T}_{mx})$:
      $ \hat{T}_{mx}= \hat{T}_{x} +  \delta _{Tx} $,~~~
      $ \hat{T}_{my}= \hat{T}_{y} +  \delta _{Ty} $
      \item  Estimate the average height of target $(\hat{T}_{z})$
       \item   Estimate the trajectory of the path by fitting  best curve after minimizing the least square errors
       \item   Estimate suitable  stand-off point $(S_{x},S_{y}, S_{z})$  near to target path for grabbing  
\end{enumerate}
\textbf{Output}: Target current position $(\hat{T}_{x}, \hat{T}_{y},\hat{T}_{z})$, predicted position $(\hat{T}_{mx},\hat{T}_{my},\hat{T}_{z})$, grabbing stand-off point $(S_{x},S_{y}, S_{z})$.
 \caption{ Target position estimation, prediction and grabbing stand-off point estimation}
 \label{alg:prediction}
\end{algorithm}  
After the estimation of target points over a duration, the future position of the target points are predicted from the previous sequence of observations. As the target is performing a figure-of-eight loop, a curve fitting method is used to estimate its trajectories based on the observed points. After estimation of the target path, the suitable grabbing stand-off points are selected such that the interceptor UAV can perform approximate head-on interception with the target. Grabbing stand-off points are those feasible locations with respect to the target trajectory where, the UAVs should wait to capture the ball. A typical interception scenario of high speed target is shown in Fig. \ref{fig:high_speed}. The grabber UAV waits at the stand-off points and follows the target to capture the ball.

\subsubsection{Computational complexity}
The total time for execution of Algorithm \ref{alg:prediction} is given below.
\begin{equation}
    T_{\text{Total}}=T_{1a}+T_{1b}+T_{1c}+T_{1d}+T_{2}+T_{3}+T_{4}+T_{5}+T_{6}
\end{equation}
where, $T_{1a},~T_{1b},~T_{1c},~T_{1d}$ denote the time involved in the steps 1a-1d, respectively. Let the time associated to compute the estimation algorithm be $T_{est}$. Steps 1d involves estimation of states. Estimation involves computing the covariance matrix, Kalman gain, and the estimated states using the state equations, and the previous measurements. For step 2, it is required to compute predictions for $m$ time steps
and hence, the run-time computational complexity of the algorithm is dependent on the number of prediction steps added to a constant value for the other computations, that is, $O$(m) + constant. Depending on the computational capability and the accuracy required, the designer can decide on a trade-off between the two considerations.

\subsubsection{Balloon interception} 
\begin{figure}[t]
    \centering
    \includegraphics[width=0.8\linewidth]{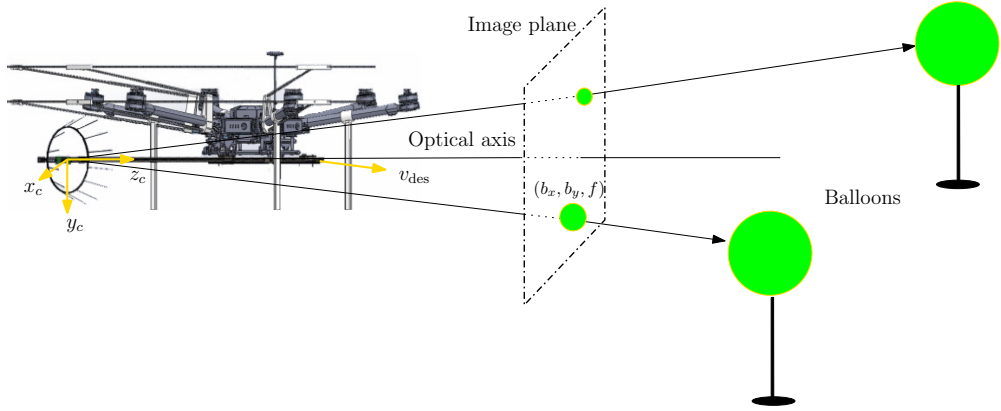}
    \caption{Balloon popping: the geometry showing the relative position of camera, image plane and the target balloons. A nearer balloon is approached first upon detecting multiple balloons.}
    \label{fig:Balloon_cam_persp}
\end{figure}
\begin{figure}
    \centering
    \includegraphics[scale=0.65]{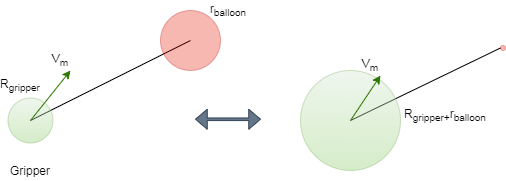}
    \caption{Balloon popper design. A comparatively small gripper is obtained for balloon interception.}
    \label{fig:gripper2}
\end{figure}

The guidance law developed for ball grabbing is modified to intercept the balloon, considering it as a stationary target. The engagement geometry during a balloon popping scenario is shown in Fig. \ref{fig:Balloon_cam_persp}. The UAV pops the nearest balloon first if multiple balloons are detected. In case of balloon interceptor UAV, the end effector is designed such that a nominal contact is sufficient to achieve successful popping. So, the equivalent radius desired for popping is the sum of the radius of the gripper and the radius of the balloon (when fit into the resultant virtual sphere, Fig. \ref{fig:gripper2}).
\begin{equation} 
    R_{eq}= R_{\text{gripper}} + R_{\text{balloon}}   \label{eqballoon}
\end{equation}
An analysis similar to that of grabbing drone gives the minimum virtual radius sphere to be 211 mm for the balloon popper speed of 2 m/s and alignment error of 25$^\text{o}$ at a distance of 0.5 m. The alignment error margin is kept high as the time of engagement for balloon popper UAV is less. Considering the balloon radius of 150 mm,  the effective gripper radius of the balloon popper UAV is found to be 61 mm (from \eqref{eqballoon}). The final value of the effective radius of balloon popper UAV is chosen as 100 mm. 

\subsection{Controller design}
\begin{figure}[t]
  \centering
  \includegraphics[width=.6\linewidth]{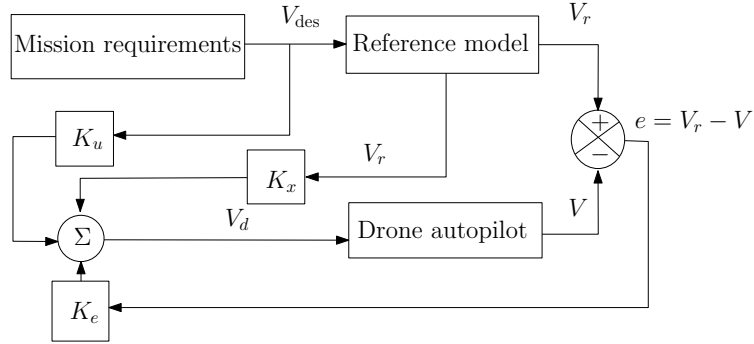}  
  \caption{Controller block diagram. This simple adaptive controller handles the possible uncertainties.}
  \label{fig:SAC_block}
\end{figure}

A variation in the system parameters is observed during and after the grabbing phase compared to other mission phases. Also, disturbances are expected due to wind gust. So, the controller needs to adapt to these variations. We have designed an add-on adaptive controller based on the concepts of Simple Adaptive Controller (SAC) to handle the uncertainty. The desired commanded velocity generated from the mission module is modified to accommodate the uncertainty. A typical block diagram is shown in Fig. \ref{fig:SAC_block}, where the desired velocity command from the mission module ($V_{\text{des}}$) is modified using add-on adaptive controller before feeding to the UAV autopilot. 
In Fig. \ref{fig:SAC_block}, $V_{d}$ is the commanded velocity to the autopilot, $V$ is the actual velocity, $V_{r}$ is the velocity output from the reference model and $K_{x}$, $K_{e}$, $K_{u}$ are the controller gains. Separate controller structure for velocity components and the yaw rate is developed and described in Algorithm \ref{alg:controller}. Following the previous method to compute the worst-case run time, the computational complexity of the controller algorithm is also a constant.

\begin{algorithm}[htb!] \label{}
\textbf{Input}:  Desired velocities ($V_{\text{desx}}, V_{\text{desy}}, V_{\text{desz}}$), desired yaw rate  ($r_{\text{des}}$), velocities $(V_{x}, V_{y}, V_{z}$), yaw rate ($r$), gains $(k_{ex}, k_{ey}, k_{ez}, k_{er}, k_{xx}, k_{xy}, k_{xz}, k_{xr},  k_{ux},  k_{uy}, k_{uz}, k_{ur}) $.
\begin{enumerate} 
    \item  Select individual  reference models for the desired responses of  actual velocities and yaw rates against the commanded values 
    \item  Calculate the output of reference models: $V_{rx}, V_{ry}, V_{rz}, r_{r}$ 
    \item  Calculate the error between the reference model output and the actual value: \\
         $ e_{x} \gets V_{rx}-V_{x}$ \\
         $e_{y} \gets V_{ry}-V_{y}$ \\
         $e_{z} \gets V_{rz}-V_{z}$ \\
         $ e_{r} \gets r_{r}-r $
    \item   Commanded  velocity to autopilot: \\
    $ V_{dx} \gets k_{ex} e_{x} + k_{xx} V_{rx} + k_{ux} V_{\text{desx}}$\\
    $ V_{dy} \gets  k_{ey} e_{y}  + k_{xy} V_{ry} + k_{uy} V_{\text{desy}}$ \\
    $ V_{dz} \gets  k_{ez} e_{z}  + k_{xz} V_{rz} + k_{uz} V_{\text{desz}}$
    
    \item  Commanded yaw rate to autopilot:\\
     $ r_{d} \gets  K_{er} e_{r} + k_{xr} r_{r} + k_{ur} r_{\text{des}}$
\end{enumerate}
\textbf{Output}: Commanded velocities ($ V_{dx}$, $ V_{dy}$, $ V_{dz})$, commanded yaw rate ($ r_{d} $).
 \caption{ Controller  Design}
 \label{alg:controller}
\end{algorithm} 

\subsection{Geo-fencing}
The UAVs should remain within the arena bounds while executing the challenge. A robust geo-fencing algorithm is needed to ensure that the UAVs remain within the predefined physical bounds. The arena is 100 x 60 x 20 m and is covered with mesh. The boundaries of the environment are set on a global frame. So, the convex 3D hull is chosen as a candidate geometric representation of the geo-fence. Since the boundaries of the arena are known, we use the 3D quick hull algorithm \cite{barber1996quickhull} to compute the set of faces, which is a set of vertices, corresponding to the face created by the boundary points.  The 3D quick hull is an iterative algorithm that adds individual points one after the other to create intermediate hulls. It creates an initial hull with faces and iteratively visits all the faces to find the horizon. After termination, a list of all the horizons is obtained in a counter-clockwise direction, which are the edges of the convex hull. The faces of the physical boundaries are obtained, and the nearest point projected by the UAV on each face is calculated. These projected points are treated as obstacles for the UAV, and at a certain distance, the UAV is subjected to the repulsive force along the line of sight of the projected point and the  UAV. A commanded velocity keeps the UAVs away from the arena boundary.

\subsection{Collision avoidance}
To avoid unforeseen inter-UAV collision cases, an artificial potential field algorithm is designed based on collision cone criteria \cite{chakravarthy2012generalization}.  Collision avoidance based on proportional navigation using the visual information is reported in \cite{clark2015vision}. However, capturing and processing the visual information from all sides is not feasible considering the space, power, and payload constraints. The collision avoidance algorithm is designed by sharing the position and velocity information among the UAVs. The relative velocity along the line of sight $(V_{r0})$ and its normal $(V_{\theta_{0}}, V_{\phi_{0}})$ are used to check the collision condition. If the distance between the pair of UAVs is $r_0$ and the radius of the safe sphere to avoid the collision is $R$, then the following condition is used to check if they are in the collision cone \cite{chakravarthy2012generalization}. 
\begin{equation}
     r_{0}^{2}(V_{\theta 0}^{2} + V_{\phi 0}^{2}) \leq  R^{2}  (V_{\theta 0}^{2} + V_{\phi 0}^{2}+ V_{r 0}^{2})
\end{equation}
\begin{equation}
     V_{r0}< 0
\end{equation}     
Once a UAV is in the collision cone and within a certain distance from an incoming UAV, it is subjected to acceleration away from the UAV along the line of sight of both UAVs. The magnitude of the acceleration is selected based on the velocity and initial distance between the UAVs. 
In some cases, one UAV is assigned as a high priority if it is involved in carrying out any critical tasks and the other UAVs as low priority. In those cases, only the low priority UAV maneuvers to avoid the collision. 
\section{Mission Modes} \label{s5}
The overall mission is classified into several tasks so that it can be allocated to different UAVs efficiently. The tasks are designed to minimize the conflicts among the them as well as to expedite the overall completion time of the mission. The important tasks are take-off, exploration, tracking, search and grabbing, popping and landing. They are described below. 

\subsection{Exploration}
\begin{figure}[t]
    \centering
    \begin{subfigure}{0.45\columnwidth}
        \centering
        \includegraphics[scale=0.5]{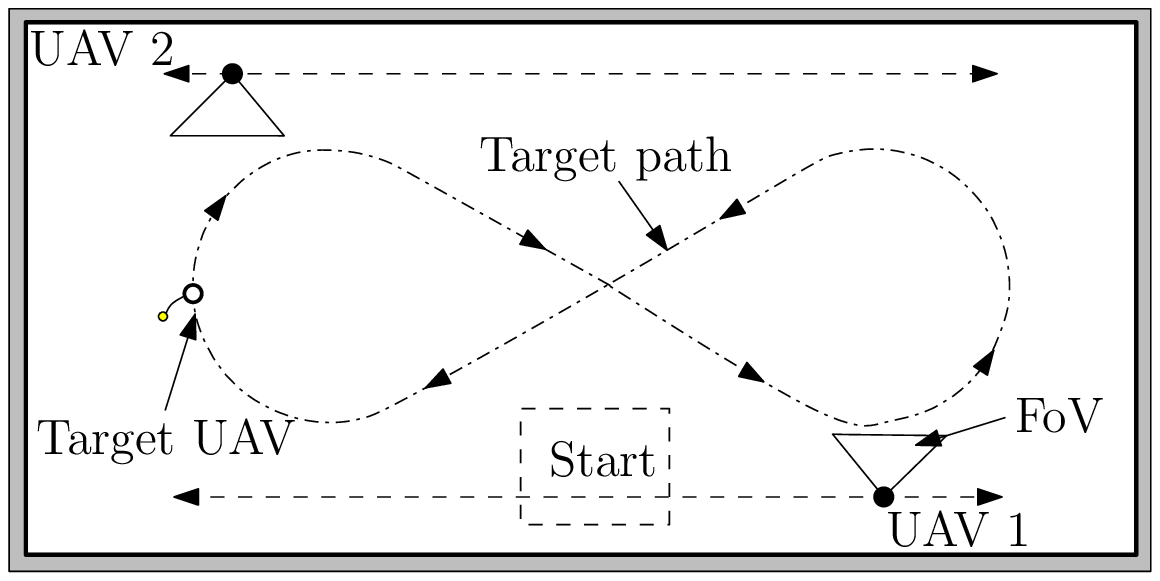}
        \subcaption{}
    \end{subfigure}
    \begin{subfigure}{0.45\columnwidth}
        \centering
        \includegraphics[scale=0.5]{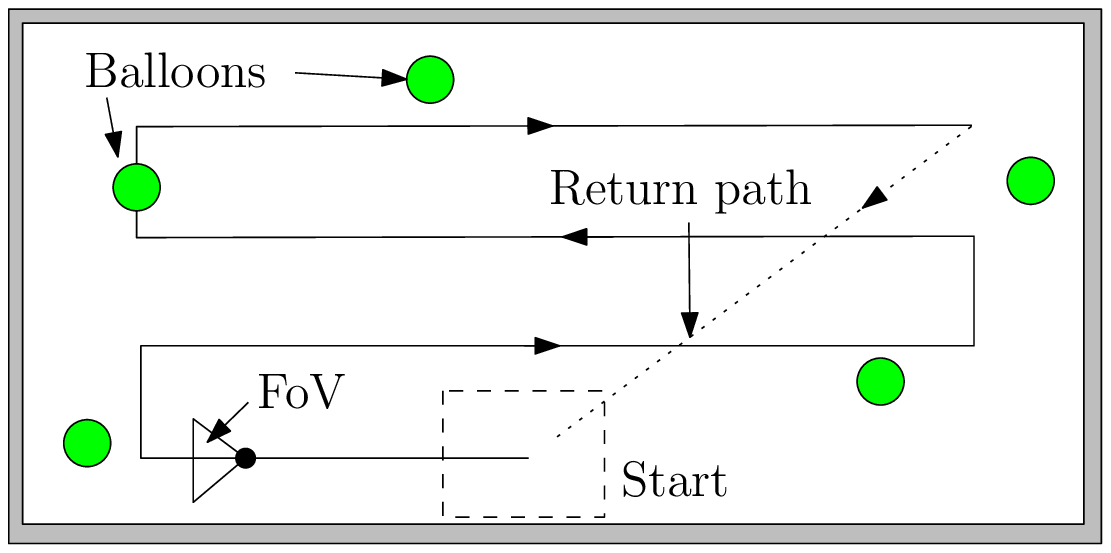}
        \subcaption{}
    \end{subfigure}
\caption{Exploration pattern for (a) grabbing (b) popping. The UAVs face the camera to the inside of the arena while exploring to detect the ball, while for exploration to detect balloons, the UAV heading is along the path.}
\label{f1}
\end{figure}

The exploration task is accomplished through waypoint-based navigation wherein a set of waypoints are supplied to the UAVs. Grid-based waypoints in an exhaustive lawn-mower pattern are fed to the UAV performing the balloon popping mission. On the other hand, waypoints with a horizontal scanning pattern between two locations are fed to the UAVs performing the ball grabbing mission. The UAVs, while navigating through waypoints, orient themselves in a direction that keeps the camera field of view  perpendicular to the waypoint trajectory and toward the inside of the arena, as shown in Fig. \ref{f1}. The UAV performing the balloon popping task, however, always orients towards its next waypoint.      

\subsection{Target tracking}
The purpose of the target tracking task is to provide the position data of the path followed by the target. This position data is used to estimate the type of repetitive curve followed by the target and its parameters. The target tracking task is triggered if a UAV detects the target while performing the exploration task. This task is considered complete once the curvature parameters of the target UAV's path are estimated. This is done cooperatively by the UAVs. Only one UAV at a time can perform this task while the other UAV stays at a standby position. The standby position is fed into the algorithm before execution, considering the feasibility of detection and grabbing within the given arena. Upon losing the target while tracking, the UAV goes to a standby position and the other UAV resumes tracking.

\subsection{Grabbing}
The grabbing task is two-fold. The interceptor UAV should be guided to detach the ball, and once it is detached, feedback needs to be generated. Thus, the main sub-tasks associated with the grabbing operation are as follows.
\begin{enumerate}[label=(\alph*)]
    \item {\bf {Aerial manipulation:}} While performing ball grabbing operation, the UAV guides the camera attached to the manipulator to the center of the ball. While doing so, the rigid part of the manipulator must hit the magnetic link attached to the target ball. During the terminal phase, when the ball fully covers the field of view of the camera, additional forward momentum is generated by the UAV so that it can detach the magnetic link with the detachment mechanism of the manipulator. The detached ball drops into the manipulator net.
    \item {\bf{Grab detection:}} On the bottom of the manipulator end effector, limit switches are placed over a circular mount. They are calibrated to activate when the ball falls on them. The calibration is such that a minimum weight of 50 g is enough to detect the ball's presence in the basket. The triggering of limit switches sends the desired feedback and completes the grabbing task. The mesh is fixed to the end-effector in a conical shape to ensure the falling of the ball on the limit switches.
\end{enumerate}
The UAVs cooperate to achieve successful grabbing. In case a UAV loses the ball from its FoV, the other UAV resumes its grabbing mission while the former moves to the grabbing stand-by location. This not only improves the probability of grabbing but also makes the grabbing operation quick and efficient.
\subsection{Balloon popping} 
\begin{figure}
    \centering
    \includegraphics[scale=0.8]{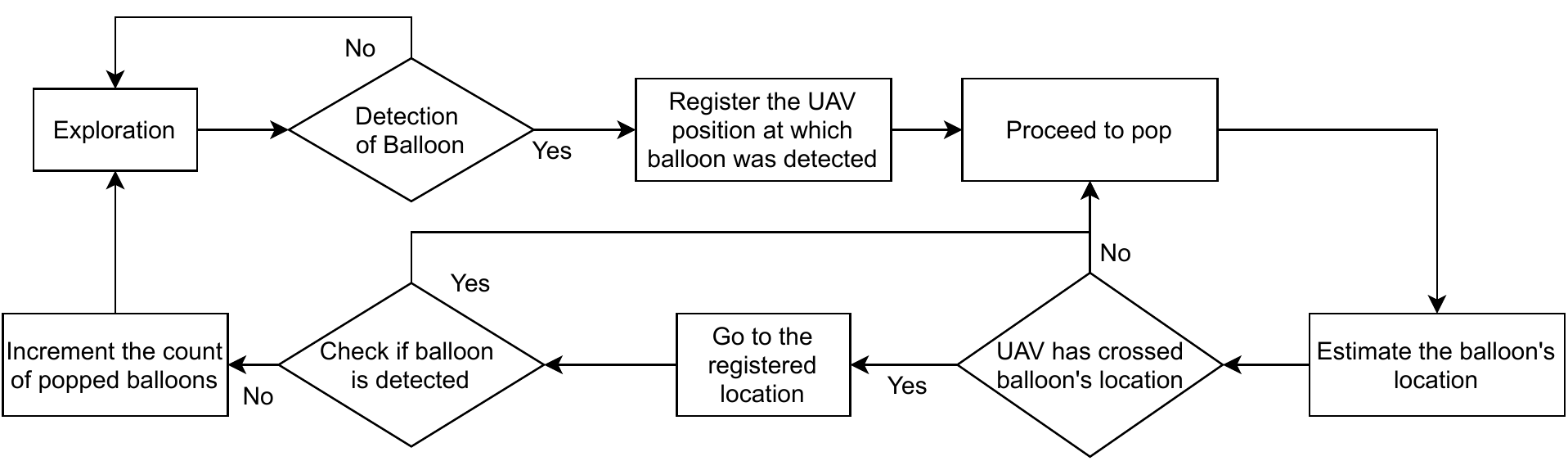}
    \caption{Balloon popping mission.}
    \label{fig:balloon_popping_mission}
\end{figure}

This task starts when a balloon is detected during the balloon exploration. The UAV registers its position at which the detection occurred and estimates the position of the detected balloon. This helps in the confirmation of popping later. The UAV proceeds to pop the balloon, by servoing towards the tracked center of the balloon in the image. Forward velocity commands are generated until there is no balloon in the camera field of view for a few continuous frames, and the current position of the UAV is beyond the position of the balloon estimated earlier. To determine the relative position of the UAV with respect to the balloon, two vectors, originating from the current UAV position and one ending at the earlier registered UAV position and the other at the balloon's estimated position, are used to compute a dot product. The sign change of this vector determines if the UAV has gone beyond the balloon's position. Further, to confirm a successful pop, the UAV returns to the previously registered position and checks if there is a balloon in its field of view. If not, the UAV gets back to the exploration mode and continues its mission. The program also keeps count of the popped balloons. The mission is assumed to be complete when the number of popped balloons is equal to five or if no balloon is detected in two rounds of exploration. A flow chart for the balloon popping module is shown in Fig. \ref{fig:balloon_popping_mission}. It summarises the overall flow of control while the balloon popping module is being executed.

\subsection{Other subsidiary modules}
Apart from the above modules, the two other essential modules are the take-off and landing modules. The take-off module is run through a standard UAV take-off Software Development Kit (SDK) routine, whereas landing is performed using assistance from vision and a landing SDK routine. Another module is grabbing standby, where the UAV moves to the standby location and waits until it receives the command to switch to grabbing.  Detection of the ball, balloon, and landing zone is performed through different cameras. Detection modules are run in parallel to the main task node so that visual information is obtained continuously. To handle resets in the competition, a restart module is also included along with the subsidiary modes. This will ensure that all the mission parameters are saved for resuming the mission after the reset is executed. This task essentially acts as a `pause' for the mission execution.

\section{Operation Management System (OMS)}\label{s6}
\begin{figure}[t]
     \centering
     \includegraphics[scale=0.55]{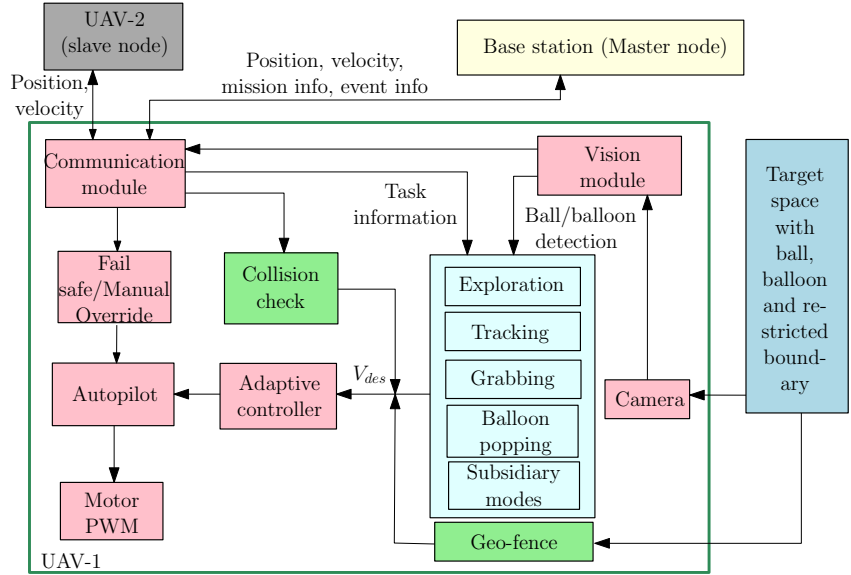}
     \caption{The OMS architecture and flow of control among UAVs and the base station. }
     \label{fig:Overall_block.png}
\end{figure}

We developed a complete software architecture on ROS for deploying multiple UAVs simultaneously to carry out the challenge autonomously. Two UAVs collaborate to grab the ball while the third one is dedicated to pop the balloons. The software architecture handles the flow of control to execute the entire mission. It is also responsible for coordinating the UAVs to execute the operation smoothly. The OMS coordinates task allocation amongst the UAVs by sharing the task information and their state information through a multi-master network setup. The OMS  architecture is shown in  Fig. \ref{fig:Overall_block.png}. It shows how the different mission modules integrate within the system. The mission modes coordinate with other system modules and with the base station. Specific aspects like collision avoidance are achieved by communication with the other UAVs.

\begin{figure}
    \centering
    \includegraphics[scale=0.5]{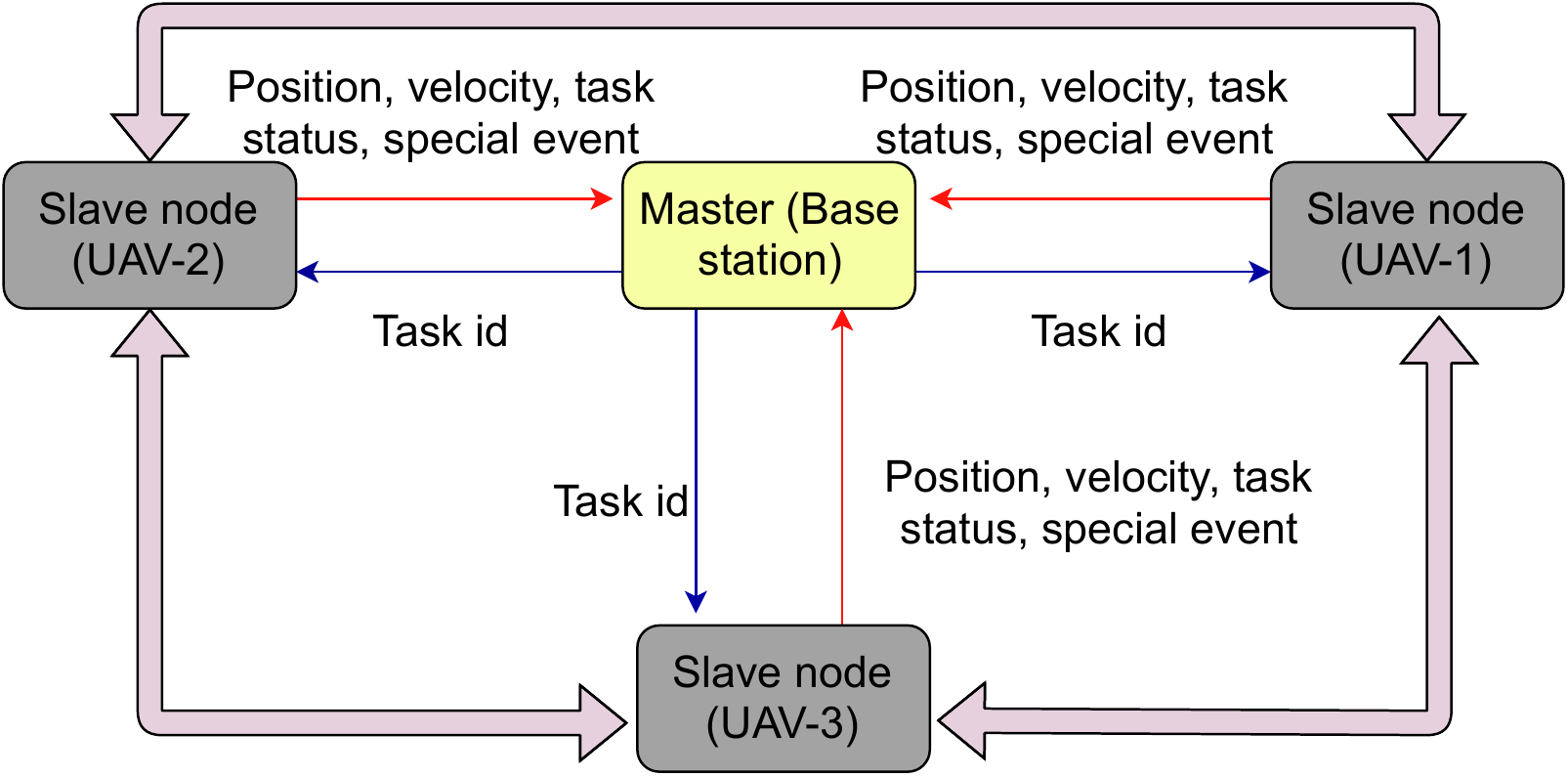}
    \caption{Information sharing in the base station-UAV network.}
    \label{fig:multimaster}
\end{figure}

In a multi-master network, the multi cast feature of network adapters is used to share the topics amongst the different nodes on different computers over the network. The IP addresses of all the computers in the network are added to the multi cast group in each local computer. This allows for inter-UAV communication of the task and agents state information among the agents in the multi cast group. In this setup, the master node does task allocation and provides the task information to the slave nodes based on the current states of all agents and mission requirements. Master nodes receive important state information such as position and velocity, along with the status of the current task of the slave nodes.  Slave nodes share only the position and velocity information among themselves to avoid collision among themselves. The master node can be on the base station or on any of the UAVs. However, keeping the master node in the base station is preferable, considering the computational requirement for the master node as well as to improve the safety of the overall system. The information sharing among the different agents of the network is kept at a minimum to avoid the load on the network. The schematic representation of the communication system is shown in Fig. \ref{fig:multimaster}.
Besides managing the mission tasks, OMS also does some basic functionalities. These include task allocation, reallocation in case of failure, fail-safe mechanism and system monitoring. 
\begin{enumerate}[label=(\alph*)]
\item   \textbf{Task allocation:} OMS schedules the tasks requiring the control inputs; whereas, tasks like capturing images and ball/balloon detection are run as a separate module continuously. The tasks are classified based on priority. Safety-critical tasks like inter-agent collision avoidance, geo-fencing, etc.  are given high priority; whereas, mission-specific tasks like exploration, tracking, and grabbing are given low priority. The high priority tasks are run in parallel while only one mission specific task is selected at a time for execution. OMS selects the mission specific tasks based on the state of the agents and mission requirement. The desired acceleration from the mission specific tasks and the safety-critical tasks are combined to obtain the commanded velocities for the current mission. 

The mission specific tasks are further classified into static tasks and dynamic tasks.  Static tasks are predefined for a mission, whereas dynamic tasks are event-triggered tasks. These triggering events can be the detection of a target, grabbing of the ball, or popping of a balloon, etc.  Some examples of static tasks are exploration for balloon and ball detection, whereas dynamic tasks are target tracking triggered by the detection of the ball by the UAV. OMS can create or delete tasks as per the mission requirement.

\item\textbf{Task reallocation and redundancy:} Task reallocation is useful if the agents are capable of performing multiple missions. A UAV can have a hybrid manipulator capable of performing ball grabbing and balloon popping missions and thus can participate in other tasks after finishing its first assigned task. Situations may arise where agents are redundant. One such example is when two UAVs are deployed for grabbing the ball, and one of them finishes its mission. In such cases, OMS identifies the redundant task and associated UAV. Further, it reassigns that particular agent to another task, speeding up the latter.

\item\textbf{State monitoring and fail-safe mechanism:} The master receives the information from all its agents, thus helping it monitor their states (position, velocity, etc.) apart from the status of their tasks. This ensures that a smooth reallocation of tasks can be achieved in the event of agent failure or redundancy. In case of any unexpected situations, a fail-safe mechanism can help redirect the UAVs' control to a pre-programmed set of instructions. This will prevent crashes and safety breaches.
\end{enumerate} 

\section{ROS Simulations}\label{s7}
\begin{figure}[htb!]
\centering
\begin{subfigure}{0.45\columnwidth}
     \centering
     \includegraphics[scale=0.115]{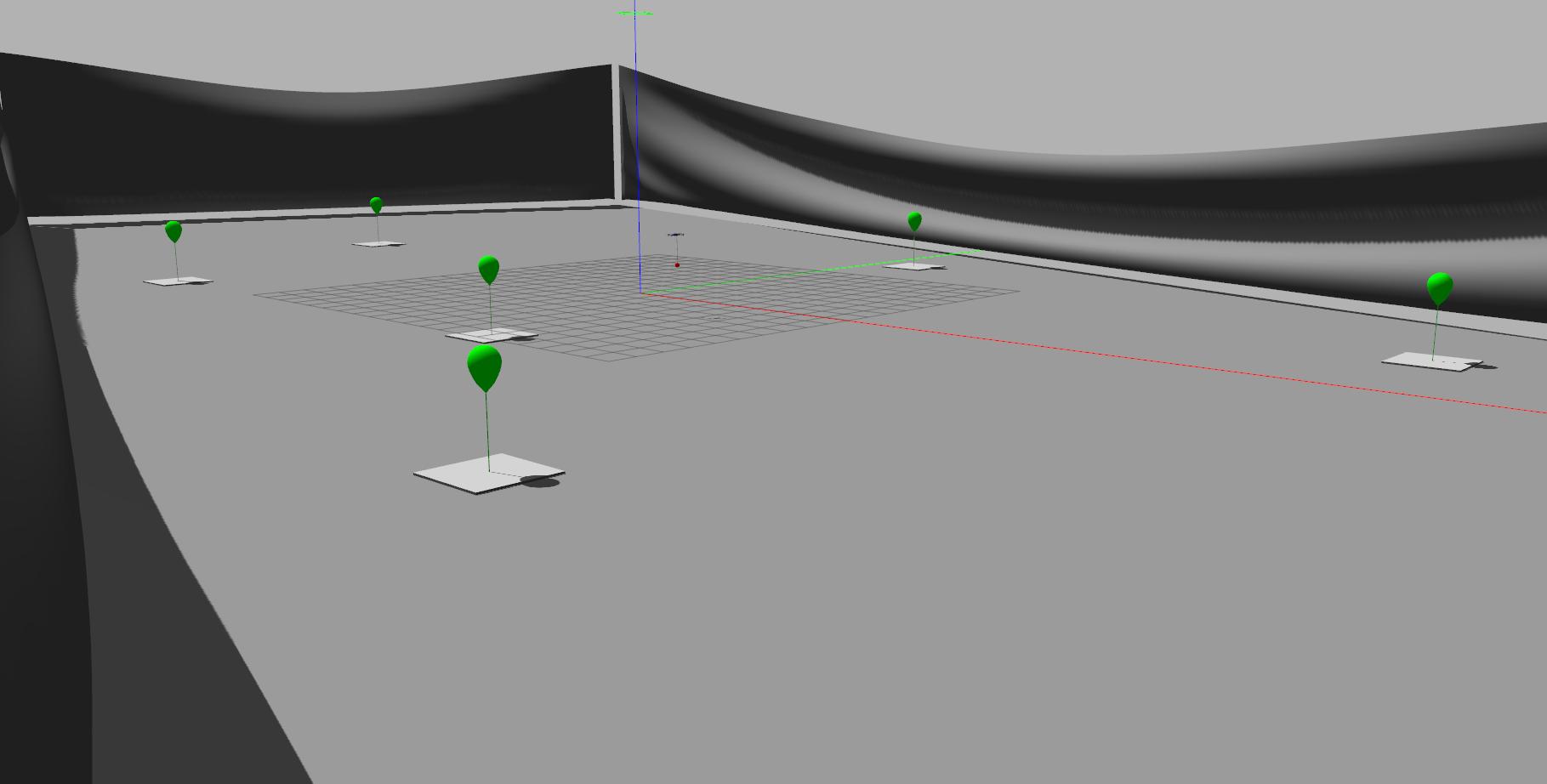}
     \caption{}
     \label{fig:arena_corner}
\end{subfigure}
\begin{subfigure}{0.45\columnwidth}
     \centering
     \includegraphics[scale=0.115]{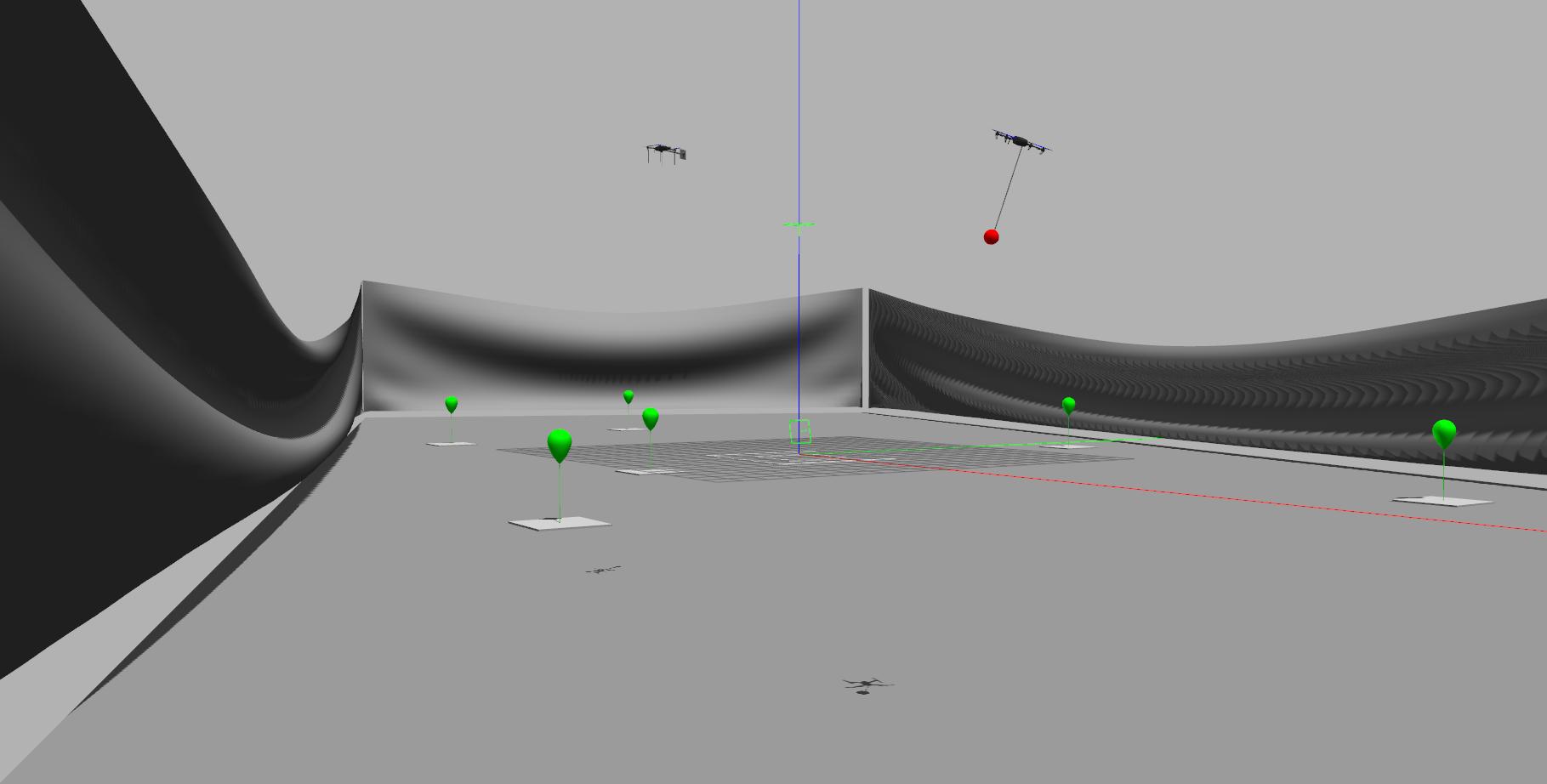}
     \caption{}
     \label{fig:arena_side}
\end{subfigure}
        \begin{subfigure}{0.45\columnwidth}
     \centering
     \includegraphics[scale=0.108]{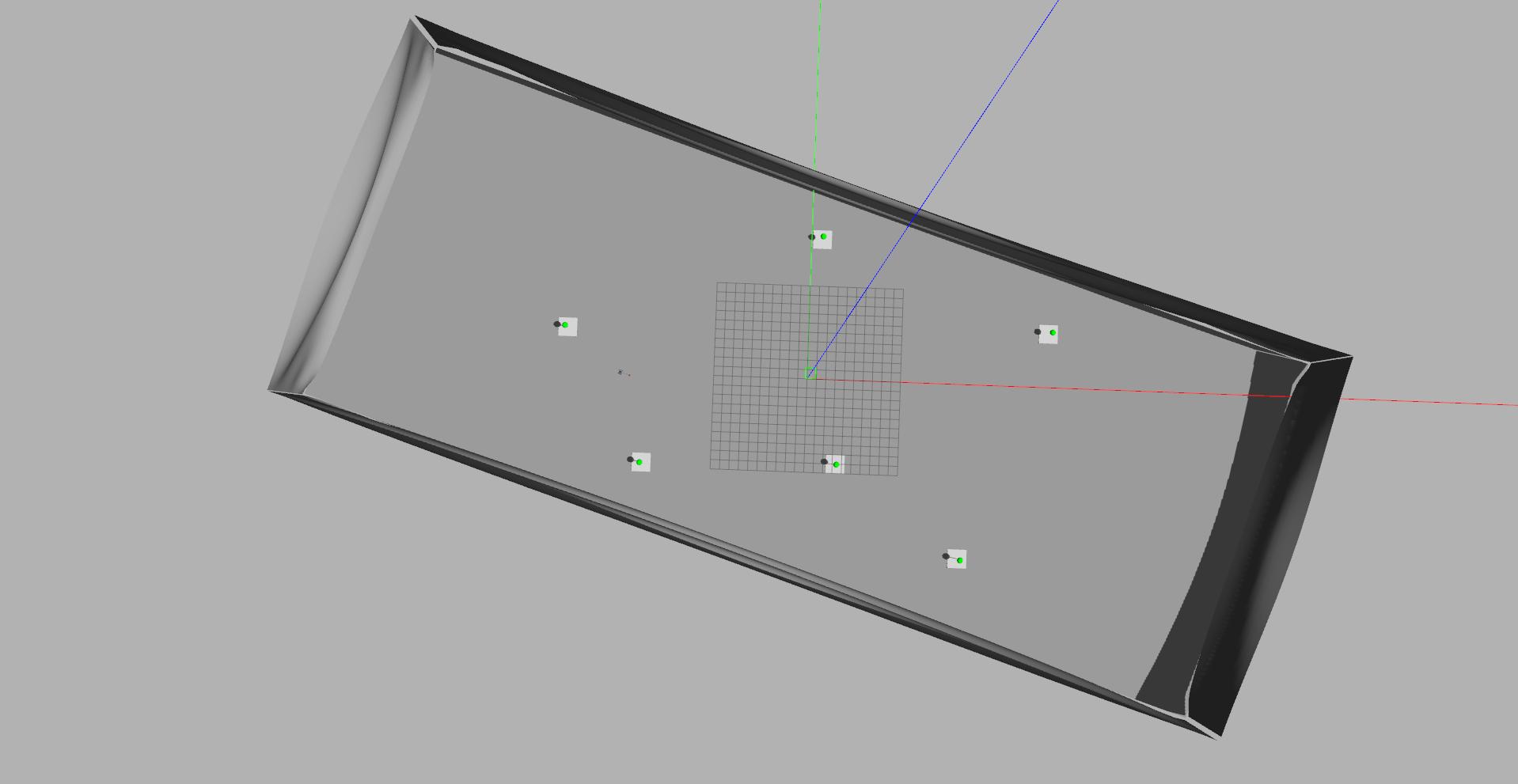}
     \caption{}
     \label{fig:arena_top}
\end{subfigure}
\begin{subfigure}{0.45\columnwidth}
     \centering
     \includegraphics[scale=0.115]{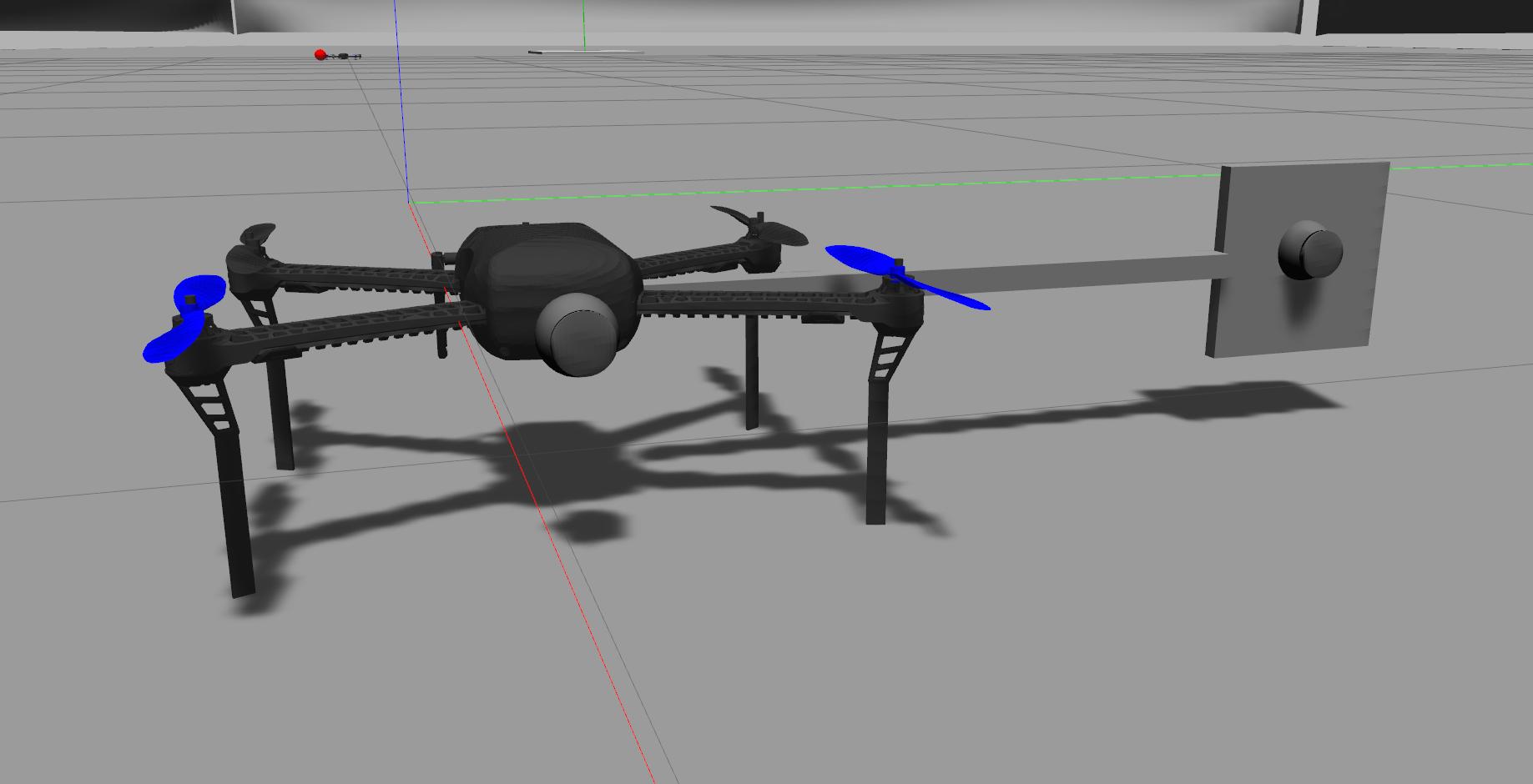}
     \caption{}
     \label{fig:iris}
\end{subfigure}
\caption{(a)-(c) Arena for Challenge 1 in Gazebo. The balloons are erected as per the specifications. The net enclosure of the competition arena is recreated using a custom model. The orientation of the arena is made similar to the venue (d) Iris drone with a sideways manipulator arm and camera attached to it.}
\label{fig:arena_Gaz}
\end{figure}
Owing to the complexity of the tasks, the inconvenience of field testing of three UAVs at once frequently, and rapid algorithm testing, we developed a virtual environment in Gazebo that simulates the arena of Challenge 1 in terms of dimensions and objectives. The following aspects were simulated.

\begin{enumerate}
\item Base Arena: The arena of 100 m $\times$ 60 m $\times$ 20 m is created using a simple rectangle of the same dimensions (Fig. \ref{fig:arena_corner}-\ref{fig:arena_side}). To simulate the nets, a cloth object around the arena was simulated physically in the 3D modeling software Blender and then exported to Gazebo. The whole arena is tilted by 18 degrees with respect to magnetic North, to match with the inclination of the Challenge 1 arena at ADNEC, Abu Dhabi (Fig. \ref{fig:arena_top}). 

\item Balloons: Six balloons of prescribed size and color are erected, and the wind effect is emulated using a sinusoidal varying force applied to the balloon link, in random directions. This gives the desired sway to the balloons.  The sway is facilitated by the presence of a ball joint between the balloon link and the rectangular base.

\item Target UAV: The target UAV used is 3DR Iris. The ball of 0.1 m diameter is hung from this UAV using a thin cylindrical link of the specified dimensions by a ball joint so that the ball sways freely as it would on the real UAV. The target UAV repeatedly moves in a figure-of-eight path, whose major and minor axes, yaw, pitch, and roll, can be configured.

\item Interceptor UAVs: The interceptor UAVs are again 3DR Iris, with the manipulation mechanism fixed at the side using a fixed joint. To simulate the grabbing manipulator, a flat plate of approximate size to the real grabber is attached at 0.6 m from the UAV center of gravity (Fig. \ref{fig:iris}). It also has a simulated camera on the flat plate, as in the real manipulator. The simulated camera has the same field of view and resolution as the real one. 
\end{enumerate} 

\begin{figure}[htb!]
    \centering
    \includegraphics[scale=0.25]{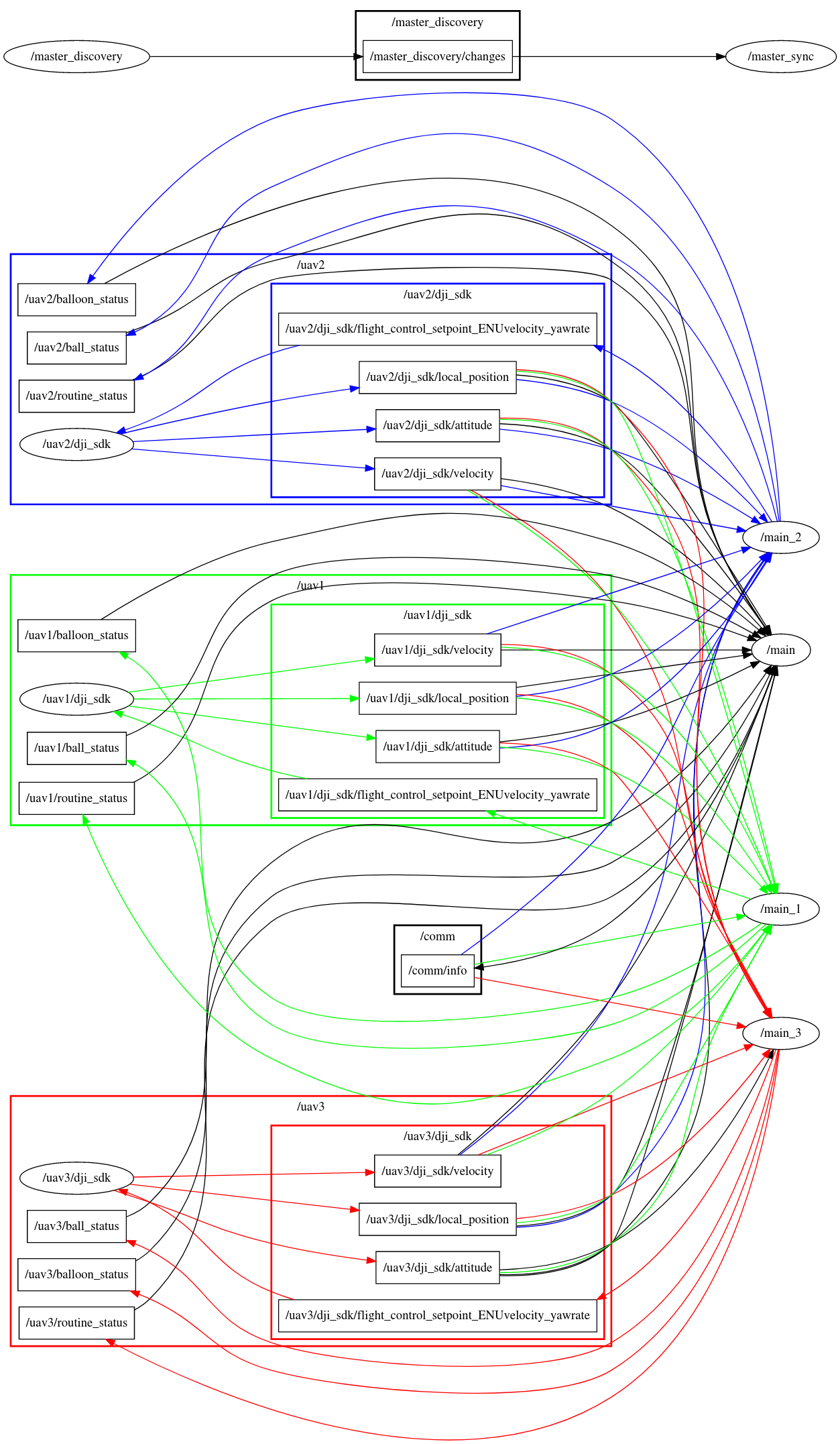}
    \caption{Network representation of the ROS nodes involved in the operation manager.}
    \label{fig:ROS_multi_master.png}
\end{figure}

The ROS node communication architecture for the three UAVs, is shown in Fig. \ref{fig:ROS_multi_master.png}.
In the figure, oval shapes denote the ROS nodes, while the rectangular shapes denote the ROS topics being published and subscribed from these nodes. They are further grouped according to their namespaces. The topics belonging to each UAV are assigned the corresponding color to visualize the data flow. It can be seen that each UAV (main\textunderscore1, main\textunderscore2, and main\textunderscore3) is listening to the position, velocity, and orientation topics of the other UAVs for collision avoidance calculations. The Ground Control Station node (main) listens to all of these plus the individual task state topics reported by each local node (ball\textunderscore status and  balloon\textunderscore status) and issues tasks on the routine\textunderscore status topic. 
\begin{figure}[htb!]
    \centering
    \includegraphics[scale=0.5]{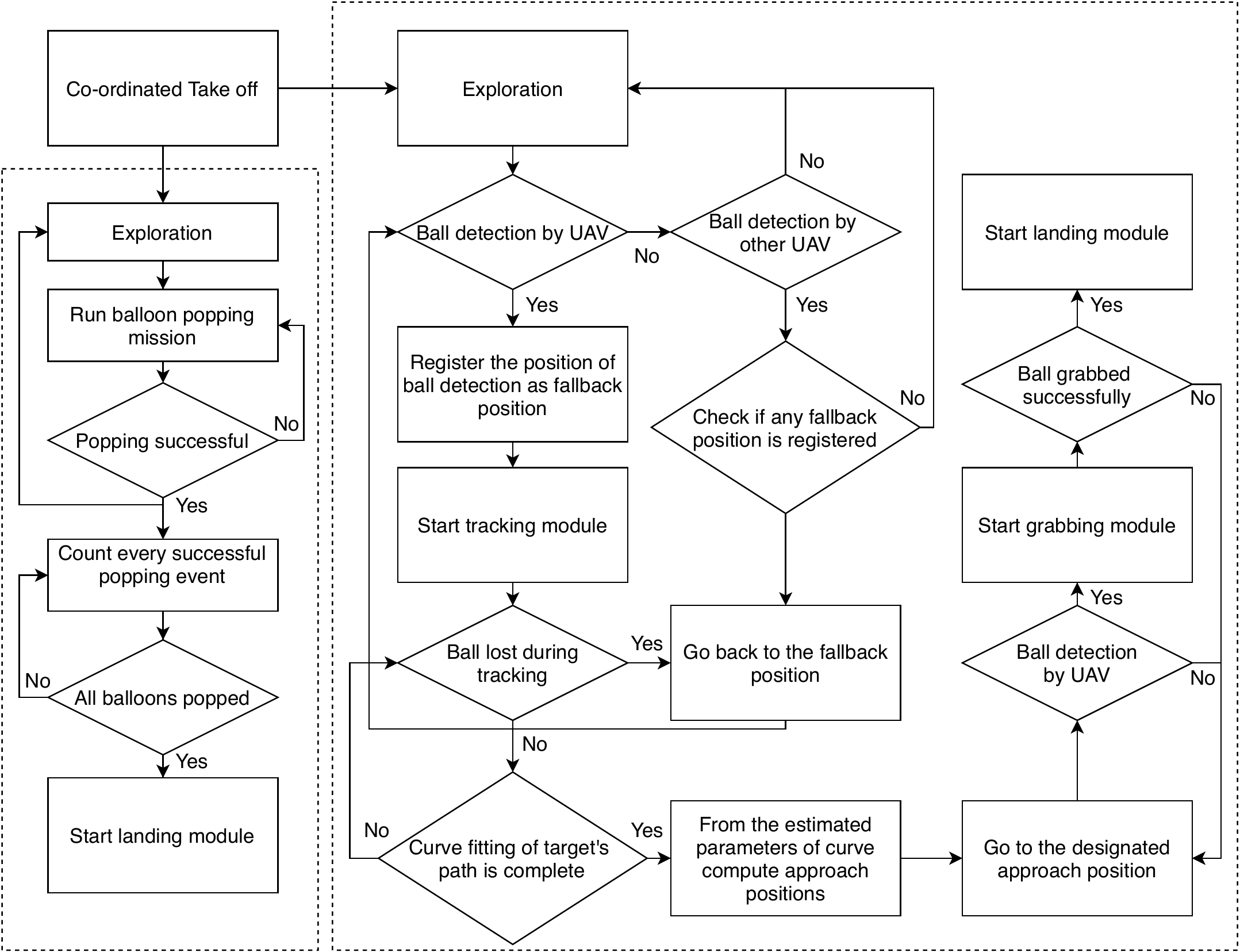}
    \caption{Flow of control for the complete mission.}
    \label{fig:ball_grabbing_mission}
\end{figure}
\begin{figure}[htb!]
    \centering
    \includegraphics[width=10cm, height=7.7cm]{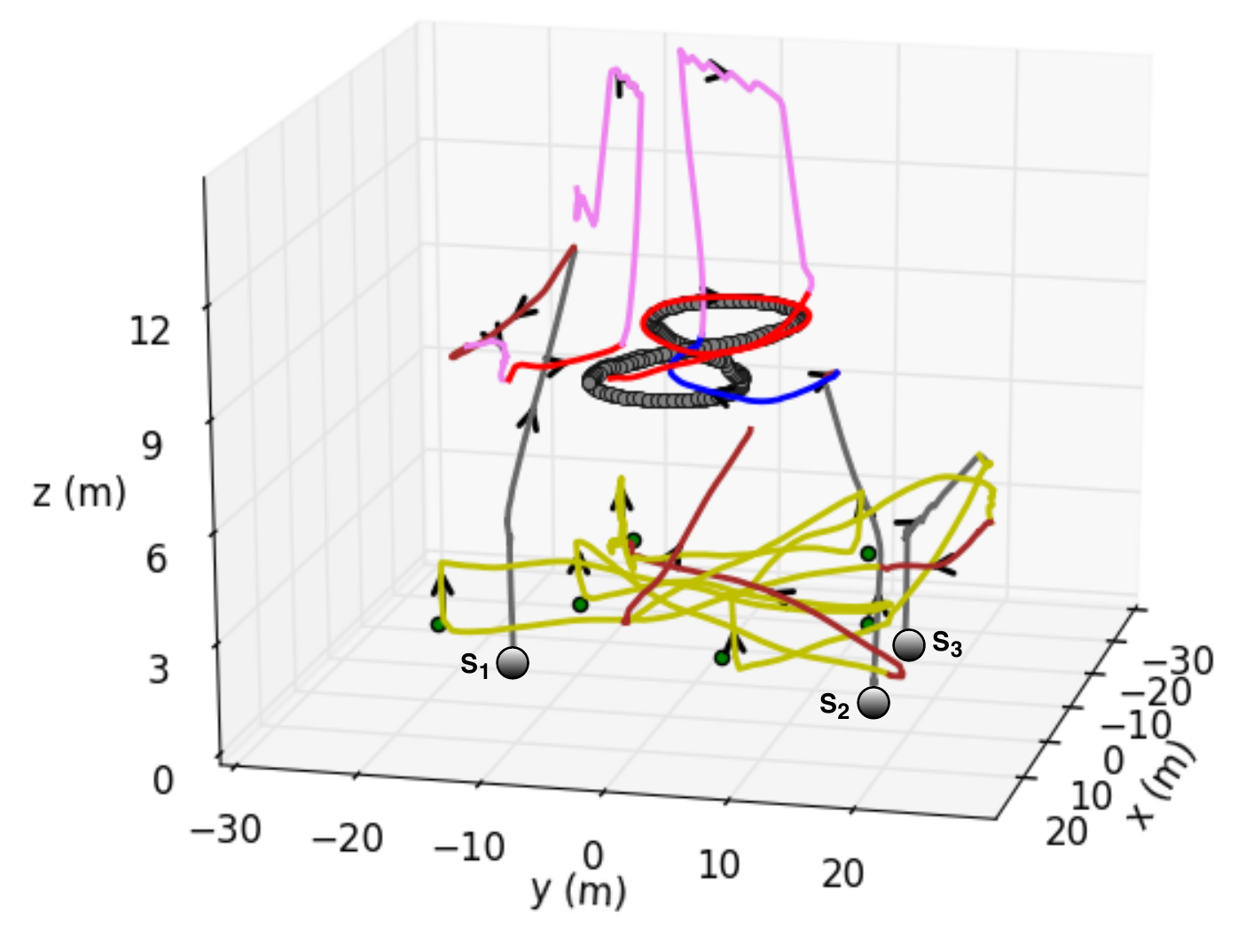}
    \caption{Task allocation during complete mission. The color codes correspond to the task switching as shown in Fig. \ref{fig:OMS}.}
    \label{fig:ch1_full_sim_trace}
\end{figure}
\begin{figure}[htb!]
    \centering
    \includegraphics[width=1.05\columnwidth, height=8cm]{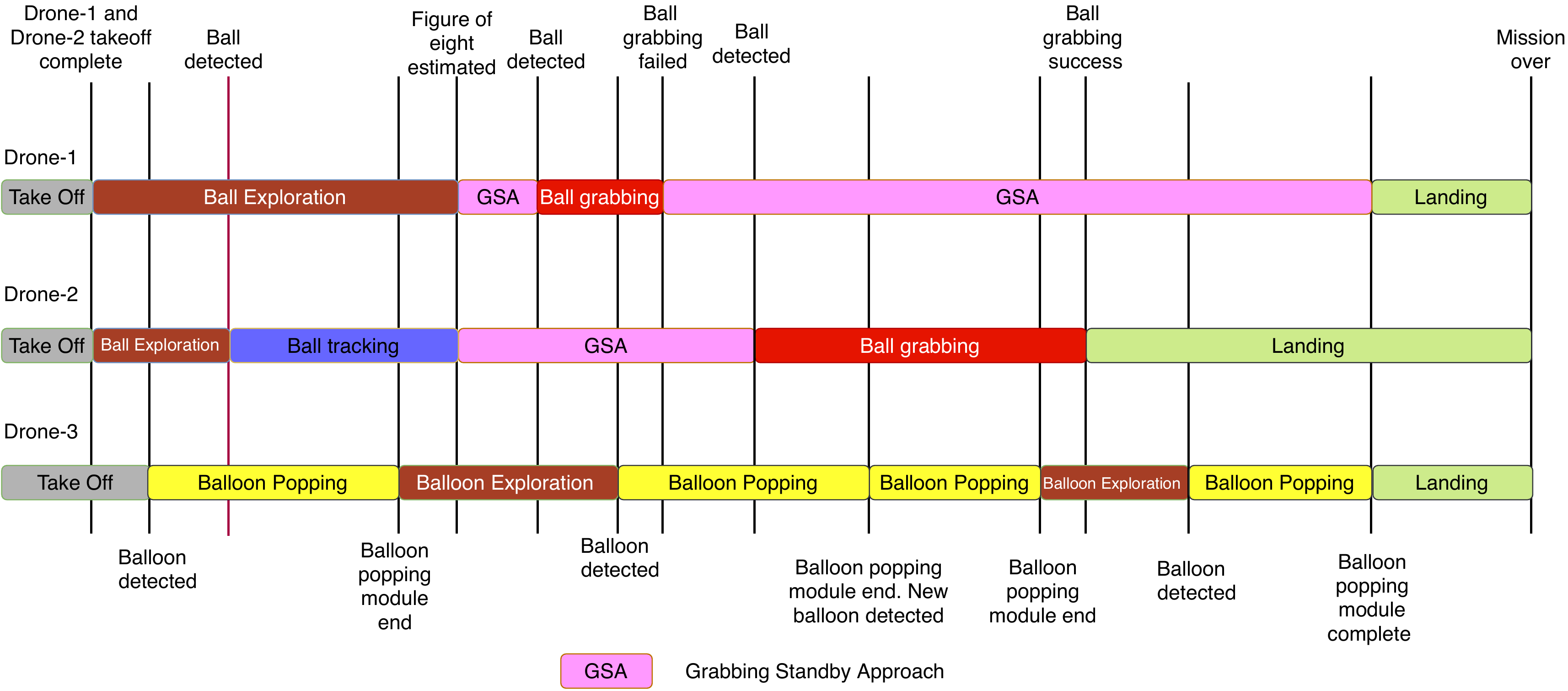}
    \caption{Task switching  showing the switching of control between different tasks in the OMS. }
    \label{fig:OMS}
\end{figure}

\begin{figure}[htb!]
    \begin{subfigure}{0.5\columnwidth}
        \centering
        \includegraphics[scale=0.4]{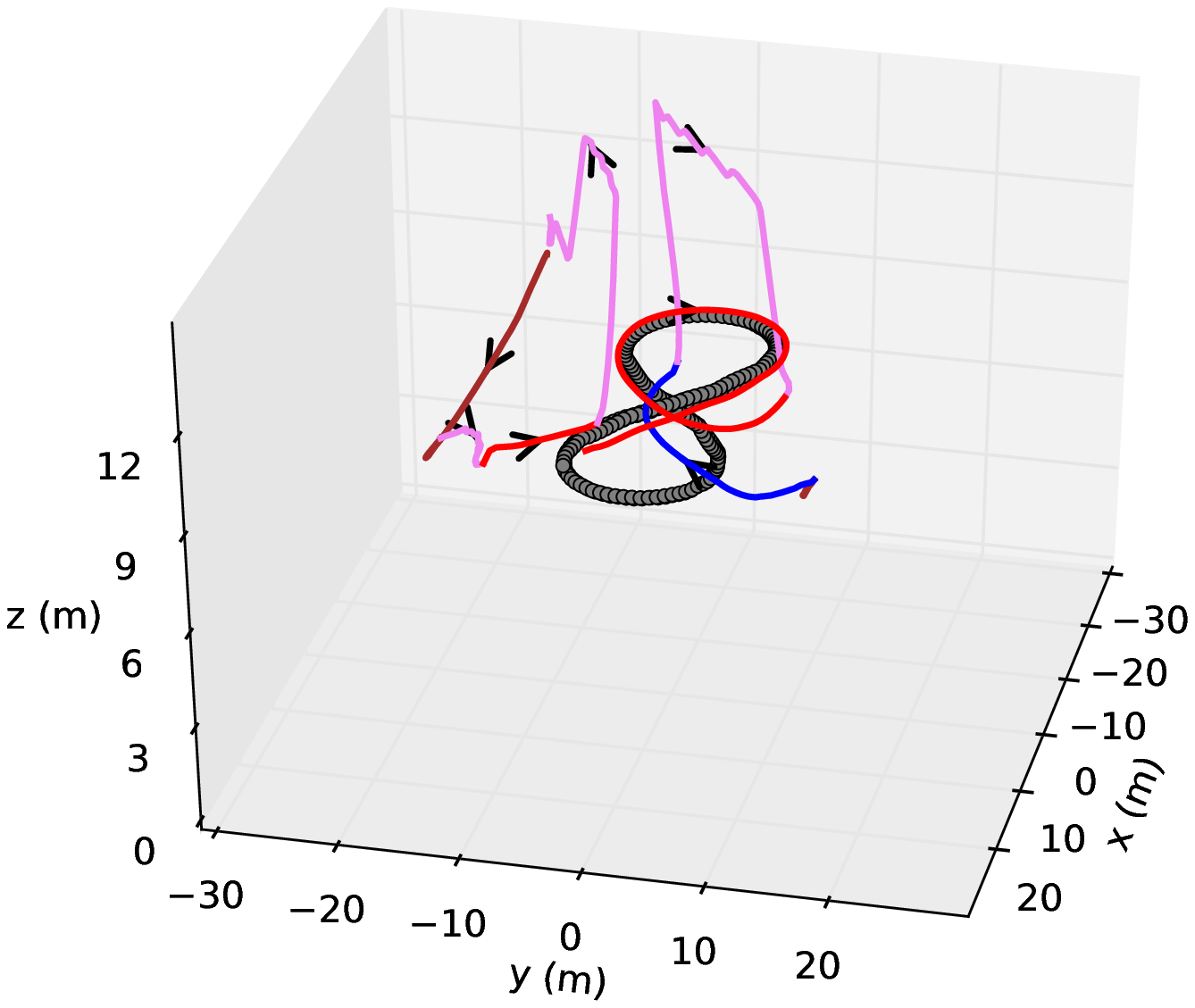}
        \caption{}
        \label{fig:grabbing_and_tracking_trace_full}
    \end{subfigure}
    \begin{subfigure}{0.5\columnwidth}
        \centering
        \includegraphics[scale=0.6]{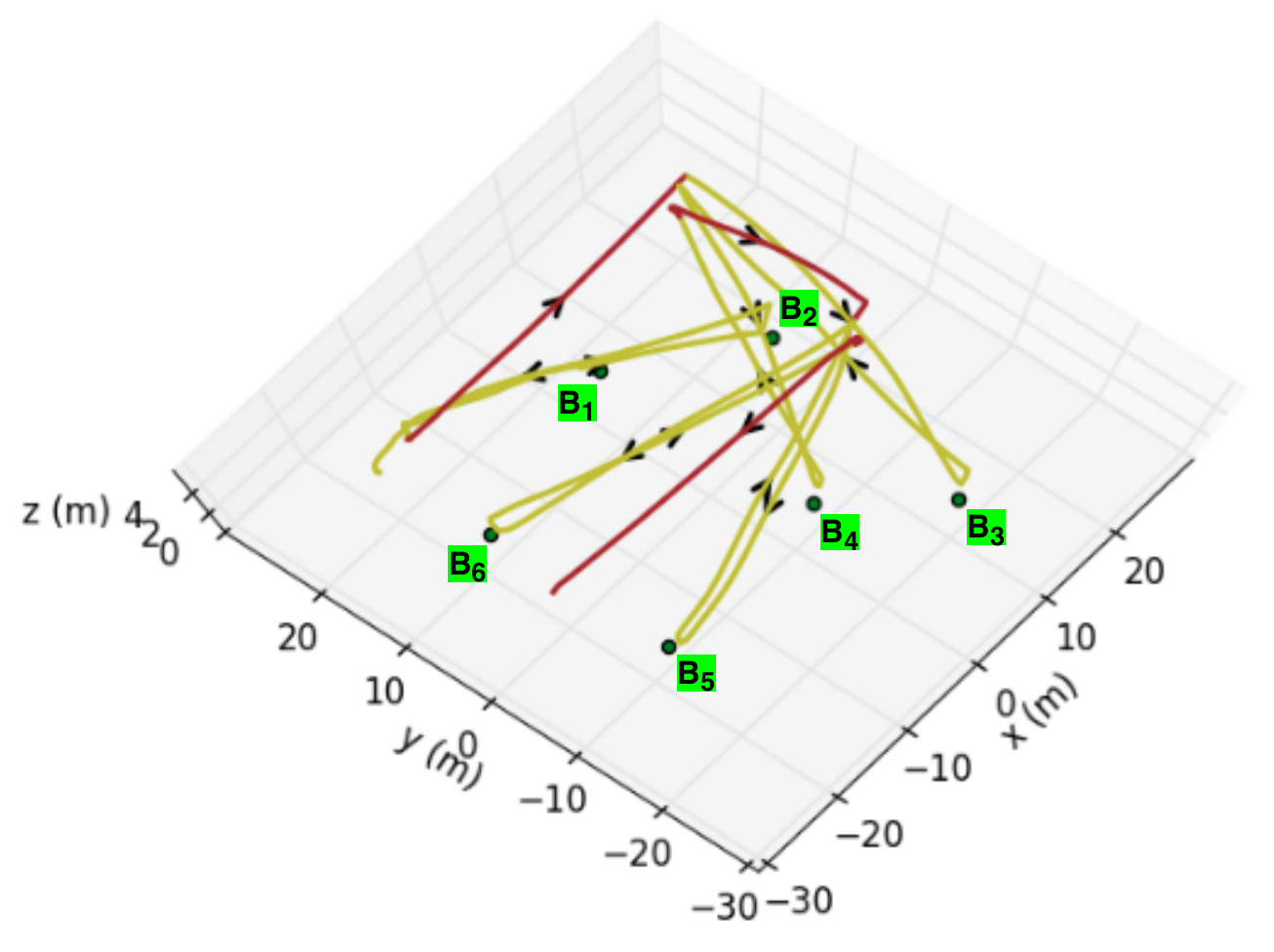}
        \caption{}
        \label{fig:popping_full_trace_line}
    \end{subfigure}
    \caption{(a) Ball grabbing mission (b) Top view of balloon popping mission.}
\end{figure}

Simulations are performed to execute the complete mission and task allocation as per the mission described in Fig. \ref{fig:ball_grabbing_mission}. Two UAVs cooperatively grab the ball while a single UAV achieves the balloon popping mission. The flow of control among the UAVs are shown in the flowchart. The trajectory of three UAVs is plotted in Fig. \ref{fig:ch1_full_sim_trace}, where the different color indicates the different tasks for the UAVs as commanded by OMS. S1, S2, and S3 are the take-off points for UAV 1, UAV 2, and UAV 3, respectively. The allotted tasks by OMS among the different agents are presented through the task switching diagram,  shown in Fig. \ref{fig:OMS}, and corresponding trajectories (Fig. \ref{fig:ch1_full_sim_trace}) of the UAVs are plotted in the same color.  UAV 1 and UAV 2 are initially allocated static tasks of ball exploration after take-off, while UAV 3 is allocated the static task of balloon exploration. The vertical lines in Fig.  \ref{fig:OMS} indicate the switching of a task due to the completion of a task or a special event like a ball detection or failure to grab. The trajectories of the UAVs specifically during ball grabbing and balloon popping are plotted in  Fig. \ref{fig:grabbing_and_tracking_trace_full} and Fig. \ref{fig:popping_full_trace_line}. In  Fig. \ref{fig:popping_full_trace_line}, B$_1$-B$_6$  are the different balloon locations.The balloons are popped in the order from B$_1$ to B$_6$. The simulation videos\footnote{\href{https://youtu.be/v4laz8aNPpA}{Simulation videos}} demonstrate the mission execution and performance of the developed system in the Gazebo environment.  

\section{Hardware Design}\label{s8}
As  per the specifications  and design considerations mentioned in  Section \ref{s1}  and Section \ref{s3}, the basic requirements for the interceptor UAV along with manipulator is listed in Table \ref{tab:hardware}.  Apart from these, the ball grabbing UAV has to be stable during the grabbing process, as it needs to exert a force of 4 N for the detachment of the ball. Also, the ball grabbing UAV has to withstand the downwash from the target UAV during the grabbing process. \begin{table}
\centering
\caption{Design specifications}
\begin{tabular}{|l|c|}
	\hline
	Specifications & Requirements \\
	\hline\hline
	 Max size & 1.2 m $\times$ 1.2 m $\times$ 0.5 m \\\hline
  	 Endurance & At least 15 minutes \\\hline
 	 Manipulator location & Sideways \\\hline
 	 Gripper & Fixed to manipulator \\\hline
    Payload requirement & At least 1-2 kg\\
	\hline
\end{tabular}
\label{tab:hardware}
\end{table}
To satisfy the above specifications, the DJI M600 Pro is selected. It is a hexacopter with triple-redundant GPS and 18 minutes of flight time for 5 kg payload. The usage of the hexacopter aids better stability while grabbing the ball, where quick maneuvers are desired. The different equipment used for Challenge 1 are as below. 
\begin{enumerate}[label=(\alph*)]
     \item \textbf{UAV and on-board computers:} As mentioned above, DJI M600 Pro (Fig. \ref{fig:m600}) is used as the UAV for grabbing and popping. The UAV is commanded via the DJI A3 Pro flight controller, which is a highly reliable autopilot. The companion computer is NVIDIA Jetson TX2 (Fig. \ref{fig:tx2}). This computer is mounted on an Auvidea J130-2k4k carrier board. This is a well-equipped carrier board with four USB ports, 1 GB/s ethernet port, and HDMI input and output. This helps accommodate all the sensors on-board with ease. An Arduino Mega 2560 is used to control the manipulator's arm and an Arduino Nano for the propeller guards.
\begin{figure}[t]
\centering
\begin{subfigure}{0.24\columnwidth}
    \centering
    \includegraphics[scale=0.235]{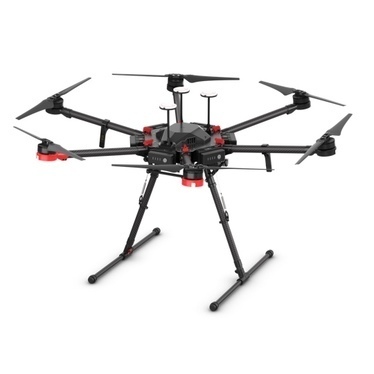}
    \subcaption{}
    \label{fig:m600}
\end{subfigure}
\begin{subfigure}{0.24\columnwidth}
    \centering
    \includegraphics[scale=0.295]{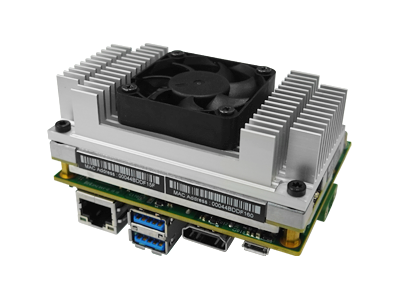}
    \subcaption{}
    \label{fig:tx2}
\end{subfigure}
\begin{subfigure}{0.24\columnwidth}
    \centering
    \includegraphics[scale=0.2625]{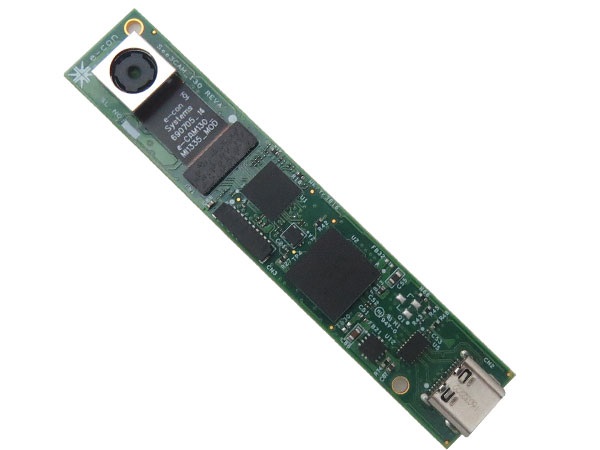}
    \subcaption{}
    \label{fig:see3}
\end{subfigure}
\begin{subfigure}{0.245\columnwidth}
    \centering
    \includegraphics[scale=0.115]{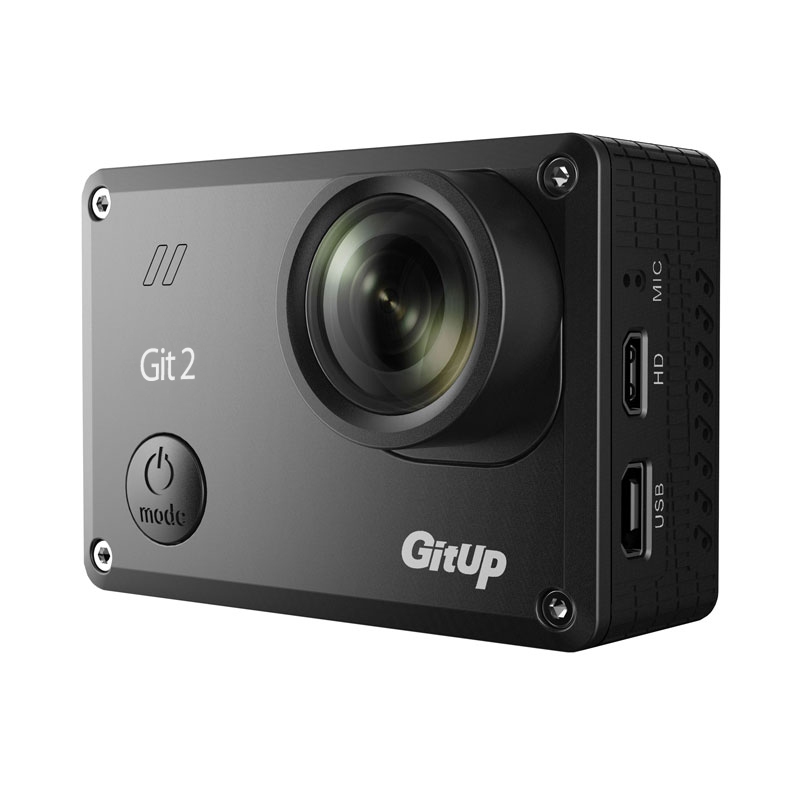}
    \subcaption{}
    \label{fig:gitup}
\end{subfigure}
\caption{(a) DJI M600 hexacopter (b) NVIDIA Jetson TX2 with Auvidea J130 carrier board (c) See3 130 camera (d) Git2P action camera.}
\label{f3}
\end{figure}
 
    \item \textbf{Vision sensors:} The target detection is achieved using visual feedback. The vision sensor employed for target tracking and interception is a See3CAM 130 (Fig. \ref{fig:see3}), an industrial-grade  4K USB monocular camera. This camera's choice helped in better detection of the ball from long distances and for a variety of backgrounds. This camera has an added advantage of smaller weight and size, which is essential for it to be mounted on the manipulator. The exploration was carried out using Gitup Git2 Pro (Fig. \ref{fig:gitup}). This is an action camera with 170$^\text{o}$ field of view, which aids in better visibility of the target ball in the arena. A T-3D V metal 3-Axis brushless gimbal is attached to the front of the UAV to hold the Gitup camera in place. 
 
\item \textbf{Manipulation mechanism and communication system:} An entirely in-house designed and fabricated one degree of freedom manipulator is employed for the challenge. Increasing the degrees of freedom provides a larger operational space for the end effector. However, this, in turn, brings in delay in actuation, more complexity in control of the manipulator, and increases power consumption. Therefore, a single degree of freedom manipulation mechanism was developed for the Challenge 1 tasks. 
 
\begin{figure}
    \centering
    \includegraphics[width=0.9\columnwidth, height=3.5cm]{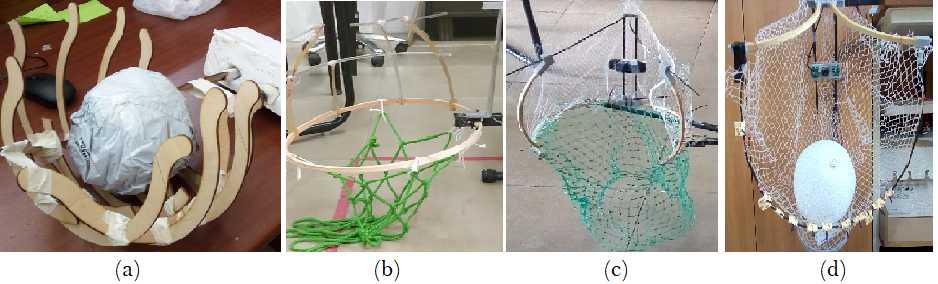}
    \caption{Evolution in design of the basket type manipulator (a) Active gripper (closes to detach) (b) Passive gripper (the crown detaches ball at any of the multiple detachment points) (c) Passive semi-open basket with dedicated camera mount (single detachment point at the center) (d) Multi-utility (grab/pop) passive semi-open basket with eye-in-hand configuration.}
    \label{f5}
\end{figure}
The design mechanism evolved as shown in Fig. \ref{f5}. The initial idea was that of an active manipulator (Fig. \ref{f5}a), a twin claw system to detach the ball, made of balsa wood. The upper portion of the gripper closes to detach the ball, and the lower portion opens to drop the ball. To make the mechanism energy optimal and lightweight, a passive basket was conceived of next (Fig. \ref{f5}b), which retained the finger like multiple detachment mechanism from the previous version in the form of acrylic but was remodeled and fixed to a wooden basket, like its crown. But the optimal placement of the camera and a clear field of view was difficult to obtain with this configuration. Besides, the front rim lowered the chances of successful grabbing. This led to the development of a newer design (Fig. \ref{f5}c), which adopted a single point of detachment mechanism rather than the many points of detachment from the previous design. This was also a semi-open basket with dedicated mounts for the camera (eye-in-hand). After extensive testing, the final design was arrived at (Fig. \ref{f5}d), where the gripper was modified such that it could grab the ball as well as pop the balloon. Sharp needles were attached to the lower portion of the basket to aid popping while the top portion of the gripper executed the ball detachment. The detachment portion is made of lightweight wood, while the lower portion of the gripper is made of thin carbon fiber strips. The basket net for collecting the ball for the final design is made from a fishing line. This made the setup lightweight and also offered less drag. 
 
\begin{figure}
\centering
\begin{subfigure}{0.24\columnwidth}
    \centering
    \includegraphics[scale=0.3]{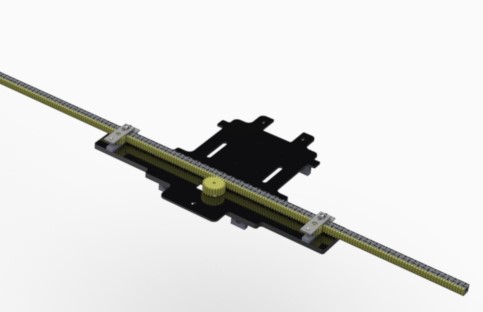}
    \vspace{0.15cm}
    \subcaption{}
    \label{fig:rack}
\end{subfigure}
\begin{subfigure}{0.24\columnwidth}
    \centering
    \includegraphics[scale=0.245]{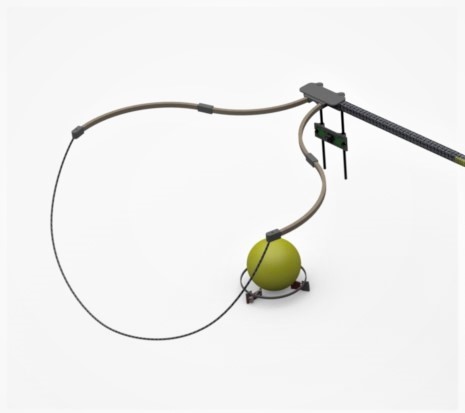}
    \subcaption{}
    \label{fig:grabber}
\end{subfigure}
\begin{subfigure}{0.24\columnwidth}
    \centering
    \includegraphics[scale=0.185]{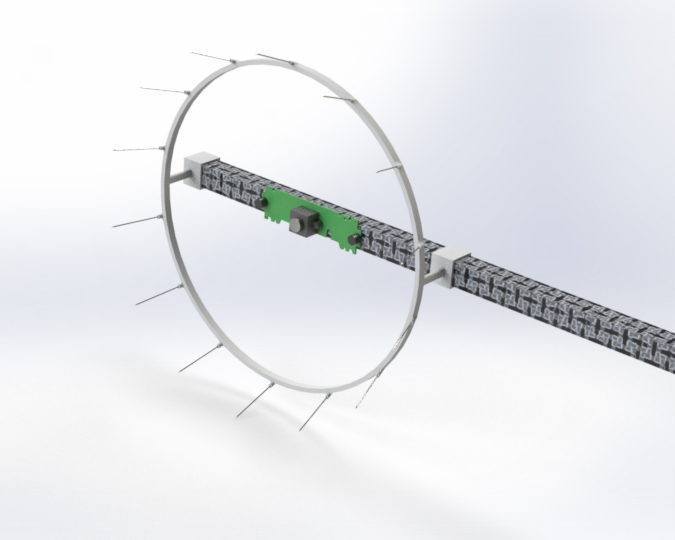}
    \subcaption{}
    \label{fig:popper}
\end{subfigure}
\begin{subfigure}{0.24\columnwidth}
    \centering
    \vspace{0.3cm}
    \includegraphics[scale=0.175]{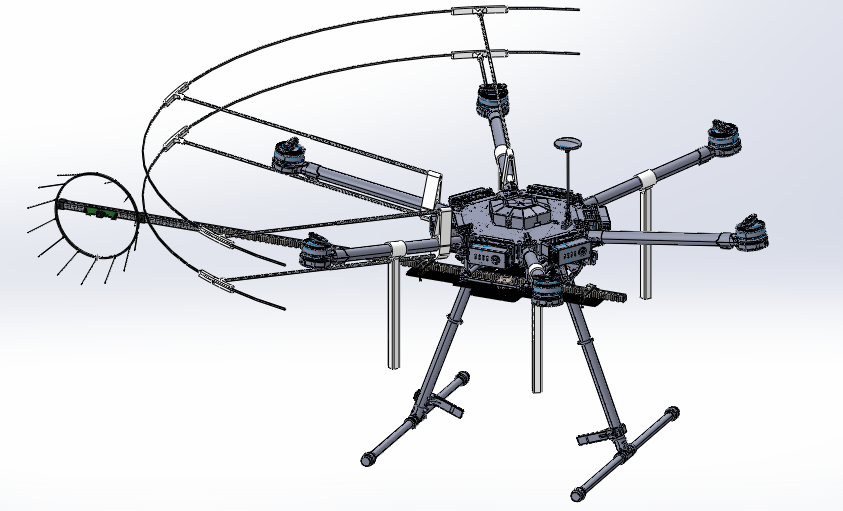}
    \subcaption{}
    \label{fig:mtotal}
\end{subfigure}
    \caption{The relevant design components of the manipulation mechanism are (a)  The rack and pinion mechanism (linear actuation) with its mount (b) The end effector to grab ball (c) The end effector for popping balloons (d) The manipulator along with propeller guards.}
\label{f4}
 \end{figure}
The manipulator used for the competition was fabricated using carbon fiber, which provides better strength to weight ratio. To be within the size constraint, the manipulator is kept retracted under the UAV and actuated linearly. The in-flight extension of the arm is achieved using a rack and pinion mechanism (Fig. \ref{fig:rack}). The range of extension is calculated to ensure stability and zero moments on the overall system while executing maneuvers. The end effector of this manipulator is a passive basket with a fruit-picker-like mechanism to detach the ball (Fig. \ref{fig:grabber}). A lightweight net is attached to the rim of this basket to collect the ball. Limit switches are attached to the bottom of this passive basket, as seen in (Fig. \ref{fig:grabber}). This acts as the feedback mechanism for detecting successful grabbing of the ball. The camera is fixed to the center of the basket, which results in an eye-in-hand configuration. This is especially useful in the approach and grab phases of the challenge. 
 
For popping balloons, a dedicated mechanism is employed. The manipulator's arm works similar to the ball grabber while the end effector has a structure shown in (Fig. \ref{fig:popper}). Sharp needles are arranged on a mount, which is 3D printed using PLA. Here too, an eye-in-hand configuration is used to help pop the balloon. 
 
Propeller guards (Fig. \ref{fig:mtotal}) are incorporated to ensure safety while carrying out the tasks. The propeller guard is designed to extend in-flight. Suitable 3D printed mounts are attached to the UAV arms to support the extension of the guards after take-off. The manipulator actuation and propeller guard extensions are carried out using an on-board Arduino Mega board. A detailed account of manipulator design and development is given in \cite{vidyadhara2021design,bv2021design}.

5GHz Wi-Fi set up is installed for field experiments, as it has better bandwidth than the 2.4 GHz counterpart. Jetson TX2 can connect to 2.4 GHz and a lower band of 5 GHz Wi-Fi. For testing purposes, TL-MR3020, a portable 3G/4G DC wireless network router, set to a static channel is used. The ground control station is an Alienware m15 R1 laptop. Its GPU is utilized for training the vision algorithms as well. 
 \end{enumerate}

\section{Experimental Results}\label{s9}
The following section details the flight experiments conducted at the IISc airfield and those during the competition. The results achieved and the major observations are reported in this section. As per the earlier competition rules, the color of the target ball was red, and the balloon was white. Therefore, some of the field experiences are performed with a red ball and white balloons.

\subsection{Field experiments}
\begin{figure}[htb!]
    \centering
    \begin{subfigure}{0.45\columnwidth}
        \centering
        \includegraphics[scale=0.4]{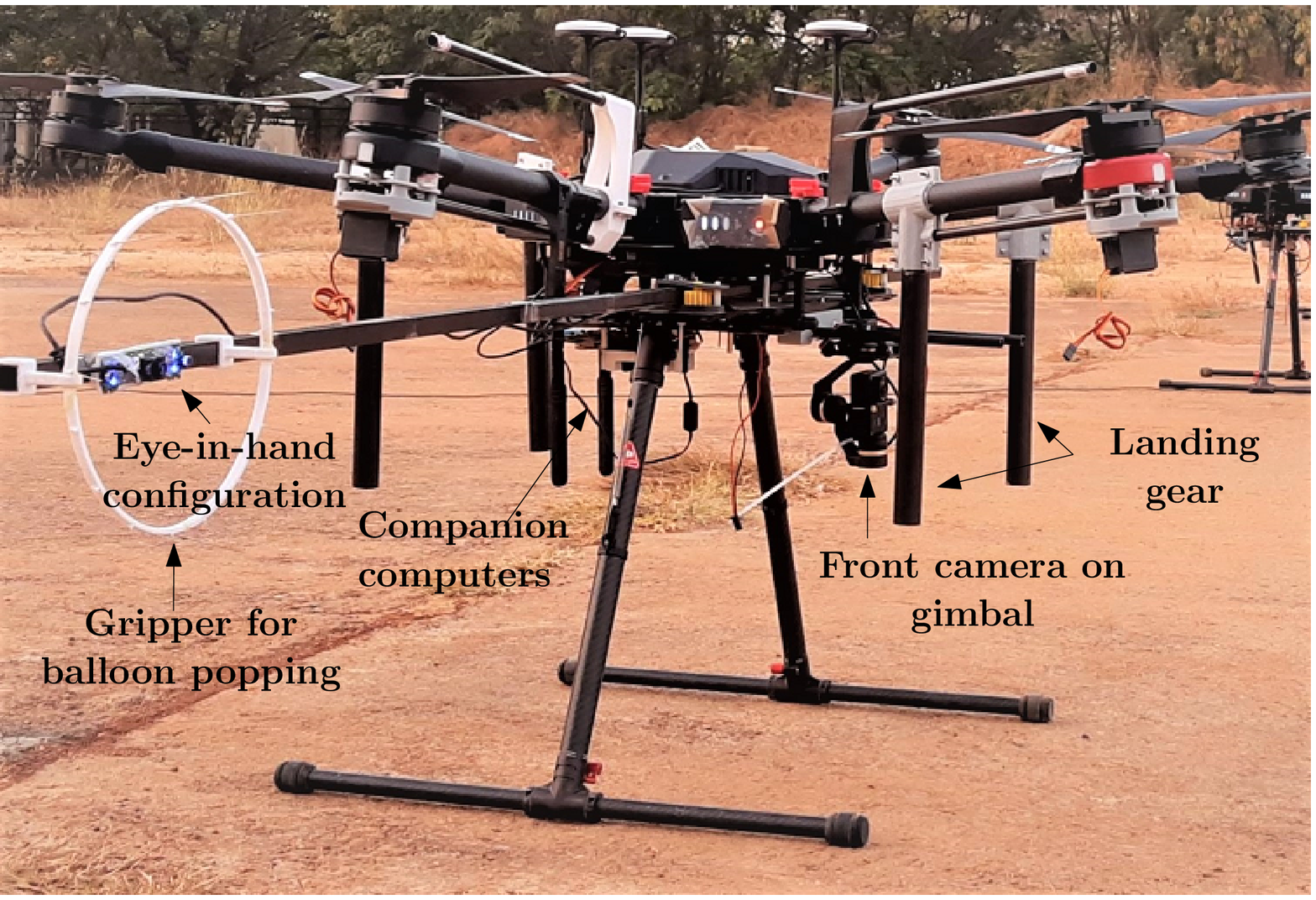}
        \subcaption{}
    \end{subfigure}
    \begin{subfigure}{0.45\columnwidth}
        \centering
        \includegraphics[scale=0.5]{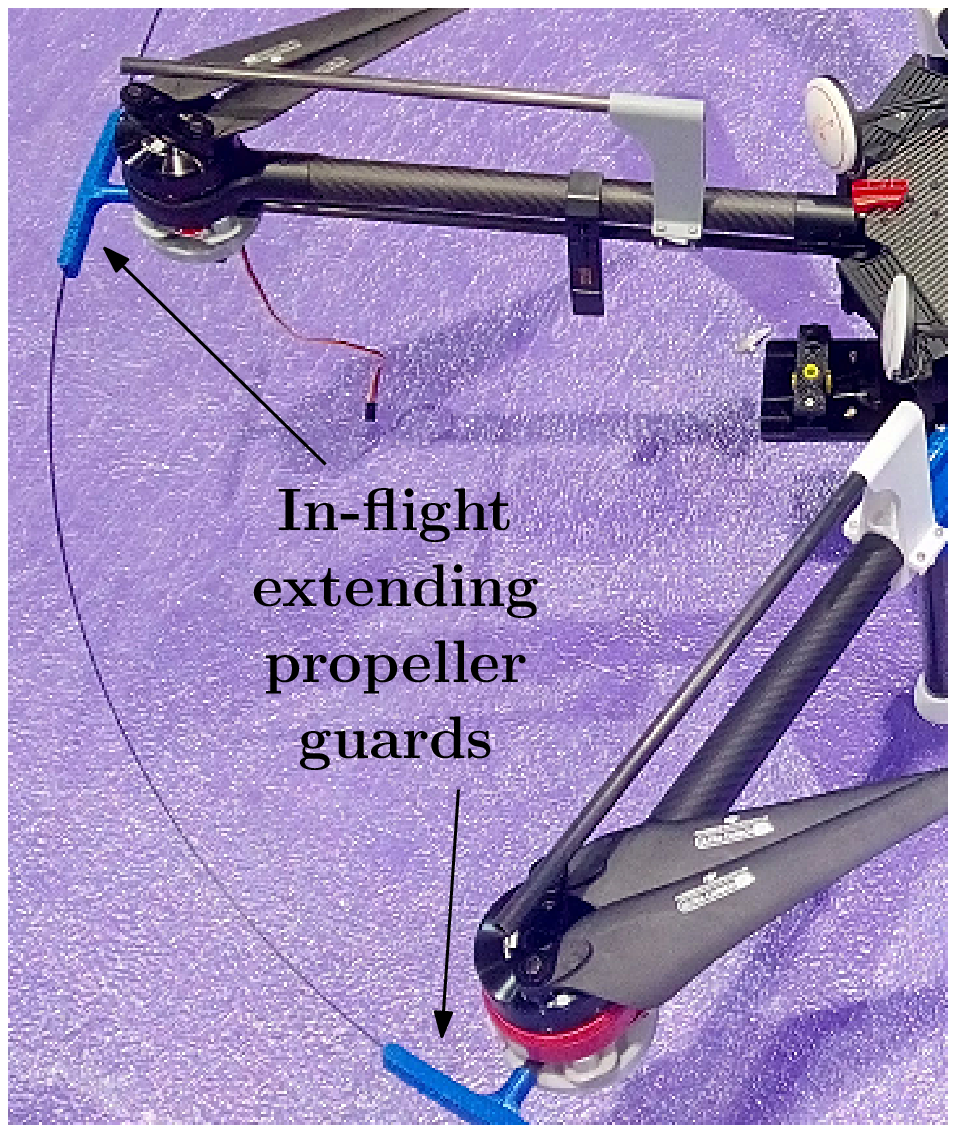}
        \subcaption{}
    \end{subfigure}
    \begin{subfigure}{1\columnwidth}
        \centering
        \includegraphics[width=0.8\columnwidth, height=4cm]{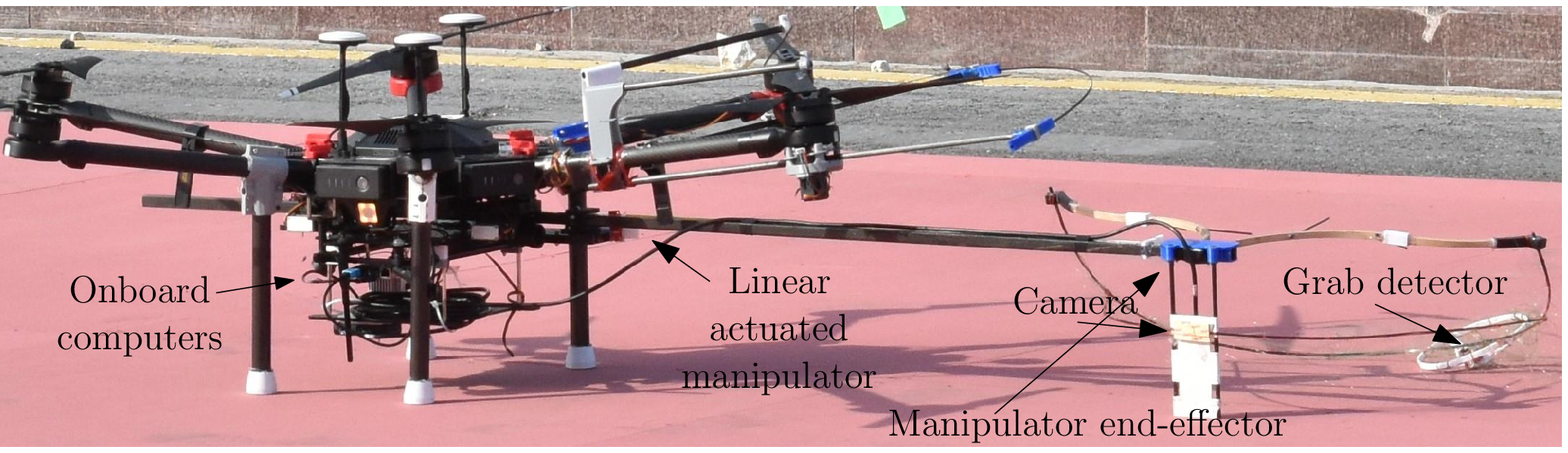}
        \subcaption{}
    \end{subfigure}
    \caption{(a) UAV set-up for balloon popping (b) The in-flight extendable propeller guard (c) The grabber UAV with passive basket type single DoF manipulator.}
    \label{f6}
\end{figure}

Simulations were performed using the PX4 autopilot in the Gazebo environment. The DJI based system was used for the competition and ROS-Gazebo was used to simulate the OMS and further deploy it on hardware. By doing so, the OMS developed on the simulator can be seamlessly transferred to the DJI environment with little to no modifications. The final configuration of the balloon popping UAV and ball grabbing UAV are shown in  Fig. \ref{f6}a and Fig. \ref{f6}c. The interceptor UAVs are equipped with propeller guards (Fig. \ref{f6}b).
\begin{figure}[htb!]
    \begin{subfigure}{0.34\columnwidth}
        \centering
        \includegraphics[scale=0.3]{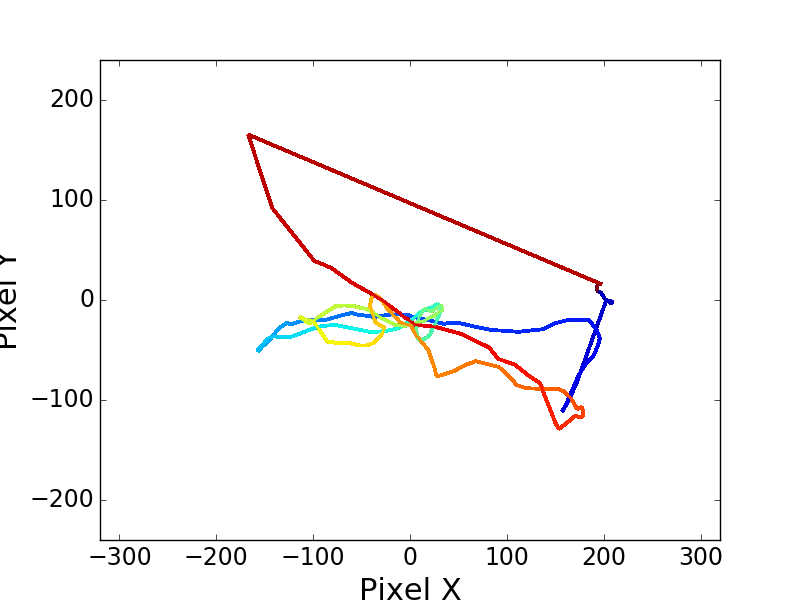}
        \caption{}
        \label{fig:pixel_xy}
    \end{subfigure}
    \begin{subfigure}{0.34\columnwidth}
        \centering
        \includegraphics[scale=0.3]{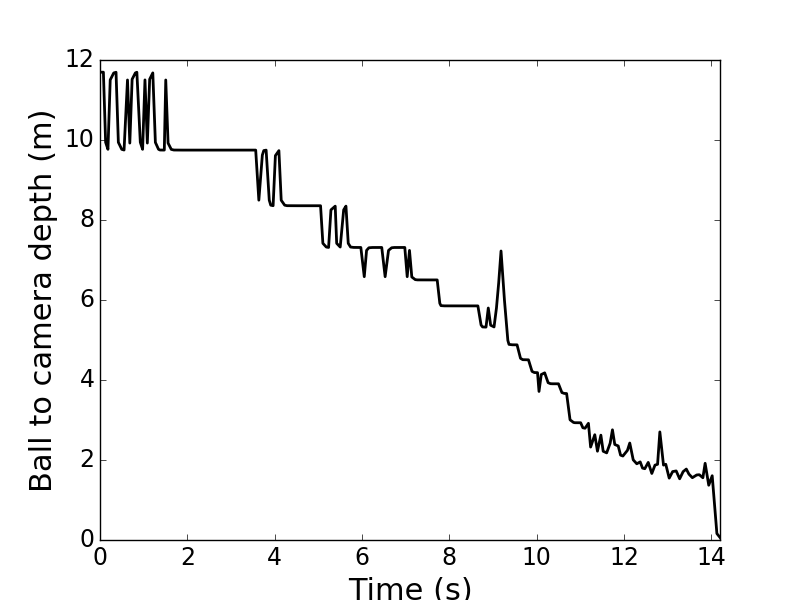}
        \caption{}
        \label{fig:depth}
    \end{subfigure}
    \begin{subfigure}{0.33\columnwidth}
        \centering
        \includegraphics[scale=0.3]{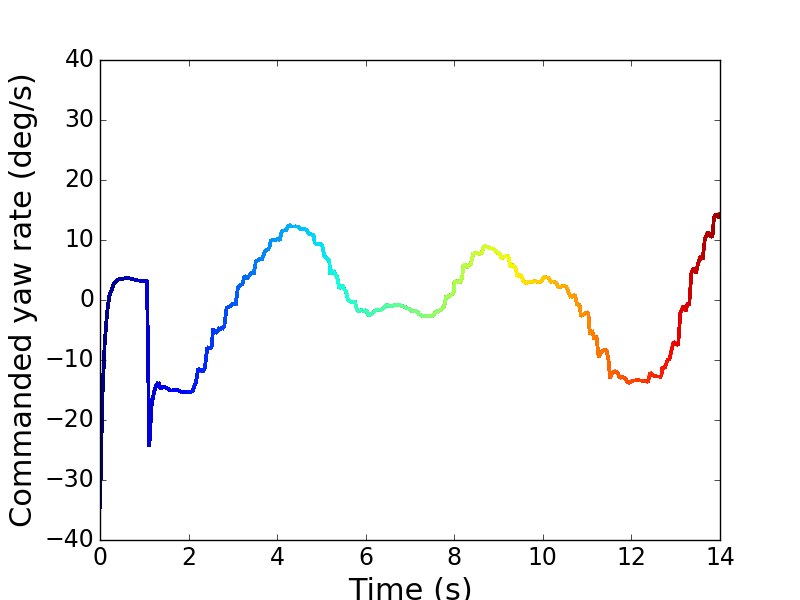}
        \caption{}
        \label{fig:comm_color_yaw_rate}
    \end{subfigure}
 \begin{subfigure}{0.246\columnwidth}
     \centering
     \includegraphics[scale=0.225]{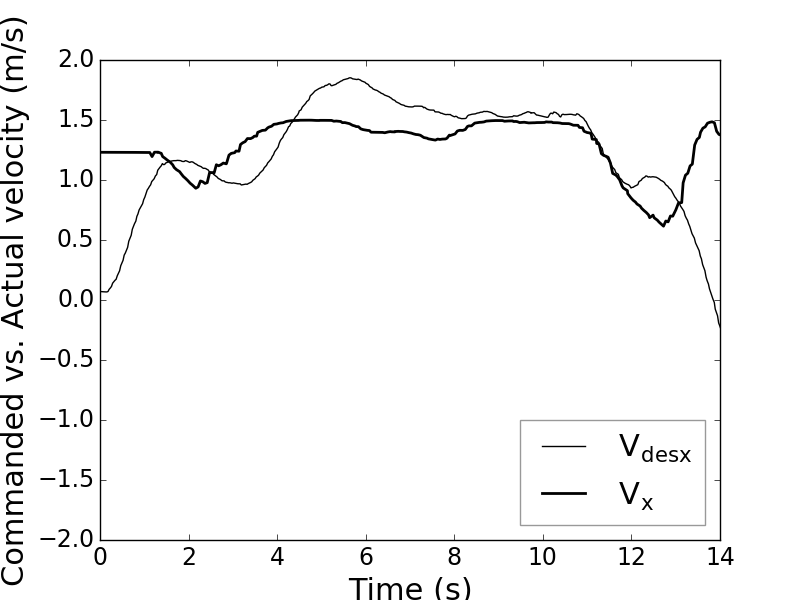}
     \caption{}
     \label{fig:comm_vs_act_x}
 \end{subfigure}
 \begin{subfigure}{0.246\columnwidth}
     \centering
     \includegraphics[scale=0.225]{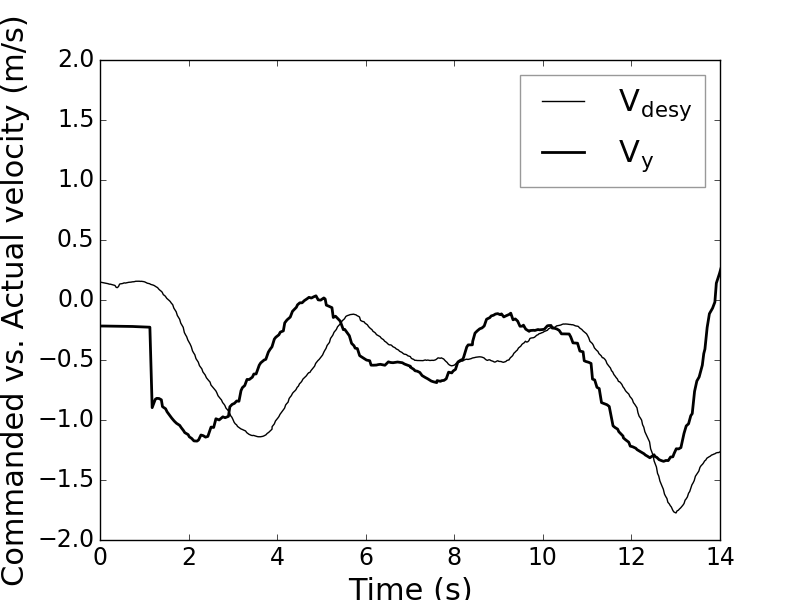}
     \caption{}
     \label{fig:comm_vs_act_y}
 \end{subfigure}
 \begin{subfigure}{0.246\columnwidth}
     \centering
     \includegraphics[scale=0.225]{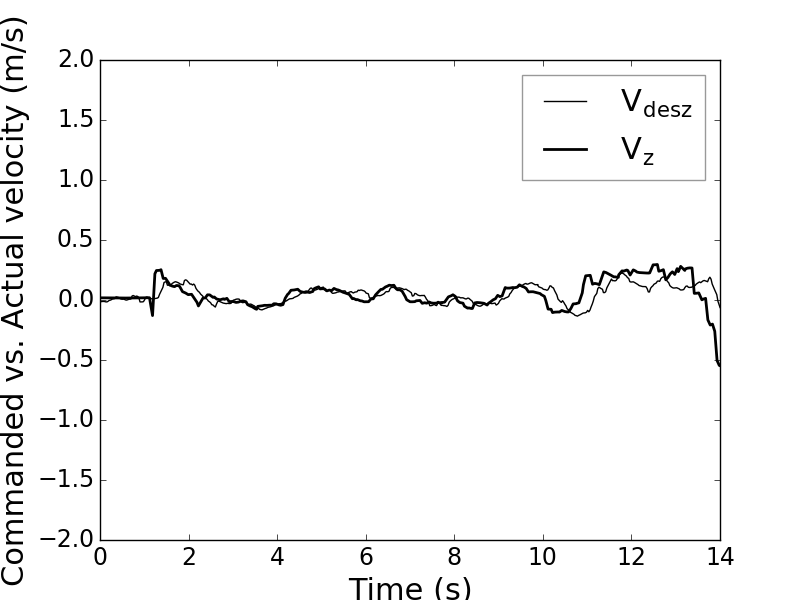}
     \caption{}
     \label{fig:comm_vs_act_z}
 \end{subfigure}
\begin{subfigure}{0.246\columnwidth}
     \centering
     \includegraphics[scale=0.225]{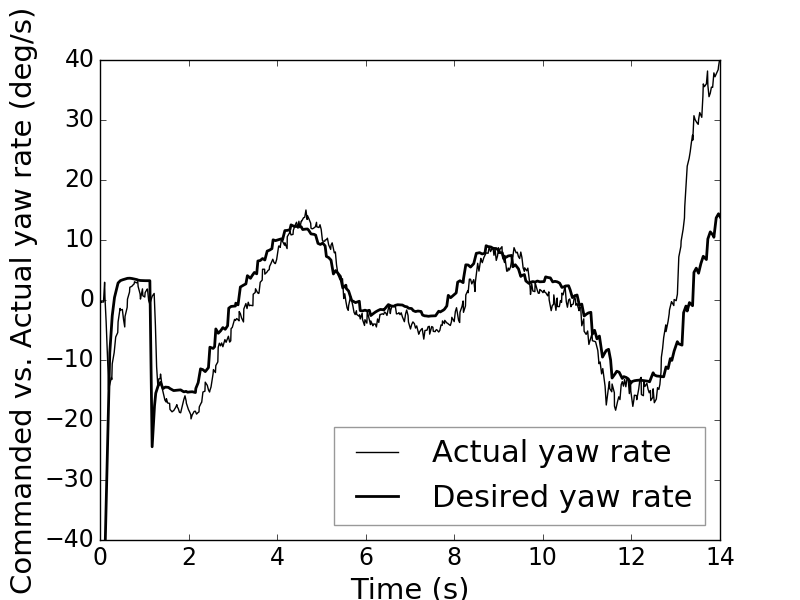}
     \caption{}
     \label{fig:comm_vs_act_yaw}
 \end{subfigure}
 \begin{subfigure}{0.33\columnwidth}
        \centering
        \includegraphics[scale=0.3]{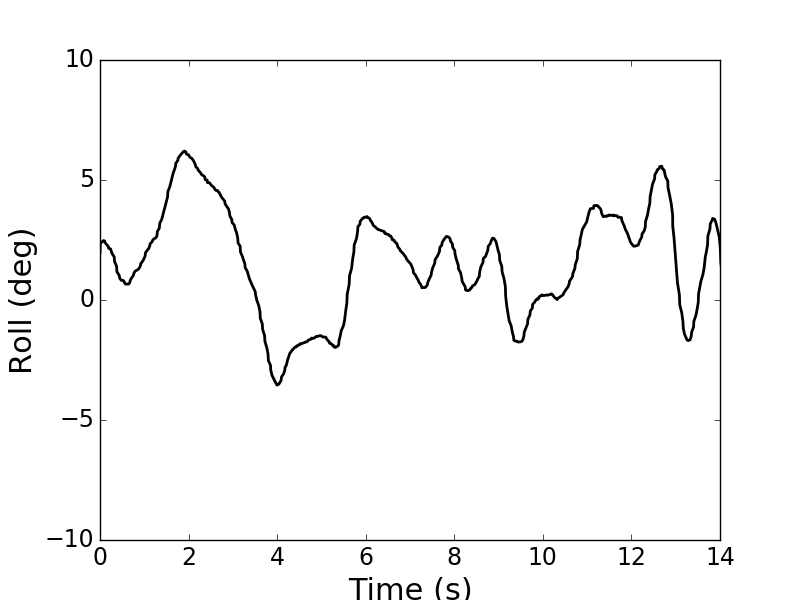}
        \caption{}
        \label{fig:roll}
    \end{subfigure}
    \begin{subfigure}{0.33\columnwidth}
        \centering
        \includegraphics[scale=0.3]{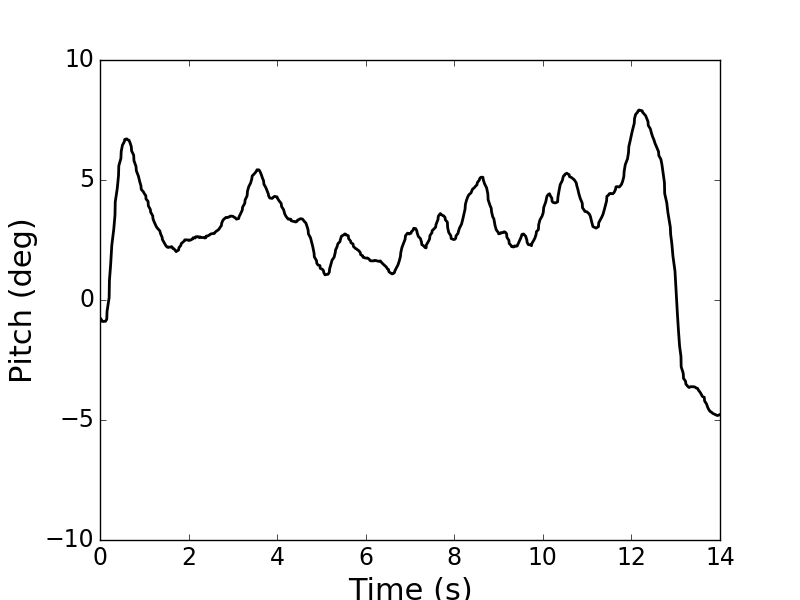}
        \caption{}
        \label{fig:pitch}
    \end{subfigure}
    \begin{subfigure}{0.33\columnwidth}
        \centering
        \includegraphics[scale=0.3]{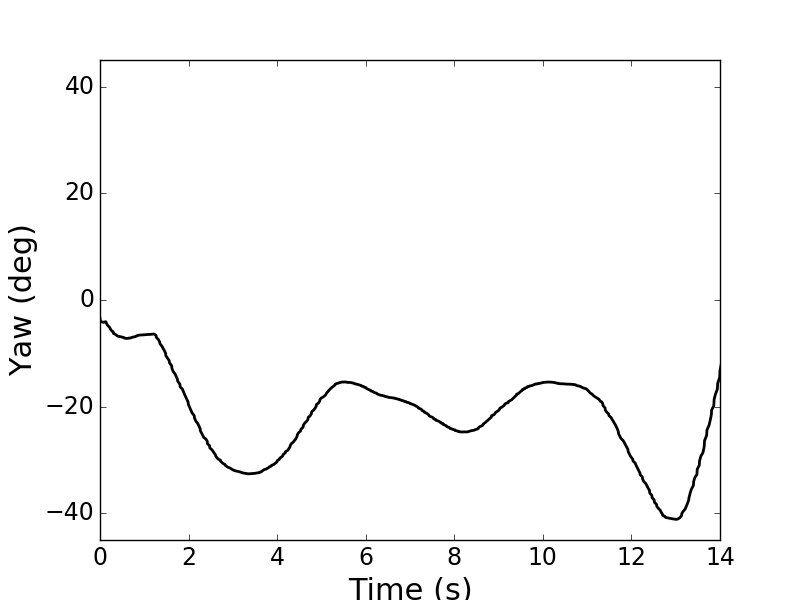}
        \caption{}
        \label{fig:yaw}
    \end{subfigure}
    \begin{subfigure}{0.33\columnwidth}
        \centering
        \includegraphics[scale=0.3]{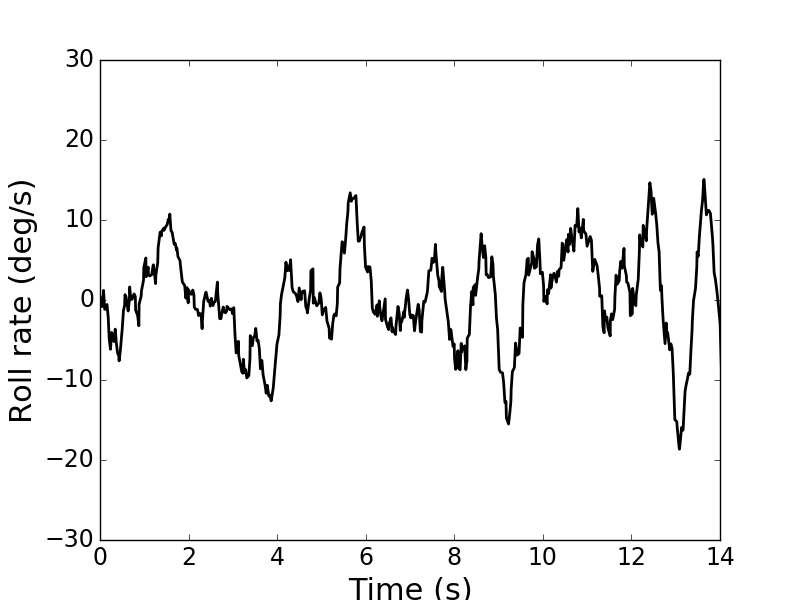}
        \caption{ }
        \label{fig:roll_rate}
    \end{subfigure}
    \begin{subfigure}{0.33\columnwidth}
        \centering
        \includegraphics[scale=0.3]{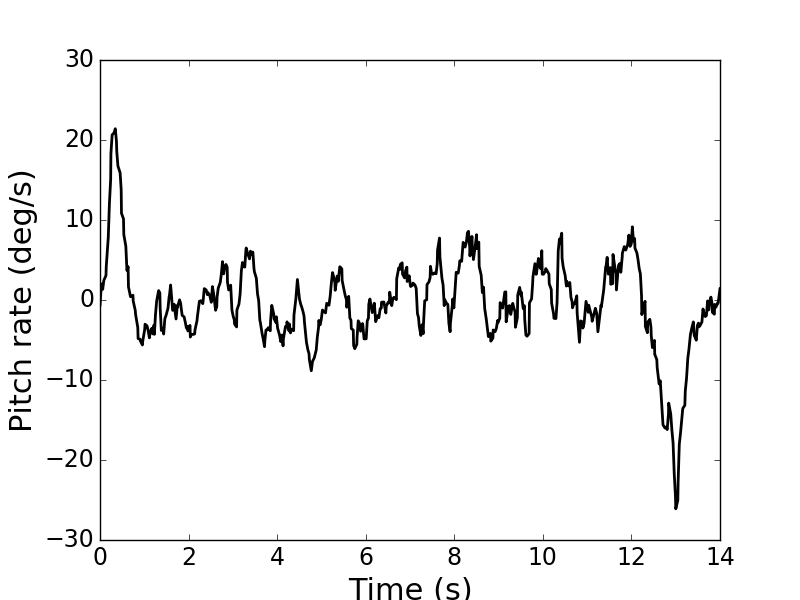}
        \caption{}
        \label{fig:pitch_rate}
    \end{subfigure}
    \begin{subfigure}{0.33\columnwidth}
        \centering
        \includegraphics[scale=0.3]{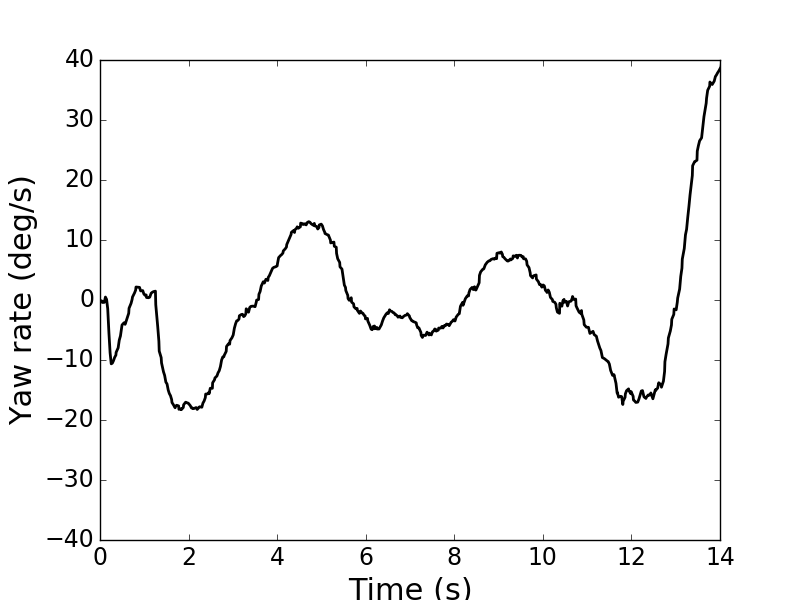}
        \caption{}
        \label{fig:yaw_rate}
    \end{subfigure}
    \caption{Experiment 1 (a) Variation of target pixel in  camera frame (b) Variation of ball depth from camera during grabbing  (c) Commanded yaw rate during grabbing phase. Commanded vs. actual velocity in (d) $x$ direction (e) $y$ direction (f) $z$ direction (g) Commanded vs. actual yaw rate (h) Roll angle (i) Pitch angle (j) Yaw angle of the interceptor UAV while in grabbing mission (k) Roll rate (l) Pitch rate (m) Yaw rate of the grabber UAV. The rates confirm stable engagement.}
    \label{fig:pixel_plot}
\end{figure}
 
\begin{figure}[t]
    \centering
    \includegraphics[width=1\columnwidth, height=5.5cm]{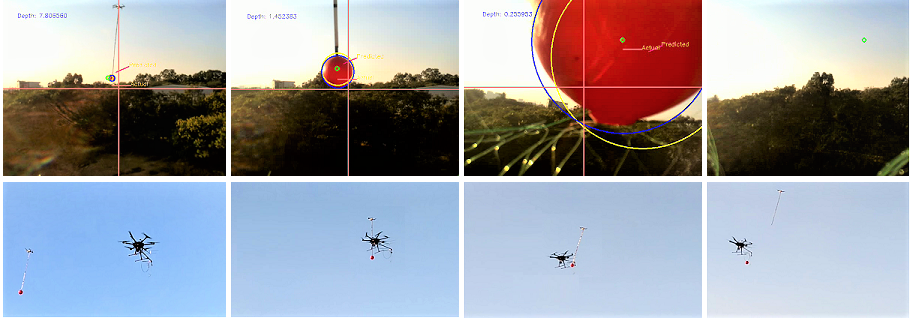}
    \caption{The ball grabbing experiment, visualised frame by frame, from the UAV camera (top) and a third person (bottom). From left: ball being detected, approached, intercepted and detached. The initial experiments were carried out using a red ball. }
    \label{fig:ball_grabbing}
\end{figure}
\begin{figure}[t]
    \centering
    \includegraphics[width=0.8\columnwidth, height=5.5cm]{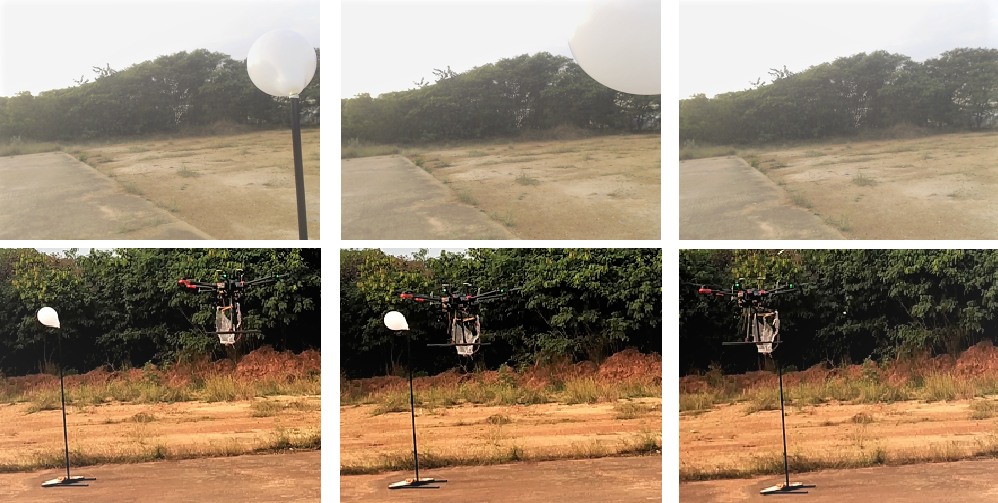}
    \caption{Balloon popping experiment from the UAV camera (top) and third person (bottom). From left: detection, approaching and popping. The experiments were carried out using a white balloon.}
    \label{fig:balloon_popping}
\end{figure}

\begin{figure}[htb!]
    \begin{subfigure}{0.34\columnwidth}
        \centering
        \includegraphics[scale=0.3]{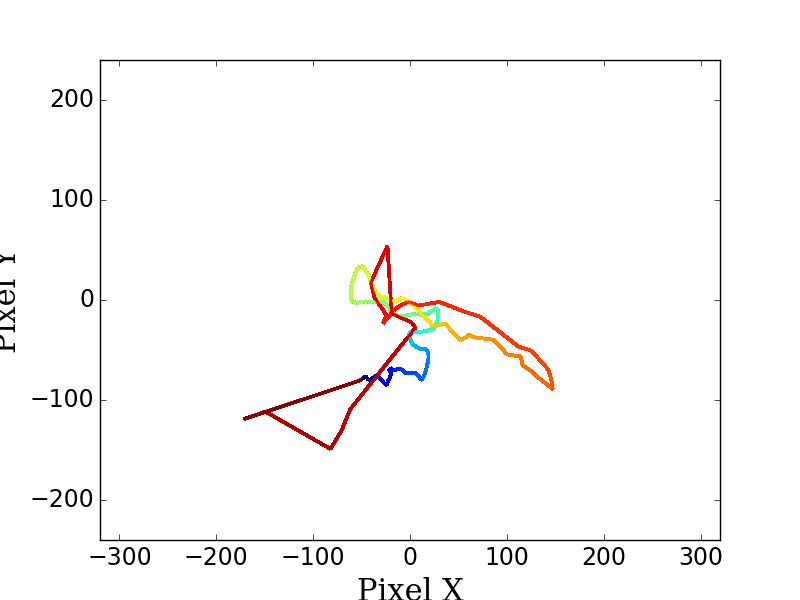}
        \caption{}
    \end{subfigure}
    \begin{subfigure}{0.34\columnwidth}
        \centering
        \includegraphics[scale=0.3]{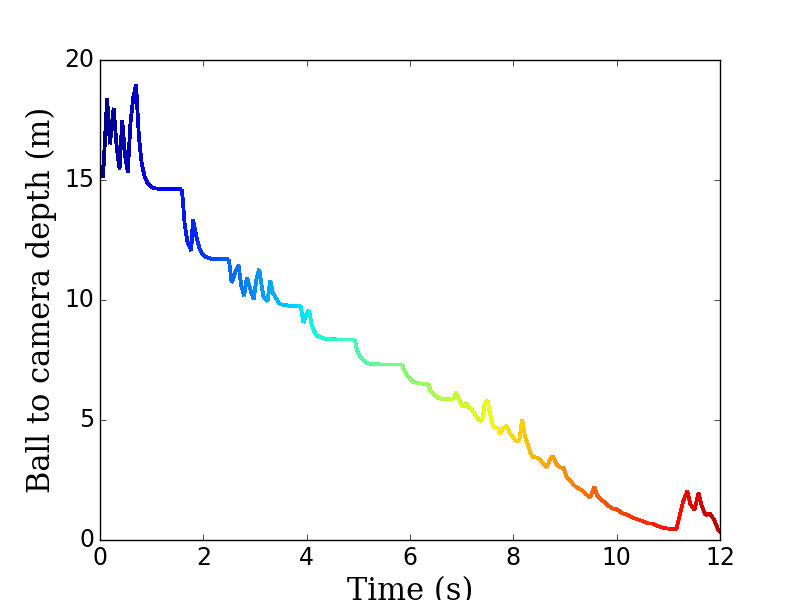}
        \caption{}
    \end{subfigure}
    \begin{subfigure}{0.33\columnwidth}
        \centering
        \includegraphics[scale=0.3]{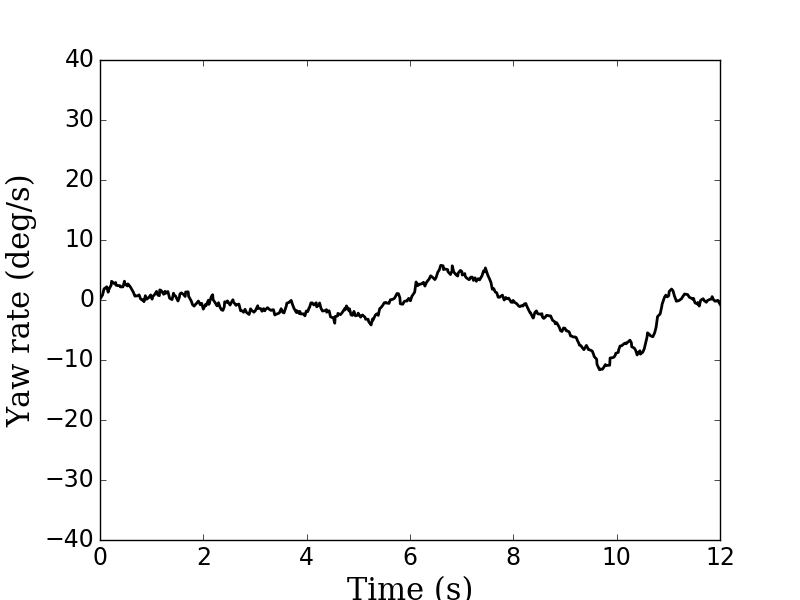}
        \caption{}
    \end{subfigure}
 \begin{subfigure}{0.246\columnwidth}
     \centering
     \includegraphics[scale=0.225]{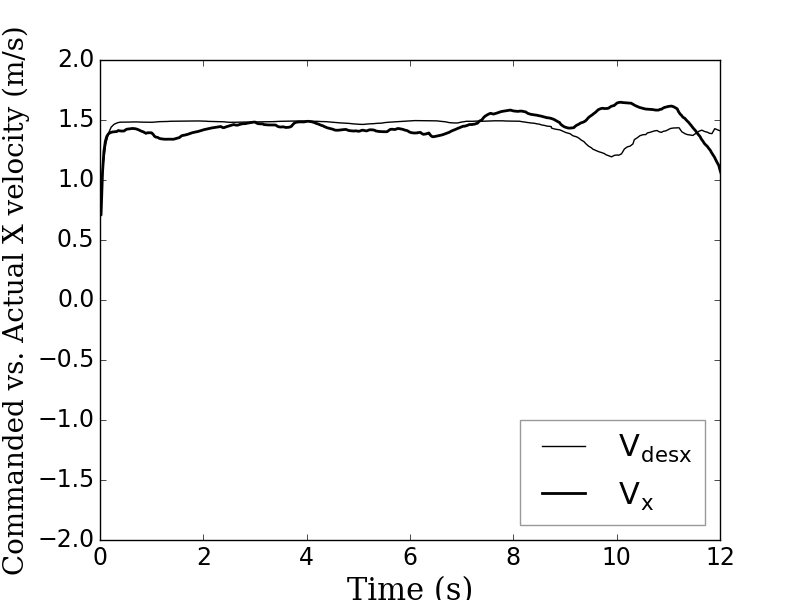}
     \caption{}
 \end{subfigure}
 \begin{subfigure}{0.246\columnwidth}
     \centering
     \includegraphics[scale=0.225]{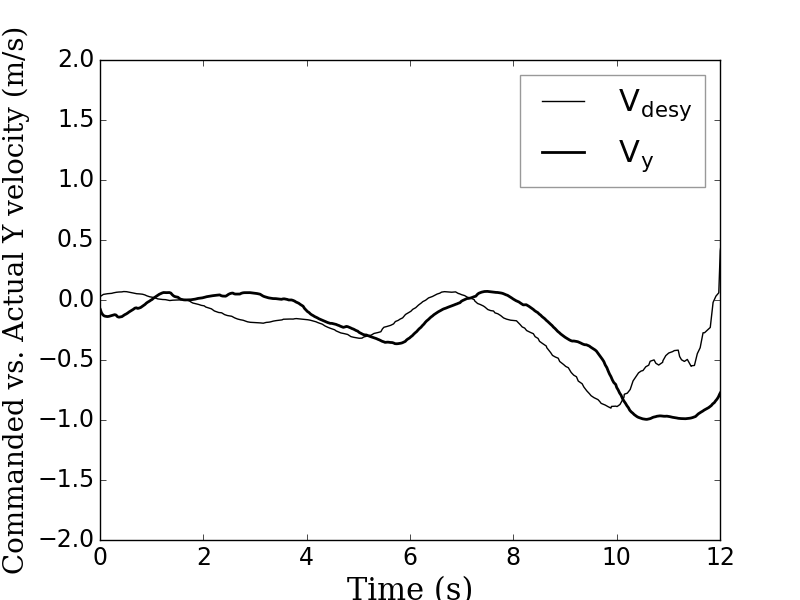}
     \caption{}
 \end{subfigure}
 \begin{subfigure}{0.246\columnwidth}
     \centering
     \includegraphics[scale=0.225]{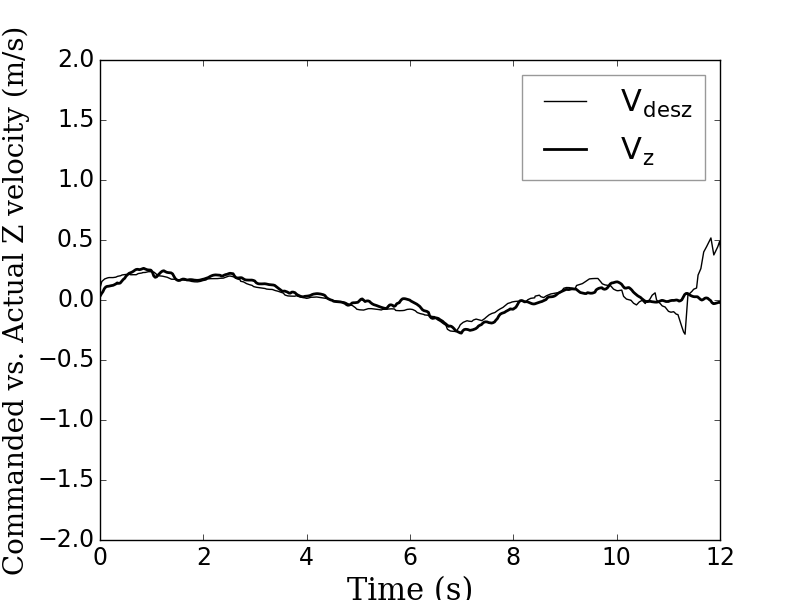}
     \caption{}
 \end{subfigure}
\begin{subfigure}{0.246\columnwidth}
     \centering
     \includegraphics[scale=0.225]{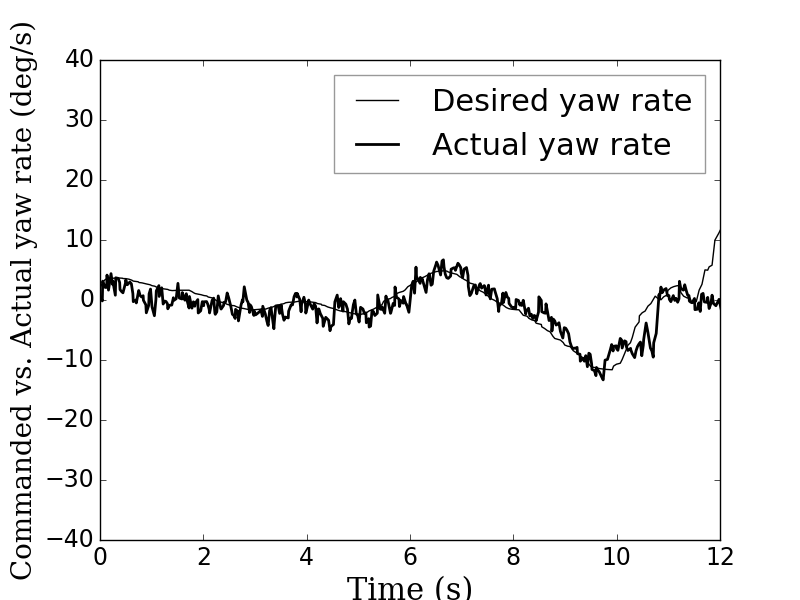}
     \caption{}
 \end{subfigure}
 \begin{subfigure}{0.33\columnwidth}
        \centering
        \includegraphics[scale=0.3]{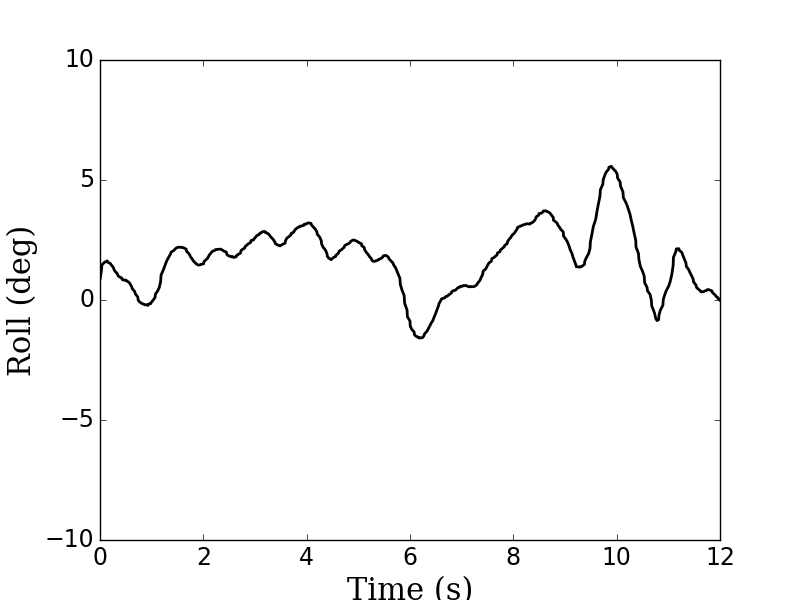}
        \caption{}
    \end{subfigure}
    \begin{subfigure}{0.33\columnwidth}
        \centering
        \includegraphics[scale=0.3]{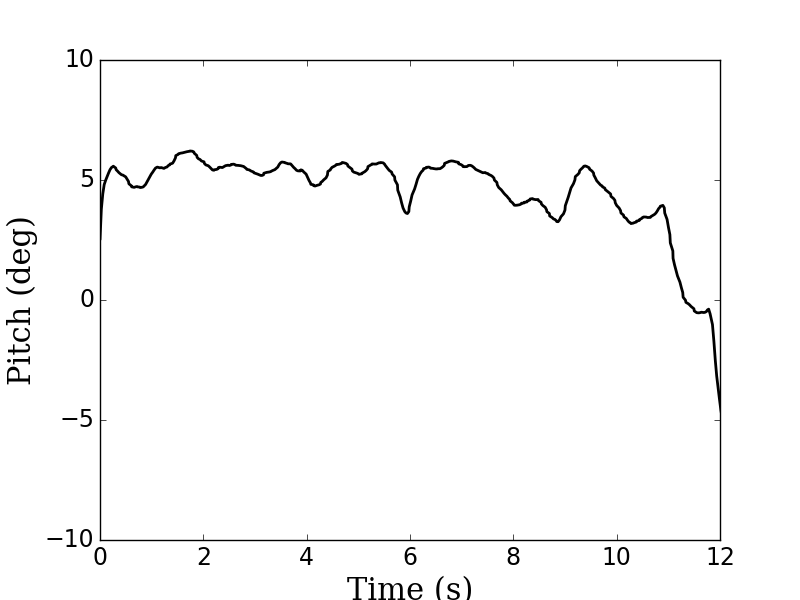}
        \caption{}
    \end{subfigure}
    \begin{subfigure}{0.33\columnwidth}
        \centering
        \includegraphics[scale=0.3]{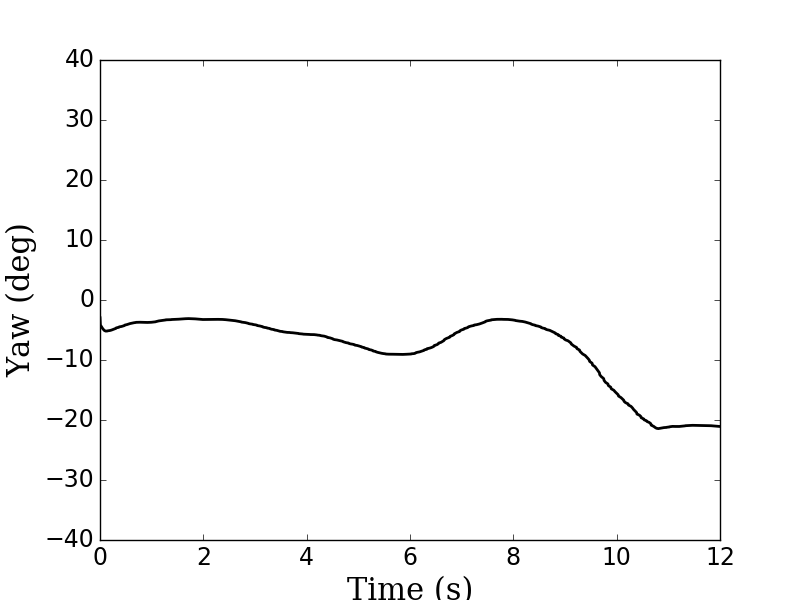}
        \caption{}
    \end{subfigure}
    \begin{subfigure}{0.33\columnwidth}
        \centering
        \includegraphics[scale=0.3]{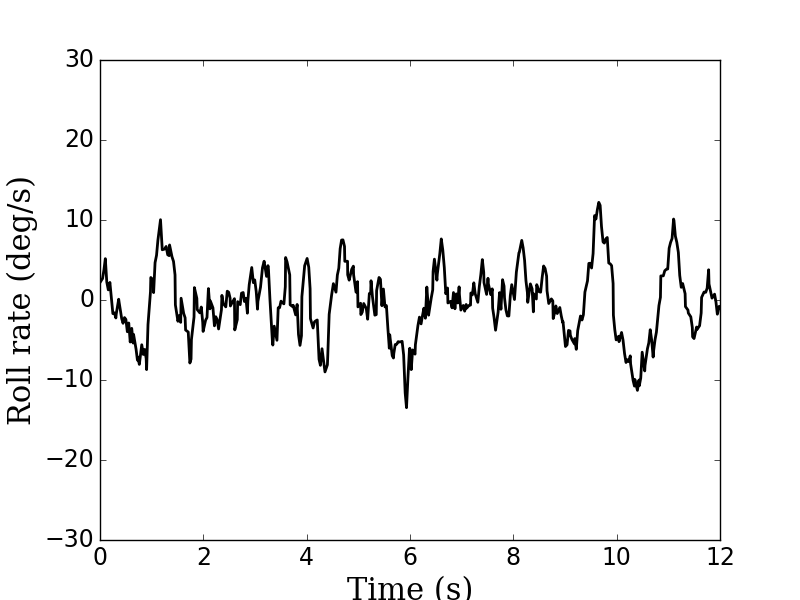}
        \caption{ }
    \end{subfigure}
    \begin{subfigure}{0.33\columnwidth}
        \centering
        \includegraphics[scale=0.3]{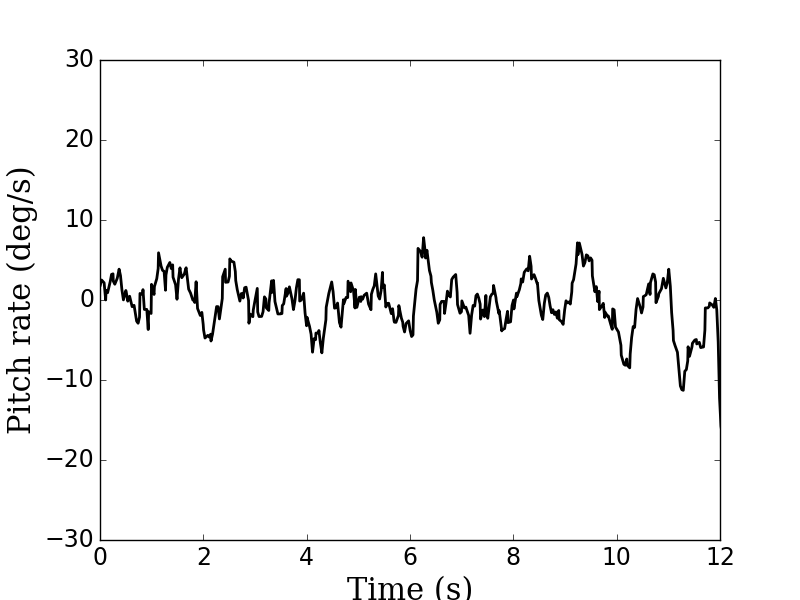}
        \caption{}
    \end{subfigure}
    \begin{subfigure}{0.33\columnwidth}
        \centering
        \includegraphics[scale=0.3]{figures/2nd grab plot/yaw_rate_1.png}
        \caption{}
    \end{subfigure}
    \caption{Experiment 2 (a) Variation of target pixel in  camera frame (b) Variation of ball depth from camera during grabbing  (c) Commanded yaw rate during grabbing phase. Commanded vs. actual velocity in (d) $x$ direction (e) $y$ direction (f) $z$ direction (g) Commanded vs. actual yaw rate (h) Roll angle (i) Pitch angle (j) Yaw angle of the interceptor UAV while in grabbing mission (k) Roll rate (l) Pitch rate (m) Yaw rate of the grabber UAV.}
    \label{fig:set2}
\end{figure}

\begin{figure}[htb!]
    \begin{subfigure}{0.34\columnwidth}
        \centering
        \includegraphics[scale=0.3]{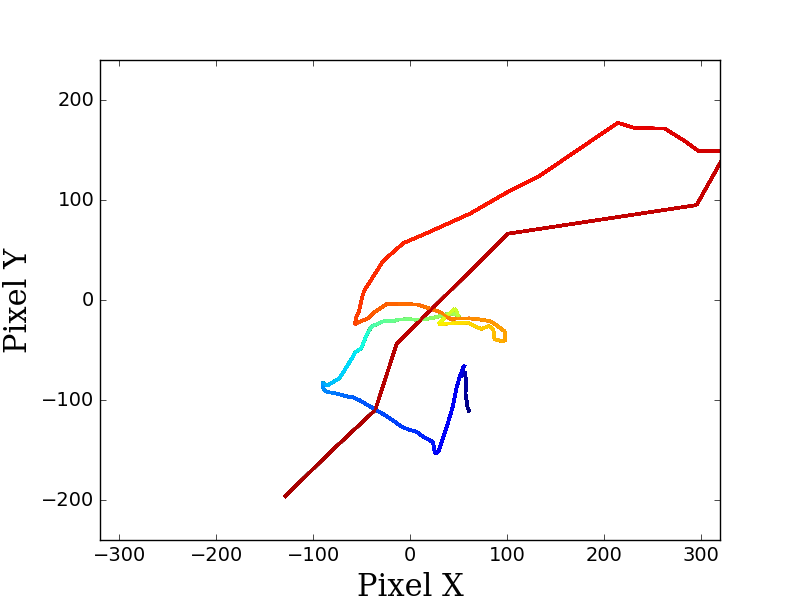}
        \caption{}
    \end{subfigure}
    \begin{subfigure}{0.34\columnwidth}
        \centering
        \includegraphics[scale=0.3]{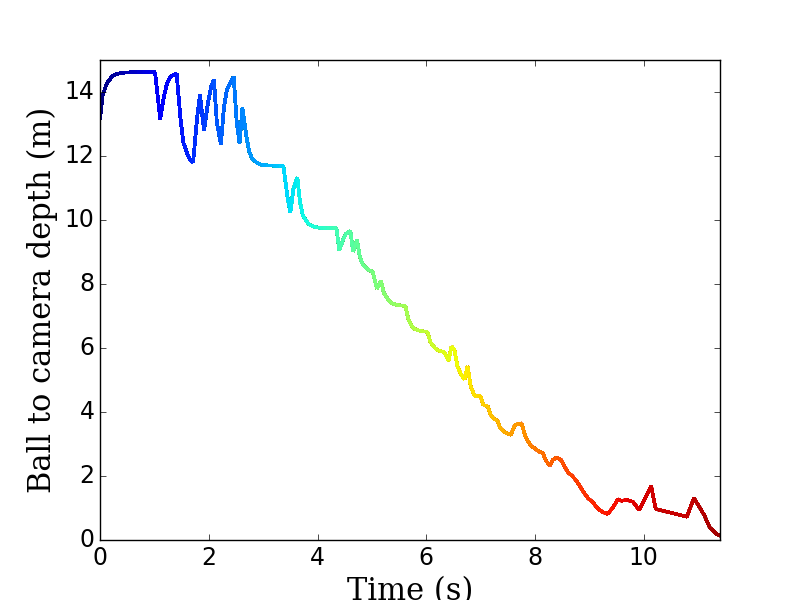}
        \caption{}
    \end{subfigure}
    \begin{subfigure}{0.33\columnwidth}
        \centering
        \includegraphics[scale=0.3]{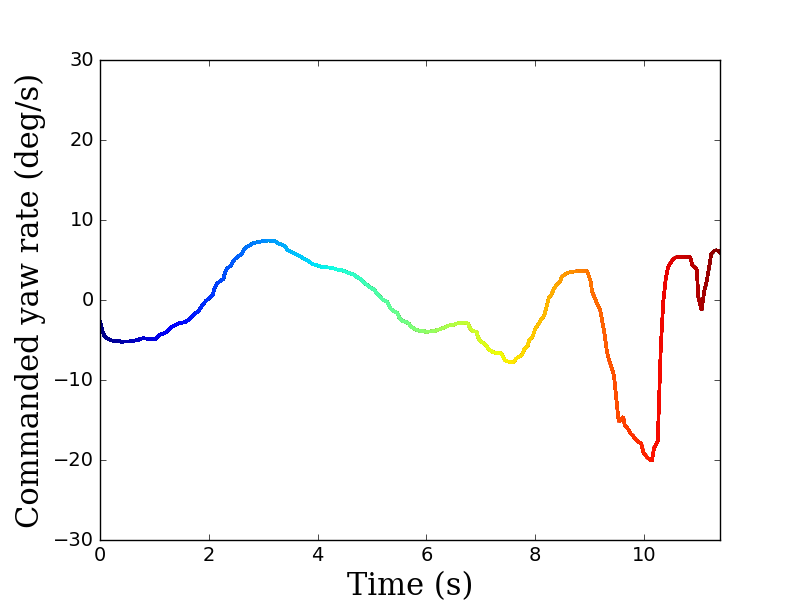}
        \caption{}
    \end{subfigure}
 \begin{subfigure}{0.246\columnwidth}
     \centering
     \includegraphics[scale=0.225]{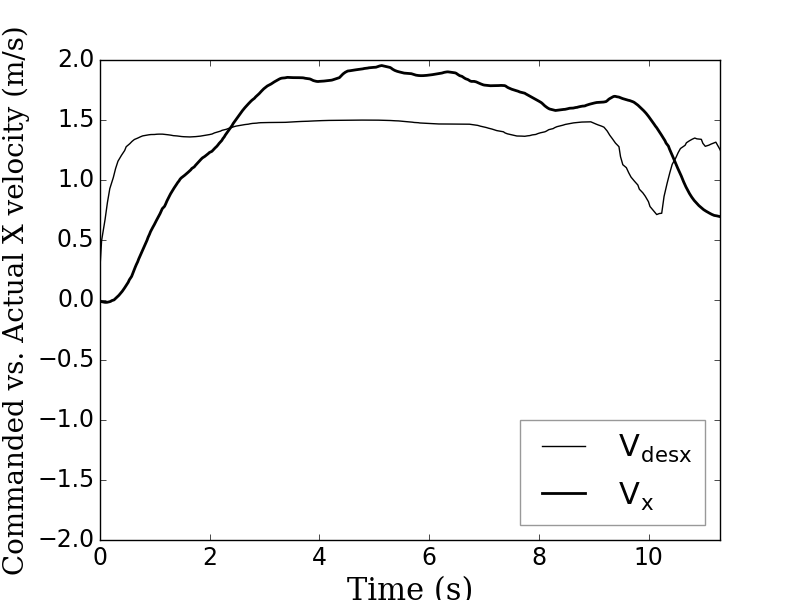}
     \caption{}
 \end{subfigure}
 \begin{subfigure}{0.246\columnwidth}
     \centering
     \includegraphics[scale=0.225]{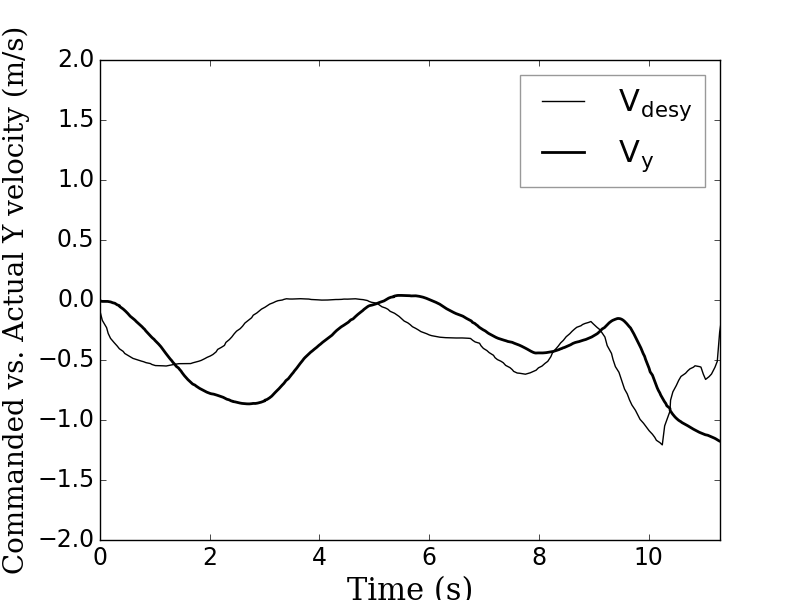}
     \caption{}
 \end{subfigure}
 \begin{subfigure}{0.246\columnwidth}
     \centering
     \includegraphics[scale=0.225]{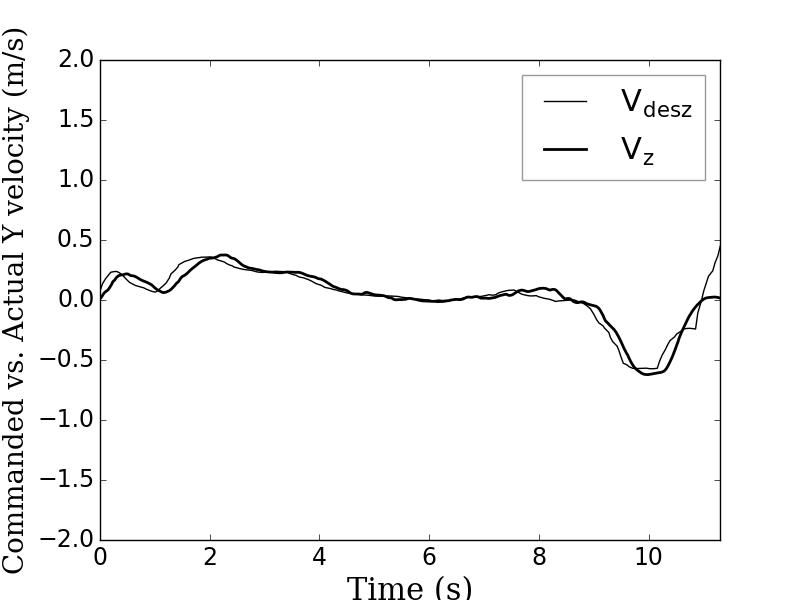}
     \caption{}
 \end{subfigure}
\begin{subfigure}{0.246\columnwidth}
     \centering
     \includegraphics[scale=0.225]{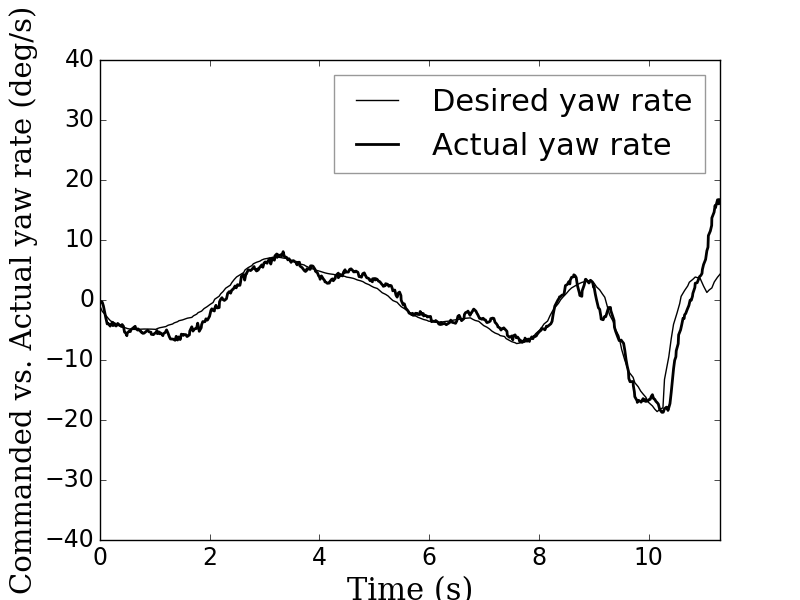}
     \caption{}
 \end{subfigure}
 \begin{subfigure}{0.33\columnwidth}
        \centering
        \includegraphics[scale=0.3]{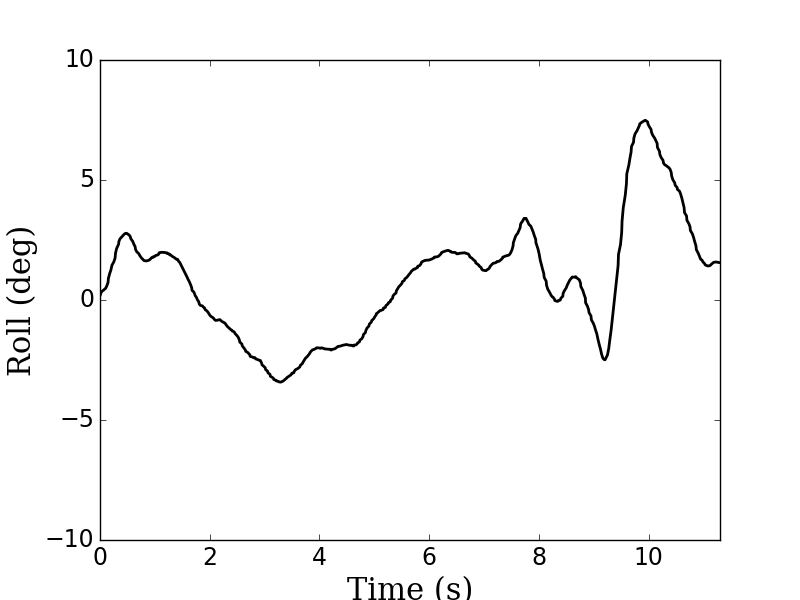}
        \caption{}
    \end{subfigure}
    \begin{subfigure}{0.33\columnwidth}
        \centering
        \includegraphics[scale=0.3]{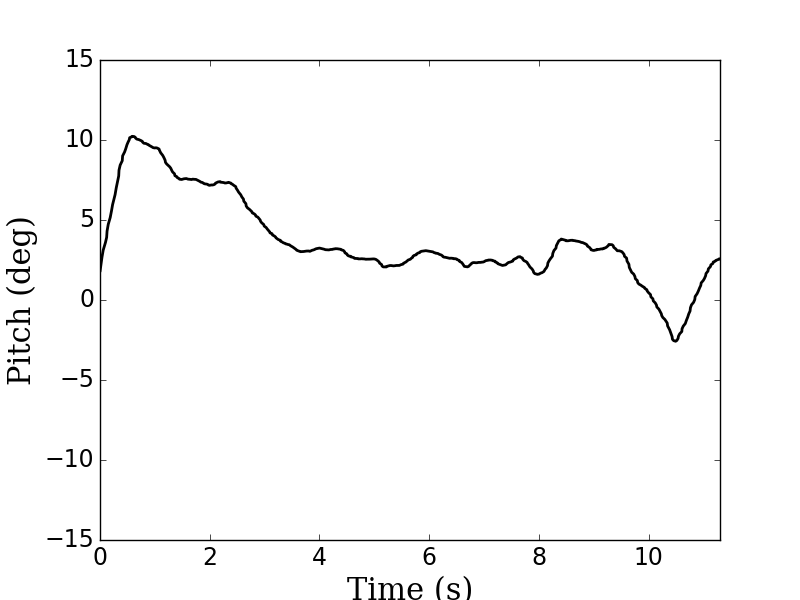}
        \caption{}
    \end{subfigure}
    \begin{subfigure}{0.33\columnwidth}
        \centering
        \includegraphics[scale=0.3]{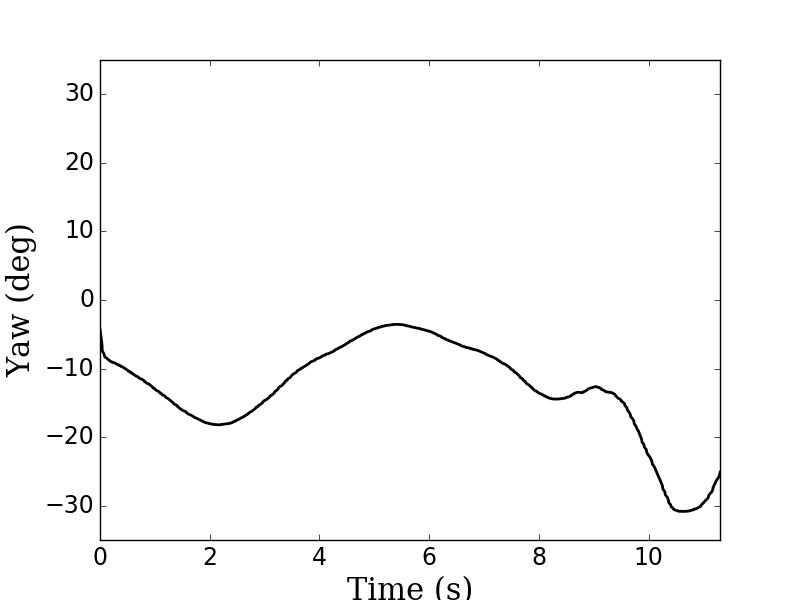}
        \caption{}
    \end{subfigure}
    \begin{subfigure}{0.33\columnwidth}
        \centering
        \includegraphics[scale=0.3]{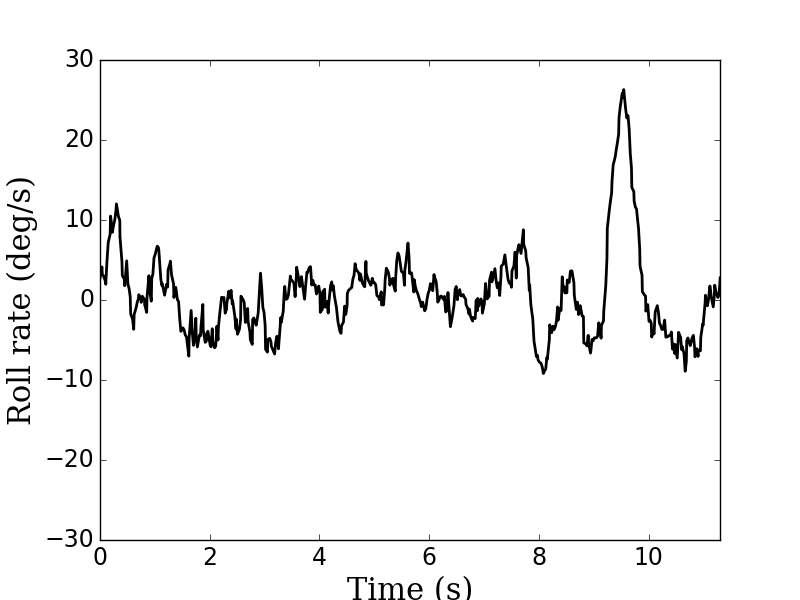}
        \caption{ }
    \end{subfigure}
    \begin{subfigure}{0.33\columnwidth}
        \centering
        \includegraphics[scale=0.3]{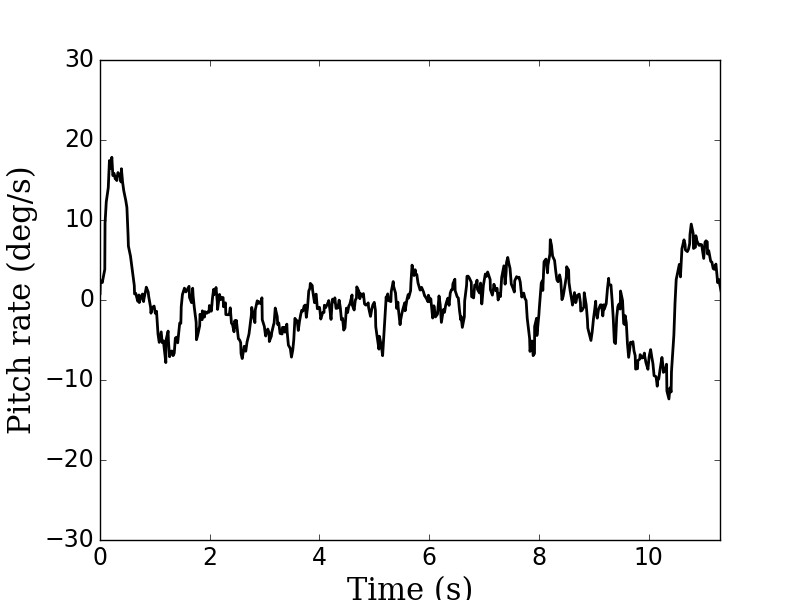}
        \caption{}
    \end{subfigure}
    \begin{subfigure}{0.33\columnwidth}
        \centering
        \includegraphics[scale=0.3]{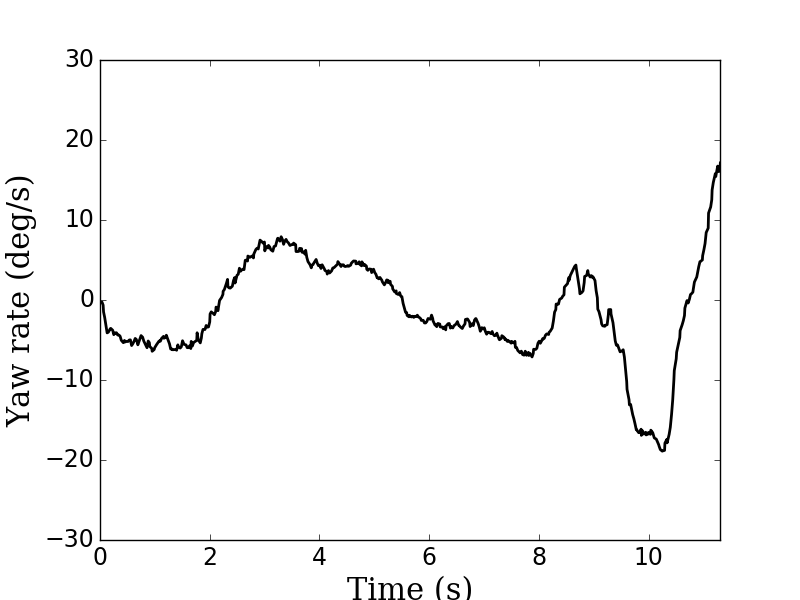}
        \caption{}
    \end{subfigure}
    \caption{Experiment 3 (a) Variation of target pixel in  camera frame (b) Variation of ball depth from camera during grabbing  (c) Commanded yaw rate during grabbing phase. Commanded vs. actual velocity in (d) $x$ direction (e) $y$ direction (f) $z$ direction (g) Commanded vs. actual yaw rate (h) Roll angle (i) Pitch angle (j) Yaw angle of the interceptor UAV while in grabbing mission (k) Roll rate (l) Pitch rate (m) Yaw rate of the grabber UAV.}
    \label{fig:set3}
\end{figure}

Individual algorithms are initially tested and the various parameters such as camera parameters, tracking distance, adaptive controller gains, etc., are tuned for the outdoor testing.  The flight results for a successful ball grabbing scenario is described here. The variation of different system states during the ball grabbing experiment, for an initial target depth of approximately 10 m, is shown in  Fig. \ref{fig:pixel_plot}.  The variation of target pixel coordinates in the camera frame during the engagement is shown in Fig. \ref{fig:pixel_xy}. The variation in the ball depth during the outdoor grabbing engagement is shown in Fig. \ref{fig:depth}. Clearly, the ball depth goes to zero for successful grabbing.  As the ball depth is measured from the visual information, the ball depth does not reduce smoothly as observed in simulation results. The desired yaw rate obtained from the guidance law is plotted in Fig. \ref{fig:comm_color_yaw_rate}, where the different colors correspond to the different pixel trajectories given in Fig. \ref{fig:pixel_xy}. The  tracking of commanded velocities and yaw  rate  by the  interceptor UAV  is shown in Fig. \ref{fig:comm_vs_act_x}-\ref{fig:comm_vs_act_yaw}. As it is evident from the plots, the system is able to track the commanded values. The variation of the attitudes  during the grabbing  engagement is shown in Fig. \ref{fig:roll}-\ref{fig:yaw}. The corresponding attitude rates are shown in Fig. \ref{fig:roll_rate}-\ref{fig:yaw_rate}. As can be observed from these, the vehicle remains stable during the entire mission. Figure \ref{fig:ball_grabbing} and  Fig. \ref{fig:balloon_popping} show the relevant frames of grabbing mission and balloon popping mission, respectively.

After the validation of individual algorithms,  the overall system architecture is verified through SITL and HITL. The successful testing is followed by deployment of the hardware described in Section \ref{s8}. One UAV (Fig. \ref{f6}a) is attached with the balloon popping mechanism, while two UAVs are attached with the basket-type passive gripper (Fig. \ref{f6}c). The operation manager is initialized with the launch and mission parameters. An end-to-end test on the flight modes was conducted, the results of which are included here. Numerous flight tests were carried out to evaluate the performance of the overall system. The mission achieved success 7 out of 10 times. The unsuccessful cases were those which operated in high wind conditions or faced failure in detection which was rectified later. Additional experimental data for two more successful missions are shown in Fig. \ref{fig:set2} and Fig. \ref{fig:set3}. Each of these are for different initial conditions of the UAVs. The corresponding flight test videos\footnote{\href{https://youtu.be/65MGSjwwWsM}{Experiment videos}} demonstrate the real time performance of the developed framework.

\subsection{Competition experiences} 
\begin{figure}[t]
    \centering
    \begin{subfigure}{0.32\columnwidth}
        \centering
        \includegraphics[scale=0.2]{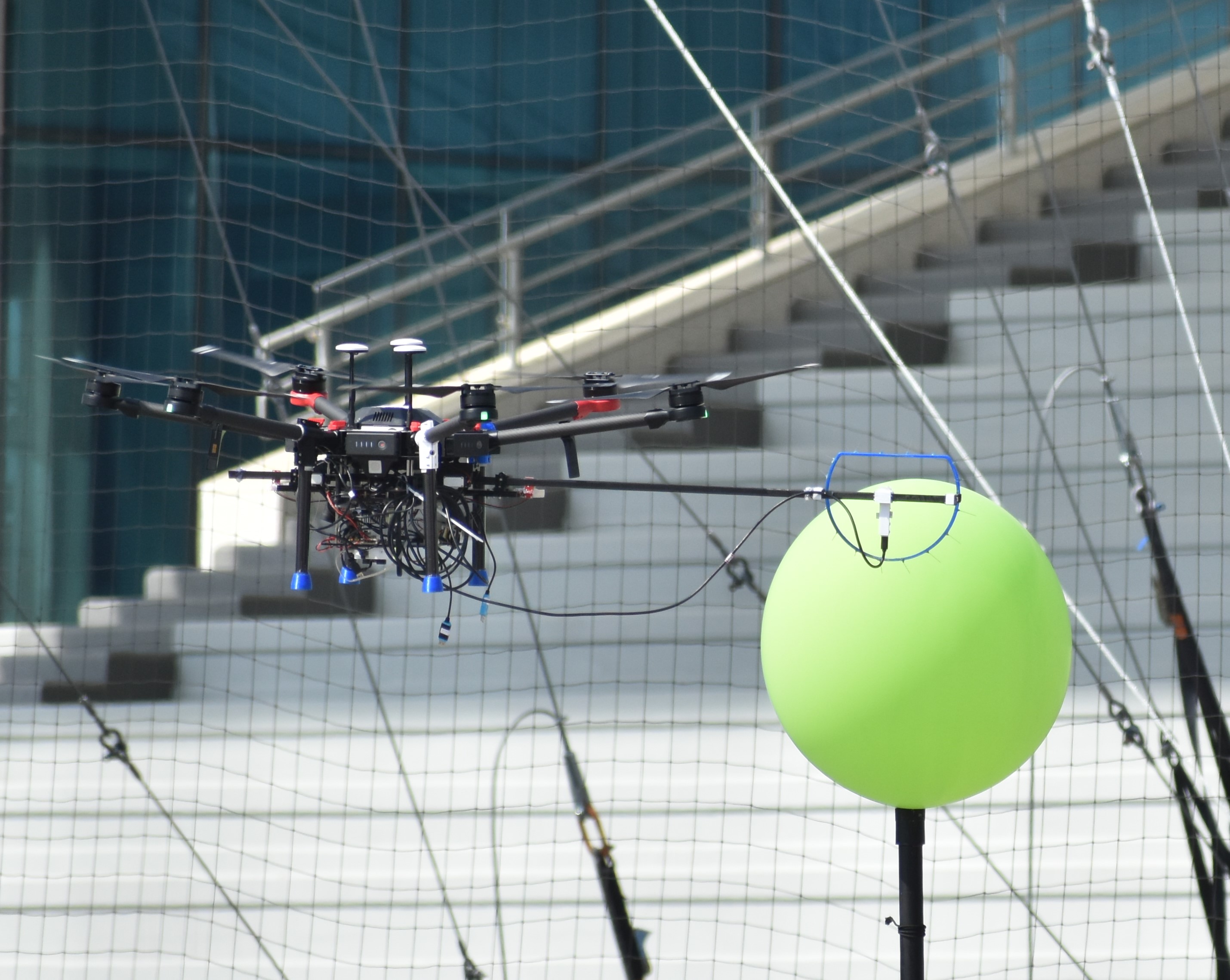}
        \subcaption{}
    \end{subfigure}
    \begin{subfigure}{0.32\columnwidth}
        \centering
        \includegraphics[scale=0.25]{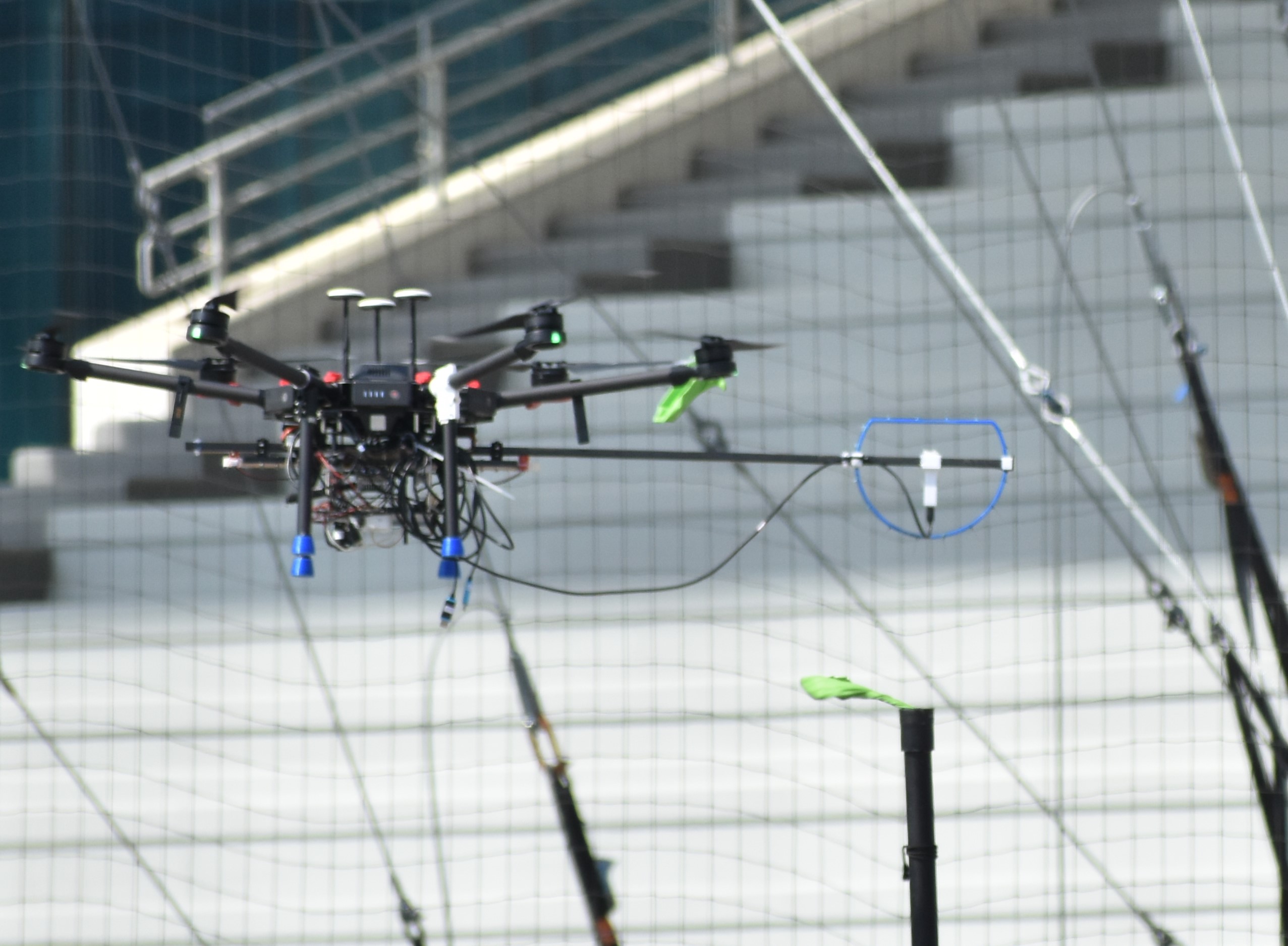}
        \subcaption{}
    \end{subfigure}
    \begin{subfigure}{0.313\columnwidth}
        \centering
        \includegraphics[scale=0.255]{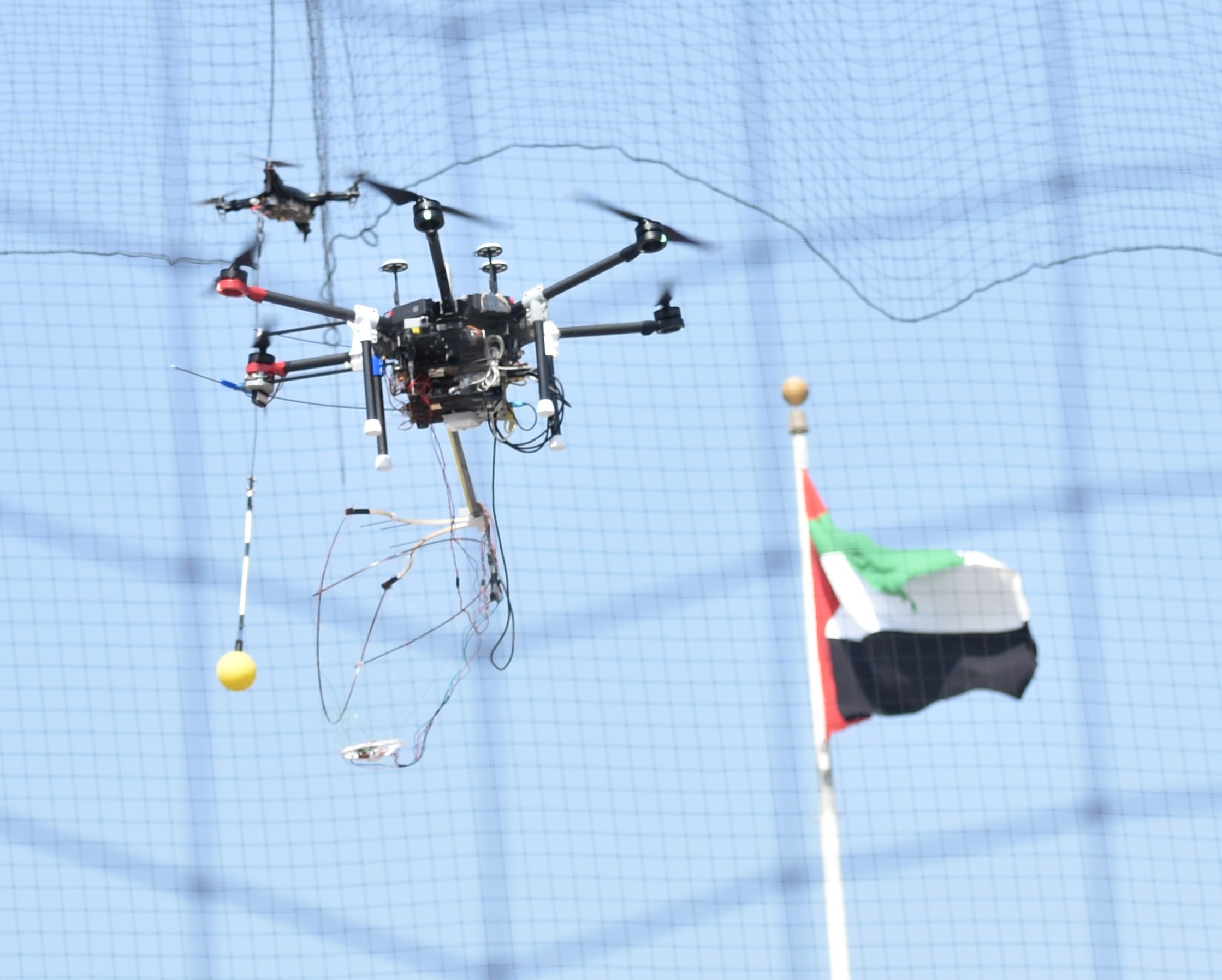}
        \subcaption{}
    \end{subfigure}
    \caption{From MBZIRC 2020 (a) UAV approaching to pop the balloon (b) Successful popping (c) UAV approaching to grab ball.}
    \label{fig:competition_snapshot}
\end{figure}

The competition arena was small in size compared to the originally announced dimensions. The speed of the target UAV was specified to be low/high without any quantitative mentions. The standby area of our architecture, for the interception, was adjusted based on the location of the sun during the time of competition, for improved performance of the vision module. The weather was very sunny during the day. The initial position of the UAVs and the angle of interception for the mission were set based on the time of the day to ensure that the cameras were facing away from the sun to minimize the glare and visual artifacts that could be caused due to lens flare. Several mission planning parameters like the level of exploration were adjusted, as full 25 m height was not available in the arena due to the sag of enclosing net. Also, the mission parameters were made more conservative due to the huge uncertainty in the GPS location accuracy at the venue. This was due to a large number of steel structures around the arena.  The communication band of our system was not fully compatible with the network provided by the organizers during the competition. So, it was not possible to establish multi-UAV communication architecture properly in the arena. The code structure was modified to perform individual missions without inter-agent communication; however, it was not robust and efficient like the previously developed architecture. We could only perform some parts of the mission. Few snapshots of the mission during MBZIRC 2020, are shown in Fig. \ref{fig:competition_snapshot} and relevant video can be found in the link\footnote{\href{https://youtu.be/1AjRck-jn1E}{MBZIRC20 videos}} below.
 
\section{Discussions} \label{s10}
The main challenges of the autonomous mission involving aerial interception and grabbing of stationary and moving targets using visual information in an unknown outdoor environment are addressed in this work. Although aerial interception using visual information is highly affected due to environmental conditions, accuracy is improved with an efficient manipulator system integrated with robust guidance law. The developed system is able to grab the ball with a high success rate from the target moving at a speed of 5 m/s. Grabbing success rate in tail-chase mode is found to be higher than the head-on mode as the time of engagement is comparatively less for head-on mode.  The system was able to track a target moving in a figure-of-eight trajectory at a speed of 4 m/s  at a distance of  8 m. At higher tracking distance, the depth from the camera is not robust for consistent tracking.  It was not easy to track the target in the curved portions of the figure-of-eight trajectory when the target speed exceeded 4 m/s.                

The following critical lessons were learned during system development and outdoor flight test experiments. 
\begin{enumerate}

    \item  In the external environment, the delay and the uncertainty in the visual information can play a significant role in a UAV mission involving moving objects. The overall system design should accommodate these factors related to visual measurements through hardware design or algorithm design. The aerial manipulator should be designed iteratively after getting feedback from the flight test results and efficiency of the visual-guidance system. 
    
    \item In outdoor conditions, vision algorithms involving target depth should be avoided for improved engagement efficiency with dynamic objects.
    
    \item  The centralized command center for task allocation during a  multi-UAV mission will have better efficiency compared to decentralized architecture; however, the whole mission can fail in the case of inter-agent communication failure. The software architecture should be robust enough to include the fail-safe scenarios of communication failure so that individual agents can perform their part of the mission instead of a complete breakdown of the whole mission.  
    
    \item The solution for the vision tasks, when designed completely using deep neural networks, can generalize well and address the corner cases that arise due to the separation of detection and tracking of the individual components in the current vision. End-to-end deep neural network designs can perform object detection and tracking by learning complex and robust features when trained properly without relying on manually tuned external components, unlike a manually integrated algorithm.

    \item GPUs with specialized deep learning capabilities like those in Jetson Xavier series can significantly boost the deep learning models in terms of inference speed.

    \item Augmenting the training data (using transformed or synthetically-generated data) while training the deep learning model can facilitate the use of a different set of training samples in each epoch. This way, when trained adequately, the model performs better than a model trained with the same (unaugmented) dataset in all of the epochs.

\end{enumerate}
 \section{Conclusions}\label{s11}
 This paper presented the software and hardware developed to address the problems in Challenge 1 of MBZIRC 2020. The challenge required to autonomously detect, track, and grab a ball from a maneuvering target as well as autonomously detect and approach stationary balloons to intercept them. An in-house designed and developed manipulation mechanism is employed to achieve the tasks of grabbing and popping. The sideways design of the manipulator ensured safe detachment of the ball and popping of balloons. The passive manipulator design reduced the energy demand of the overall system. A vision based guidance law was formulated to ensure the ball grabbing. Two UAVs were used to coordinate among themselves to explore, track and grab the ball. Popping was executed using a similar guidance philosophy, with minor modifications to suit the purpose. Underlying relations governing the algorithms were also presented in detail. The entire mission was handled by an Operation Management System (OMS), which executed multi-UAV task allocation and switching. The OMS architecture also included various features like state monitoring, fail-safe definitions, and task reallocation, which made it a robust framework for autonomous operations. The framework was developed using ROS. Simulation results of the proposed solution were presented to demonstrate the performance of the proposed solution in the virtual environments. The results of field experiments were also included, demonstrating the real time implementation and efficiency of the framework in environment with disturbances. The techniques developed in addressing the challenge have far more generality in applications that go beyond the specified challenge tasks. They can be used in many critical applications indoors and outdoors, where autonomous systems are needed to carry out aerial manipulations tasks. The proposed integrated system can be employed as counter-UAV technology in warfare. The manipulators with minor modification could be used for applications like package passing between drones in long distance delivery, repair and monitoring of inaccessible structures, fruit picking in orchards, among many others.

\section{Index to Multimedia Extensions}
The following videos are available as supporting information in the online version of this article.
\begin{table}[htb!]
    \centering
    \begin{tabular}{|c| c| l|}
    \hline
        Extension & Media type & Description\\
    \hline
    \hline
       1  & Video &ROS-Gazebo simulations \\
        2 &  Video & Field-experiments \\
      3 & Video& Performance at competition\\
         \hline
    \end{tabular}
\end{table}
\subsection*{Author credits}

\textit{Lima Agnel Tony}: Integration of different systems, project management, sub-team coordination and interface, concept development, pilot, writing the paper.

\textit{Shuvrangshu Jana}: Concept development, guidance and estimation codes, mission planner, software architecture, writing the paper.

\textit{Varun V.P.}: Coding support, Gazebo SITL simulation, GCS management, writing the paper.

\textit{Aashay Anil Bhise}: Guidance, control, and estimation software, geo-fencing codes, mission planner development, code stack finalization, Gazebo SITL simulation, GCS management, writing the paper.

\textit{Aruul Mozhi Varman S.}: Vision code development,  writing the paper.

\textit{Vidyadhara B. V.}: Manipulators and propeller guard design and prototyping, design of component mounts and landing gears.
  
\textit{Mohitvishnu S. Gadde}: Main pilot, electronics integration, communication set-up, mission planner.

\textit{Raghu Krishnapuram}: Lead in vision algorithm design and implementation, writing the paper.

\textit{Debasish Ghose}: Overall project lead, conceptualization, writing the paper.

\subsection*{Acknowledgements}
We would like to acknowledge the Robert Bosch Center for Cyber Physical Systems, Indian Institute of Science, Bangalore, and Khalifa University, Abu Dhabi, for partial financial support. We would also like to thank fellow team members from IISc and all members of Guidance, Control, and Decision Systems Laboratory, especially,  Abhishek Kashyap, Rahul Ravichandran, Aakriti Agrawal, Rohitkumar Arasanipalai, Kuntal Ghosh, Amit Singh, Kumar Ankit and Narayani Vedam for their valuable contributions towards this competition. We also thank Mayur Shewale and Vishal Pal for their valuable time. We appreciate all the help and support from our collaborator, Tata Consultancy Services.
 \FloatBarrier

\bibliographystyle{apalike}
\bibliography{reference}
\end{document}